\documentclass{article}
\usepackage[table,dvipsnames]{xcolor}
\usepackage{iclr2027_conference,times}

\usepackage[utf8]{inputenc} 
\usepackage[T1]{fontenc}    
\usepackage{url}            
\usepackage{booktabs}       
\usepackage{amsfonts}       
\usepackage{nicefrac}       
\usepackage{microtype}      
\usepackage{minitoc}
\usepackage{wrapfig}
\usepackage[colorlinks,linktoc=all]{hyperref}
\usepackage{tikz}
\usepackage{amsmath,amsfonts,mathtools}
\usepackage{graphicx}
\usepackage{amsmath}
\usepackage{algorithm}
\usepackage{algpseudocode}
\hypersetup{citecolor=Blue}
\hypersetup{linkcolor=Blue}
\hypersetup{urlcolor=Blue}
\usepackage{titletoc}

\usepackage{amsmath}
\usepackage{amssymb}
\usepackage{mathtools}
\usepackage{amsthm}
\usepackage{xspace}
\usepackage{pifont}
\usepackage{minitoc}

\usepackage{tabularx, booktabs}
\usepackage{tikz}
\usepackage{CJKutf8}
\usepackage{graphicx}
\usepackage{tcolorbox}

\usepackage{amsmath,amsfonts,bm}

\def\eqref#1{equation~\ref{#1}}

\def\1{\bm{1}}

\DeclareMathAlphabet{\mathsfit}{\encodingdefault}{\sfdefault}{m}{sl}
\SetMathAlphabet{\mathsfit}{bold}{\encodingdefault}{\sfdefault}{bx}{n}

\usepackage[capitalize,noabbrev]{cleveref}
\usepackage{paralist}
\newcolumntype{Y}{>{\centering\arraybackslash}X}

\theoremstyle{plain}

\theoremstyle{definition}

\theoremstyle{remark}

\usepackage[disable,textsize=tiny]{todonotes}
\usepackage{multirow}
\usepackage{subcaption}

\usepackage{enumitem}

\usepackage{makecell}

\newcommand{\methodname}{\textsc{\textbf{GenMem}}\xspace}

\title{{\textsc{{GenMem}}\xspace}: Generative Symbolic Memory for Self-Evolving Harness}
\author{
  \textbf{Xinke Jiang\textsuperscript{1,2,3,*}},
  \textbf{Tao Feng\textsuperscript{*}},
  \textbf{Weixuan Xu\textsuperscript{1,*}},
  \textbf{Zhixin Zhang\textsuperscript{1,2,3}},
  \textbf{Zhibang Yang\textsuperscript{1,2,3}},\\
  \textbf{Wentao Zhang},
  \textbf{Runchuan Zhu\textsuperscript{1}},
  \textbf{Xu Chu\textsuperscript{2,3,4,\ensuremath{\dagger}}},
  \textbf{Junfeng Zhao\textsuperscript{2,3,\ensuremath{\dagger}}},
  \textbf{Yasha Wang\textsuperscript{1,5,\ensuremath{\dagger}}}\\[4pt]
  \normalfont\textsuperscript{1}National Engineering Research Center of Software Engineering, Peking University, Beijing, China\\
  \normalfont\textsuperscript{2}School of Computer Science, Peking University, Beijing, China\\
  \normalfont\textsuperscript{3}Key Laboratory of High Confidence Software Technologies, Ministry of Education, Beijing, China\\
  \normalfont\textsuperscript{4}Center on Frontiers of Computing Studies, Peking University, Beijing, China\\
  \normalfont\textsuperscript{5}Peking University Information Technology Institute (Tianjin Binhai), Tianjin, China\\[2pt]
  \normalfont\small
  \textsuperscript{*}Equal contribution.\quad
  \textsuperscript{\ensuremath{\dagger}}Corresponding authors.\\
  \normalfont\small
  \texttt{\{xinkejiang, yangzb\}@stu.pku.edu.cn;}
}
\iclrfinalcopy

\usepackage{listings}
\lstdefinestyle{promptstyle}{
  basicstyle=\footnotesize\ttfamily,
  breaklines=true,
  breakatwhitespace=false,
  columns=fullflexible,
  keepspaces=true,
  showstringspaces=false,
  frame=none,
  xleftmargin=0pt,
  breakindent=0pt
}

\usepackage{colortbl}
\usepackage{booktabs,array,caption}
\definecolor{ReviewOne}{HTML}{000000}
\definecolor{ReviewTwo}{HTML}{000000}
\definecolor{ReviewThree}{HTML}{000000}
\DeclareRobustCommand{\reviewOne}[1]{\textcolor{ReviewOne}{#1}}
\DeclareRobustCommand{\reviewTwo}[1]{\textcolor{ReviewTwo}{#1}}
\DeclareRobustCommand{\reviewThree}[1]{\textcolor{ReviewThree}{#1}}

\definecolor{DeltaBlue}{RGB}{45,112,176}
\definecolor{DeltaPink}{RGB}{204,78,132}
\definecolor{ResultHeader}{RGB}{232,235,239}
\definecolor{ResultGroup}{RGB}{245,246,248}
\definecolor{ResultBest}{RGB}{210,230,250}
\providecommand{\resultcell}{}
\renewcommand{\resultcell}[2]{%
  \hbox{#1\hskip0.02em{\fontsize{5.8pt}{6.4pt}\selectfont #2}}%
}
\providecommand{\upchg}{}
\renewcommand{\upchg}[1]{{\color{DeltaBlue}\fontsize{4.2pt}{4.2pt}\selectfont $\Delta$\,#1}}
\providecommand{\downchg}{}
\renewcommand{\downchg}[1]{{\color{DeltaPink}\fontsize{4.2pt}{4.2pt}\selectfont $\Delta$\,#1}}
\providecommand{\samechg}{}
\renewcommand{\samechg}[1]{{\color{gray!70!black}\fontsize{4.2pt}{5.8pt}\selectfont $\Delta$\,#1}}

\providecommand{\methodname}{\textsc{\textbf{GenMem}}}

\tcbuselibrary{breakable,skins}
\newtcolorbox{promptbox}[1]{%
  breakable, enhanced, arc=2pt,
  colback=white, colframe=gray!55, boxrule=0.6pt,
  title={#1}, coltitle=white, colbacktitle=gray!60,
  fonttitle=\bfseries\small,
  left=4pt, right=4pt, top=3pt, bottom=3pt,
  before skip=6pt, after skip=6pt
}

\begin{document}
\maketitle
\fancyhead{}

\doparttoc \faketableofcontents
\begin{abstract}
Long-term memory supports the self-evolution of LLM agents by retaining experience and skills across tasks and enabling their \textit{retrieval, reuse}, and \textit{revision} in subsequent long-horizon decision-making.
Yet existing  memory management approaches remain limited to discriminative retrieval and to address the sparse, hierarchical, and highly redundant structure of reusable experience: only a small, task-dependent subset of trajectories and memories warrants retention, retrieval, or revision.
Learning these operations is further complicated by \textit{\textbf{sparse, delayed}}, and \textit{\textbf{indirect}} task-level feedback, with weak supervision across the memory lifecycle.
Moreover, continual memory evolution introduces an architectural tension as \emph{\textbf{addressing invariance}}: stored experience is perpetually revised, yet the addressing interface consumed by learned retrieval policies must remain stable.
To address, we present \methodname{}, which reformulates memory management as generative symbolic addressing.
Its core mechanism is the \textbf{Symbolic Identifier (SID)}, a multi-level discrete token tuple drawn from a Cartesian-product address space that factorizes a million-scale sparse memory space using fewer than one hundred discrete symbols.
Architecturally, \methodname{} couples a MemRetriever and a MemEvolver within a multi-agent harness, \reviewOne{trained via GRPO with process and outcome rewards}.
Beyond accuracy, we find that SIDs exhibit emergent \textit{\textbf{neuro-symbolic properties}}: they can be interleaved with natural-language tokens and are \emph{\textbf{compositional}}---combining multiple SIDs composes their stored experiences to address novel situations not covered by any single entry.

\end{abstract}
\begingroup
\raggedbottom
\section{Introduction}

\textbf{\textit{Long-term memory}} enables self-evolving LLM agents to retain reusable experience and skills across tasks, supporting long-horizon decision-making~\citep{Wang2023Voyager,Hong2023MetaGPT,zhang2026memoryactionautonomouscontext}. Unlike encyclopedic knowledge, such experience-centric memory is action-oriented, context-conditioned, and continuously evolving~\citep{Zhang2024MemorySurvey}. The central question therefore shifts from \emph{which memory most resembles a query} to \emph{which experience best advances the agent's current strategy}---a question that must remain answerable even as the experiences themselves evolve.
Despite promising progress~\citep{zheng2026skillrouter,zhang2025memgen,Zhang2026MemRL}, existing memory-augmented agents~\citep{chhikara2025mem0,xu2025amemagenticmemoryllm} remain largely anchored to a \textbf{dense retrieval paradigm}: they access experience through discriminative scoring, with learned addressing coupled to content representations. Despite these advances, two fundamental challenges remain:

\noindent \ding{182} \textit{\textbf{Sparsity Challenge.}} Experience is highly redundant: across the memory lifecycle (acquisition, organization, retrieval, and evolution), only a small fraction carries reusable value; only a narrow region is relevant to any given query; and refinement is concentrated on repeatedly exercised entries. The experiences that generalize tend to cluster around shared sub-skills---for instance, distinct web-navigation tasks may all reuse a ``search-then-filter'' routine---yet flat dense stores do not explicitly encode this structure, treating entries uniformly and diluting key associations as the bank grows~\citep{Zhang2024MemorySurvey}. Appendix~\ref{sec:app_sparsity} details these sparsity dimensions. Supervision compounds the problem: an experience's utility emerges only through delayed, indirect task outcomes, so reliance on sparse terminal rewards~\citep{Zhang2026MemRL,Ma2026FineMem} makes it difficult to assign supervision to individual memory operations.

\begin{wrapfigure}{R}{0.55\textwidth}
\centering
\vspace{-0.4cm}
\includegraphics[width=\linewidth]{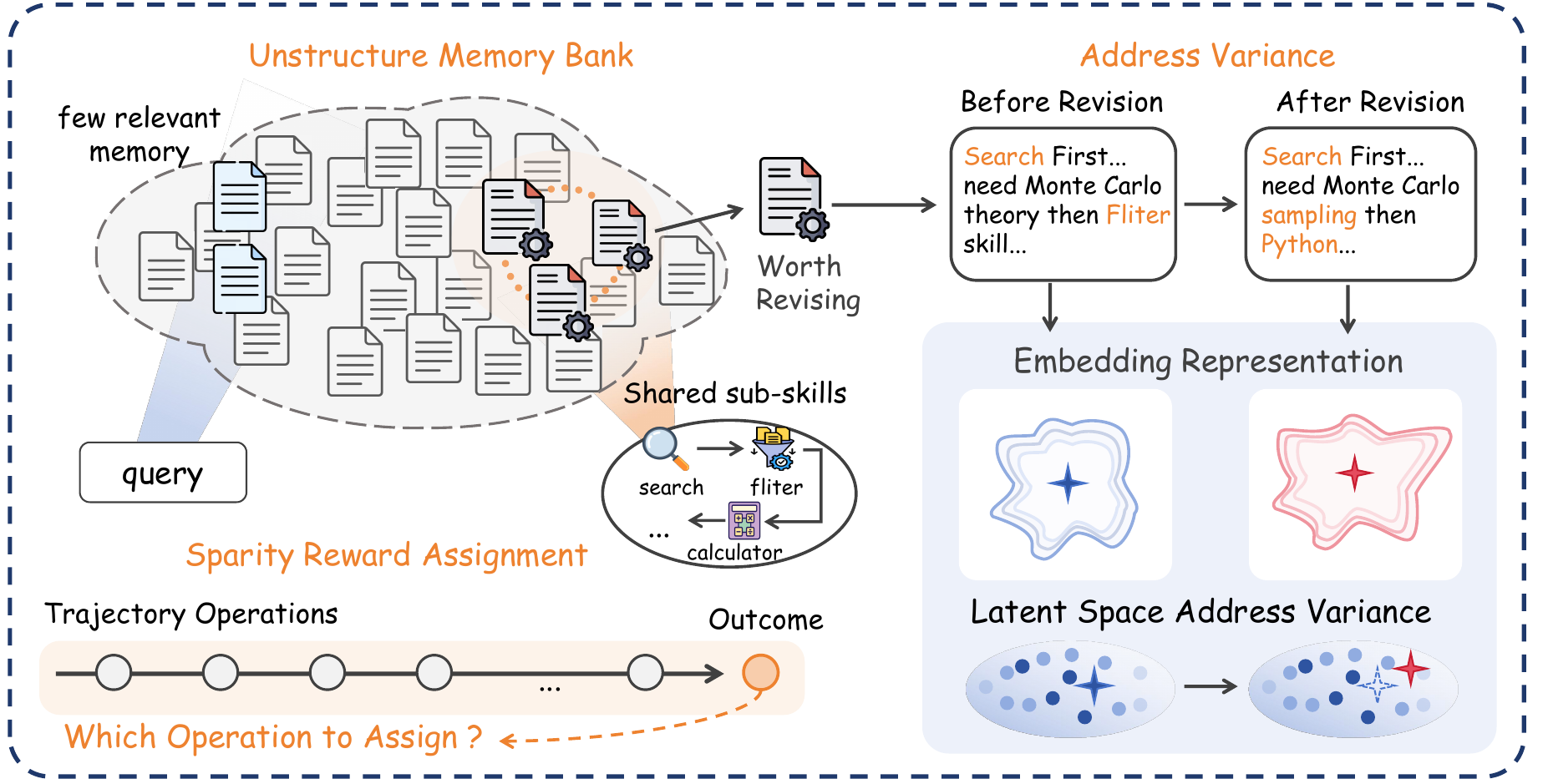}
\caption{Two challenges in evolving experience memory.
\textbf{Left:} Reusable experience is sparse and organized around
shared sub-skills, while delayed task rewards hinder credit
assignment.
\textbf{Right:} Content updates can shift embedding-based retrieval,
motivating stable addresses that remain invariant as experience evolves.}
\label{fig:intro}
\vspace{-0.2cm}
\end{wrapfigure}

\noindent \ding{183} \textit{\textbf{Address Variance Challenge.}} A deeper architectural tension arises when experience continuously evolves (Figure~\ref{fig:intro}): \textbf{the content at an address changes, but the address itself must remain stable.} A self-evolving agent refines SOPs, merges redundant skills, and discards outdated procedures---yet the memory system must remain reliably addressable throughout. When retrieval relies on representations derived from current memory content, each revision can silently change which queries reach an entry, even when the entry still serves the same functional role (Appendix~\ref{sec:app_disc_limits}). We term the requirement to preserve an entry's address across such revisions \emph{\textbf{addressing invariance}}. Satisfying it demands decoupling \emph{where} from \emph{what}: stable discrete coordinates identify functional regions of experience space, while text behind those coordinates remains freely editable. \reviewThree{Frozen dense keys also preserve addresses.} This decoupling reframes what agent should learn: not particular content versions that become outdated with each revision, but a persistent mapping from a problem to \textbf{functional role} of required experience.

Our key motivation draws on a recurring principle: \textbf{\textit{structured and sparse architectures exploit hidden regularities in high-dimensional spaces}}---LoRA through low-rank updates, MoE through sparse activation, and Engram through conditional knowledge storage~\citep{hu2022lora,fedus2022switch,cheng2026engram}. Analogously, reusable experience concentrates in \textbf{sparse, hierarchically structured regions} amid substantial redundancy, motivating selective addressing that remains stable as content evolves.
We therefore propose \methodname{}, a \textbf{generative symbolic memory framework} that unifies memory access and update targeting through \textbf{Symbolic Identifier (SID)} generation. Each SID is a hierarchical discrete token tuple \reviewThree{obtained by quantizing an experience embedding and then successively quantizing its residuals}. \reviewThree{The resulting coarse-to-fine codes form a Cartesian-product address space for indexing experiences that vary in task domain, tool modality, and strategy pattern.} Constructed offline via context-aware RQ-KMeans, SIDs organize functionally proximate experiences through shared prefixes~\citep{lee2022autoregressive,rajput2023recommender}.
The memory agent generates target SIDs from its current context for selective access, while online revisions update payloads in place, preserving existing SID assignments under fixed codebooks.

Concretely, \methodname{} couples two memory agents in a closed loop: \textbf{MemRetriever} generates a SID and rewrites the retrieved experience into task-adaptive support; a multi-agent harness executes the task using this support; and \textbf{MemEvolver} distills the resulting trajectory and generates an update SID for insertion or revision. The pipeline comprises alignment pretraining, mid-training, GRPO post-training, and online self-evolution. The first two stages teach SID generation and its integration with reasoning; \reviewOne{GRPO combines process and outcome rewards}, providing fine-grained supervision beyond sparse terminal feedback. During online self-evolution, the agents continue refining memory through the same stable SID interface with model parameters and codebooks fixed. Our contributions are fourfold:
\begin{itemize}[leftmargin=*,noitemsep,topsep=2pt]
    \item We propose \methodname{}, the first generative symbolic addressing framework for evolving experience memory, unifying memory retrieval and evolution through SID generation over a sparse, hierarchically structured space (Section~\ref{sec:prelim}).

     \item We instantiate this framework as a closed-loop system coupling MemRetriever, a multi-agent harness, and MemEvolver, with staged training using \reviewOne{process and outcome rewards} to support online memory evolution (Sections~\ref{sec:sid_construction}--\ref{sec:posttraining}).

    \item Experiments across embodied planning, web navigation, search-augmented QA, medical reasoning and deep research evaluate \methodname{} against strong memory-agent baselines (Section~\ref{sec:experiments}).

    \item \reviewTwo{We show that SIDs act as neuro-symbolic conditioning tokens that can be combined in the same input as a natural-language query. Without experience text, longer SID prefixes improve accuracy on GPQA, and combining multiple reranked SIDs improves accuracy over using a single SID (Section~\ref{sec:symbolic_language}).}
\end{itemize}
\endgroup

\section{\methodname{}: Symbolic Memory Addressing System}
\label{sec:method}
\label{sec:sid_system}

\begin{figure*}[t]
    \centering
    \vspace{-0.15cm}\includegraphics[width=0.99\textwidth]{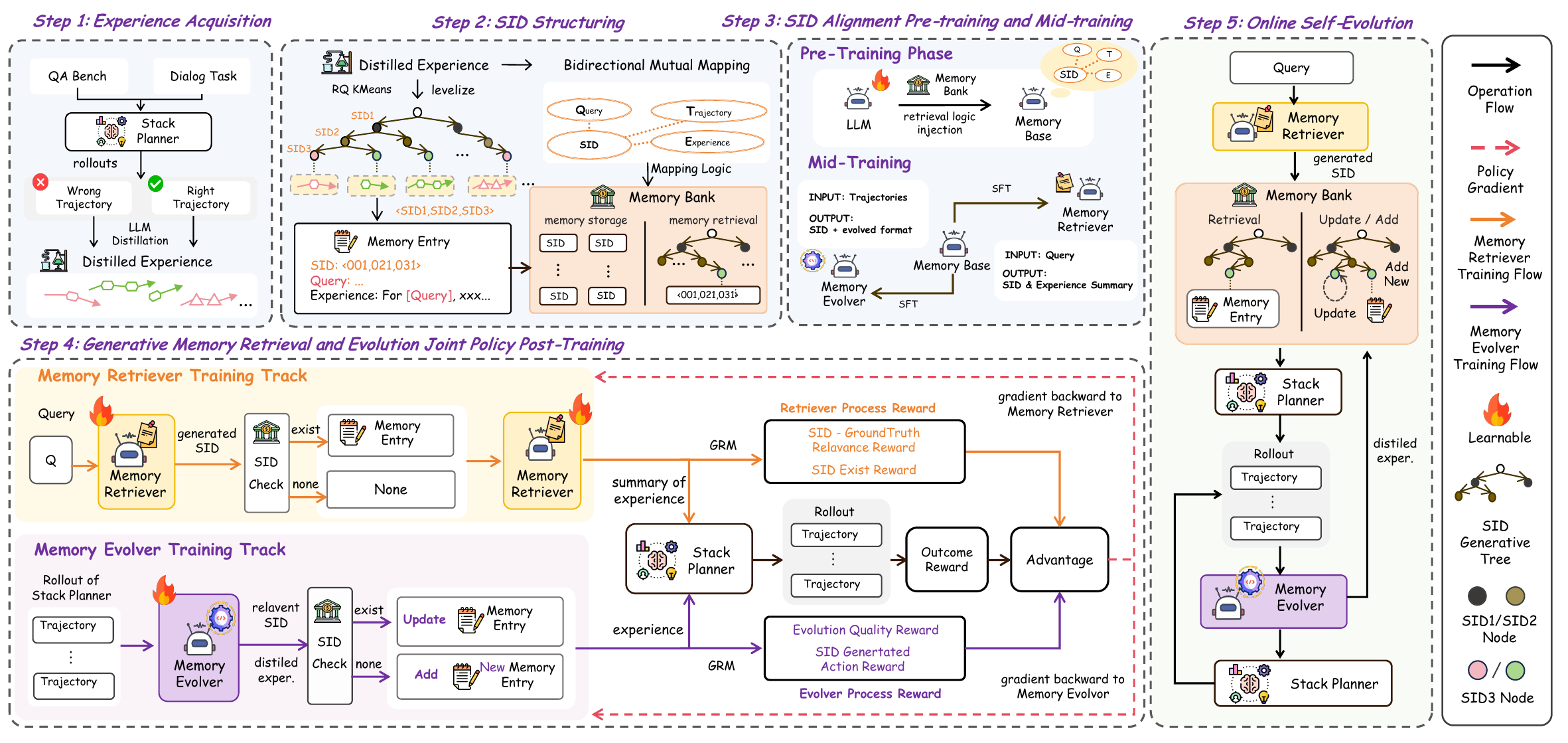}
    \vspace{-0.25cm}
    \caption{Overview of 
    \methodname{}.}
    \label{fig:framework}
    \vspace{-0.34cm}
\end{figure*}

\methodname{} unifies memory retrieval and evolution through a
shared symbolic address space, enabling selective access to
structured experience while preserving stable addresses under
content updates.

\paragraph{\ding{182} \textit{Memory Symbolic Identifier Definition}.}
\label{sec:sid_definition}
\label{sec:prelim}
\label{sec:disc_limits}
Let $\mathcal{M}=\{\textbf{m}_i\}_{i=1}^{\textbf{N}}$ denote the external memory
bank, where each entry $\textbf{m}_i=(\textbf{x}_i,\textbf{z}_i)$ pairs a textual experience
payload $\textbf{x}_i$ with a Symbolic Identifier (SID) as $\textbf{z}_i$. We define
each SID as a tuple of $L$ discrete symbols as:
\begin{equation}
    \textbf{z}_i=(\textbf{s}_i^1,\ldots,\textbf{s}_i^L)\in\mathcal{Z},
    \quad
    \mathcal{Z}=[\textbf{N}_1]\times\cdots\times[\textbf{N}_L],
    \quad
    \textbf{s}_i^\ell\in[\textbf{N}_\ell],
    \label{eq:sid_definition}
\end{equation}
where $[\textbf{N}_\ell]=\{1,\ldots,\textbf{N}_\ell\}$ indexes the symbols at
level $\ell$. Given a context $\textbf{c}$, a symbolic addressing policy
generates $\textbf{z}=\textsc{\textbf{Addr}}_{\theta}(\textbf{c})$, and lookup returns
the associated payload $\textbf{x}=\textsc{\textbf{Lookup}}(\mathcal{M},\textbf{z})$.
The context $\textbf{c}$ may be a query, a trajectory, or a newly distilled
experience, allowing the same address space to support retrieval
and update targeting.

In \methodname{}, SIDs represent $\prod_{\ell=1}^{L}\textbf{N}_\ell$ possible addresses
using $\sum_{\ell=1}^{L}\textbf{N}_\ell$ level-specific symbols, with
shared prefixes organizing experience hierarchically for
selective access. Crucially, a SID identifies a memory slot
rather than a particular content version: online revisions
transform $(\textbf{x}_i,\textbf{z}_i)\xrightarrow{\textsc{\textit{Evolve}}}(\textbf{x}_i',\textbf{z}_i)$,
allowing experience to evolve while its address remains stable.

\paragraph{\ding{183} \textit{Experience Acquisition and SID Construction}.}
\label{sec:experience_generation}
\label{sec:sid_construction}
\label{sec:coldstart}
We construct the initial memory bank through a
\emph{gradient-free cold start}. Across datasets spanning
diverse task domains, the multi-agent harness StackPlanner~\citep{stackplanner}
generates multiple trajectories for each training query.
An LLM extractor compares correct, failed, and unverified
attempts using available reference answers or environment
feedback. It distils reusable procedures and failure lessons
while preserving uncertainty where the traces lack evidence.
The resulting summaries form the seed set
$\mathcal{X}_{\mathrm{seed}}$
(Appendix~\ref{sec:app_experience_generation}).

To obtain discrete addresses that the memory agents can generate,
we quantize continuous experience representations into SID sequences.
First, we encode each summary $\mathbf{x}_i$ via a frozen text encoder:
\begin{equation}
    \mathbf{e}_i
    =f_{\mathrm{enc}}(p_{\mathrm{enc}}\oplus\mathbf{x}_i)
    \in\mathbb{R}^{d},
    \label{eq:experience_encoding}
\end{equation}
where the prefix $p_{\mathrm{enc}}$ emphasizes reusable
strategies, applicability conditions, and procedures.
This semantic embedding $\mathbf{e}_i$ 
encourages the resulting addresses to reflect shared problem-solving
patterns rather than surface-level textual similarity.

We then construct the discrete address space by fitting RQ-KMeans
to the seed embeddings~\citep{jin2025onerec}: 
At the first level, K-means clusters the embeddings into $\textbf{N}_1$
groups; the resulting cluster centers form the first
\emph{\textbf{codebook}}. Each embedding is assigned to its nearest center,
whose index becomes the first SID symbol. We then subtract this
center from the embedding and fit the next codebook to the
remaining residuals. Repeating this procedure yields $L$
codebooks, with successive levels capturing information not
represented by earlier ones.
Formally, let $\mathcal{C}^{(\ell)}
=\{\mathbf{c}_j^{(\ell)}\}_{j=1}^{N_\ell}$ denote the
resulting cluster centers at level $\ell$. Starting from
$\mathbf{r}_i^{(1)}=\mathbf{e}_i$, 
\reviewThree{we assign codes using the
weighted distance in Eq.~\ref{eq:rq_weighted_distance}, with $w_\ell=0.5$ at every level. We then update the residual as follows.}
{\color{ReviewThree}%
\begin{equation}
\begin{aligned}
    \textbf{s}_i^\ell
    =\operatorname*{arg\,min}_{j\in[\textbf{N}_\ell]}
      d_{\mathrm{weighted}}^{(\ell)}
      \bigl(\mathbf{r}_i^{(\ell)}, \mathbf{c}_j^{(\ell)}\bigr), \quad
    \mathbf{r}_i^{(\ell+1)}
    =\mathbf{r}_i^{(\ell)}
      -\mathbf{c}_{\textbf{s}_i^\ell}^{(\ell)},
    \quad \ell=1,\ldots,L.
\end{aligned}
\label{eq:rq_assignment}
\end{equation}
}
The sequence of selected center indices
$\mathbf{z}_i=(\textbf{s}_i^1,\ldots,\textbf{s}_i^L)$ forms the SID.
Shared prefixes group experiences with common quantized
representations, while later symbols provide finer distinctions.
This yields structured addresses without requiring a separate
symbol for every memory entry. In \methodname{}, we use codebook sizes $(48,16,8,8)$,
yielding $49{,}152$ possible addresses from $80$ tokens, and summarize multiple experience texts assigned to the same SID
into a single memory entry. The codebooks are fixed after
construction. 
\reviewThree{Appendix~\ref{sec:app_rqkmeans} defines the Euclidean and Weighted distances. Appendix~\ref{sec:app_rqkmeans_metrics} reports the codebook comparisons and the three-level prefix-prompt experiment.}

\paragraph{\ding{184} \textit{Generative Memory Retrieval and Evolution}.}
\label{sec:memory_operations}
Given the SID-indexed bank $\mathcal{M}$,
\textbf{MemRetriever} and \textbf{MemEvolver} implement
generative read--write operations around StackPlanner
(Figure~\ref{fig:framework}). 
\begin{itemize}[leftmargin=*,noitemsep,topsep=0pt]
    \item \textbf{\textit{Memory Retrieval}.} MemRetriever generates an address
    $\mathbf{z}_r$ from the query and reasoning context,
    retrieves the associated experience, and rewrites it
    into task-adaptive support $\tilde{\mathbf{x}}_r$.

    \item \textbf{\textit{Memory Execution.}} StackPlanner executes the task
    using this support, producing an answer $\hat{y}$
    and an execution trajectory $\tau$.

    \item \textbf{\textit{Memory Evolution}.} MemEvolver distils the trajectory,
    generates an update address $\mathbf{z}_e$, and consults
    the associated entry to determine an operation $o$
    and updated payload $\tilde{\mathbf{x}}_e$.
\end{itemize}
For a query $q$, the closed memory loop comprises three steps:
\begin{equation}
\begin{aligned}
    \textit{Retrieve:}\quad
    &(\mathbf{z}_r,\tilde{\mathbf{x}}_r)
    =\textsc{\textbf{MemRetriever}}(q,\mathcal{M}), \\
    \textit{Execute:}\quad
    &(\hat{y},\tau)
    =\textsc{\textbf{StackPlanner}}(q,\tilde{\mathbf{x}}_r), \\
    \textit{Evolve:}\quad
    &\mathcal{M}'
    =\textsc{\textbf{Update}}\!\left(
    \mathcal{M};
    \textsc{\textbf{MemEvolver}}(\tau,\mathcal{M})
    \right).
\end{aligned}
\label{eq:pipeline}
\end{equation}
Here, MemEvolver returns an update tuple
$(o,\mathbf{z}_e,\tilde{\mathbf{x}}_e)$, specifying the
operation, target SID, and updated payload.
The operation $o\in\{\texttt{insert},\texttt{revise}\}$
either adds a new entry or updates an existing one.
Revision supports retention, consolidation, and deletion
through unchanged, merged, or empty payloads, respectively.

\section{Memory Policy Optimization and Self-Evolution}
\label{sec:memory_optimization}

Building on the symbolic addressing system, we develop
MemRetriever and MemEvolver through four stages:
SID alignment pretraining, memory-operation mid-training,
joint policy optimization, and online self-evolution.
The first three stages teach the agents to generate,
use, and update symbolic memory; the final stage applies
these learned policies to continually evolve the memory bank.

\paragraph{\ding{182} \textit{SID Alignment Pretraining}.}
\label{sec:pretraining}
We first align task contexts and experience content with
the constructed SID space through five text--SID mappings.
The forward mappings
$q\mapsto\mathbf{z}$, $\tau\mapsto\mathbf{z}$, and
$\mathbf{x}\mapsto\mathbf{z}$ teach address generation,
while the reverse mappings $\mathbf{z}\mapsto\mathbf{x}$
and $\mathbf{z}\mapsto\mathbf{d}$ associate SIDs with
experience content and natural-language semantic
descriptions $\mathbf{d}$, respectively.
For an input context $\mathbf{c}$ and target SID
$\mathbf{z}=(s^1,\ldots,s^L)$, the forward loss and
combined alignment objective are
\begin{align}
    \mathcal{L}_{\mathrm{SID}}(\mathbf{c},\mathbf{z})
    &=-\sum_{\ell=1}^{L}
    \log p_\theta(s^\ell\mid\mathbf{c},s^{1:\ell-1}),
    \quad \mathbf{c}\in\{q,\tau,\mathbf{x}\},
    \label{eq:sid_loss} \\
    \mathcal{L}_{\mathrm{pre}}
    &=\begin{aligned}[t]
        \lambda_q\mathcal{L}_{q\to\mathbf{z}}
        +\lambda_\tau\mathcal{L}_{\tau\to\mathbf{z}}
        +\lambda_x\mathcal{L}_{\mathbf{x}\to\mathbf{z}}+\lambda_{\mathbf{z}\to\mathbf{x}}\mathcal{L}_{\mathbf{z}\to\mathbf{x}}
        +\lambda_{\mathbf{z}\to\mathbf{d}}\mathcal{L}_{\mathbf{z}\to\mathbf{d}},
      \end{aligned}
    \label{eq:pretrain_loss}
\end{align}
where the coefficients
weight the contributions of the respective mappings
specified in Appendix~\ref{sec:app_sid_mapping}.
This stage establishes bidirectional associations
between symbolic addresses and experience, providing
the pretrained checkpoints for subsequent
MemRetriever and MemEvolver training.

\paragraph{\ding{183} \textit{Memory-Operation Mid-Training}.}
\label{sec:midtraining}
SID alignment establishes associations between symbolic
addresses and experience, but effective memory use requires
the model to incorporate these addresses into reasoning
and action. We therefore construct \emph{ReAct-style}
trajectories to train MemRetriever and MemEvolver
separately from the pretrained checkpoints, teaching
each policy to reason about memory, invoke SID-based
operations, and use the returned experience. Specifically, 
\begin{itemize}[leftmargin=*,noitemsep,topsep=0pt]
    \item For \textbf{MemRetriever}, role-specific prompts guide it to reason about the current task,
generate a retrieval SID, inspect the retrieved experience,
and rewrite it into task-adaptive support.
\item For \textbf{MemEvolver}, the prompts guide the model to
analyze an execution trajectory, identify transferable
lessons, generate an update SID, and revise addressed
experience or propose a new entry.
\end{itemize}
We apply \emph{rejection sampling} to retain trajectories
that satisfy interaction format and task-correctness
criteria. Then we apply a stronger teacher model to provide
distillation targets for the reasoning traces and
role-specific outputs: experience rewriting for
MemRetriever and experience evolution for MemEvolver.
Using a standard next-token objective, we fine-tune
each policy on the resulting demonstrations, turning
isolated SID prediction into executable memory
read--write behavior.

\paragraph{\ding{184} \textit{Joint Policy Post-Training}.}
\label{sec:posttraining}
To align memory operations with downstream objectives
and alleviate sparse feedback, we jointly optimize
MemRetriever and MemEvolver from their mid-trained
initialization.
\begin{itemize}[leftmargin=*,noitemsep,topsep=0pt]
    \item \textbf{Process reward.}
    {\color{ReviewOne}%
    MemRetriever receives rewards for level-wise SID agreement, retrieval
    relevance, and rewrite quality. MemEvolver receives rewards for a single SID
    action, protocol validity, SID agreement, lookup of an existing experience,
    evolution quality, and a valid draft. Their process rewards are
    \begin{equation}
    \begin{aligned}
        R_{\mathrm{process}}^{\mathrm{R}}
        &= R_{\mathrm{SID}}^{\mathrm{R}} + R_{\mathrm{rel}} + R_{\mathrm{rewrite}},\\
        R_{\mathrm{process}}^{\mathrm{E}}
        &= R_{\mathrm{act}} + R_{\mathrm{proto}} + R_{\mathrm{SID}}^{\mathrm{E}}
         + R_{\mathrm{lookup}} + R_{\mathrm{evol}} + R_{\mathrm{draft}} .
    \end{aligned}
    \label{eq:process_reward}
    \end{equation}
    SID validity, lookup success, and output format gate MemRetriever's
    reward terms. For MemEvolver, protocol validity gates SID agreement,
    and draft and evolution rewards apply when their corresponding checks pass.
    Appendix~\ref{sec:app_reward} gives the individual weights and conditions.
    }

    \item \textbf{Outcome reward.}
    {\color{ReviewOne}%
    StackPlanner uses the experience rewritten by MemRetriever, falling back
    to the retrieved payload when no rewrite is available. For MemEvolver,
    StackPlanner uses the evolved experience. Both agents receive
    \begin{equation}
        R_{\mathrm{out}}=2\,\mathbb{I}[\text{StackPlanner answers correctly}].
        \label{eq:outcome_reward}
    \end{equation}
    }
\end{itemize}
\vspace{-0.7cm}
For optimization, we adopt \textbf{GRPO}~\citep{shao2024deepseekmath}
\reviewOne{for both memory agents, sampling $G=8$ rollouts per prompt.
For MemEvolver, we assign process rewards to the corresponding output
token spans and the outcome reward at termination. For MemRetriever,
we sum the process and outcome rewards for each rollout, normalize the
totals once within the group, and minimize}

\begin{equation}
\begin{aligned}
    \mathcal{L}_{\mathrm{GRPO}}
    =-\mathbb{E}\Biggl[
    \frac{1}{G}\sum_{i=1}^{G}\frac{1}{|u_i|}
    \sum_{t=1}^{|u_i|}
    \Bigl(
    &\min\!\bigl(
        \rho_{i,t}\hat{A}_i,
        \operatorname{clip}(\rho_{i,t},1-\epsilon,1+\epsilon)
        \hat{A}_i
    \bigr) -\beta D_{\mathrm{KL}}^{i,t}
    \Bigr)
    \Biggr],
\end{aligned}
\label{eq:grpo}
\end{equation}
where 
$\rho_{i,t}$ is the token-level importance ratio
relative to the rollout policy,
\reviewOne{$\hat{A}_i$ is the group-normalized sum of the process and outcome
rewards for MemRetriever rollout $u_i$}, and $D_{\mathrm{KL}}^{i,t}$
penalizes divergence from the reference policy.
\reviewOne{Reward components are specified in}
Appendix~\ref{sec:app_reward}--\ref{sec:app_reward_outcome}.

\paragraph{\ding{185} \textit{Online Memory Self-Evolution}.}
\label{sec:online}
After training, model parameters, encoder, and SID codebooks are frozen. The three modules form a closed loop on a stream of new queries: MemRetriever generates a SID and fetches task-adaptive support, StackPlanner executes, and MemEvolver generates an update address and distils the trajectory into an updated payload. An \texttt{insert} populates an unoccupied SID slot, whereas a \texttt{revise} updates an existing entry. In both cases, the target SID is generated by MemEvolver, not recomputed by encoding and quantizing the updated payload. We run this loop for $T$ steps until the memory bank converges. Adaptation proceeds entirely through external memory: payloads are added or revised within the fixed SID space, requiring neither gradient updates nor reassignment of existing addresses. The update rule and design rationale are provided in Appendices~\ref{sec:app_online_updates} and~\ref{sec:app_design_choices}, respectively.

\section{Experiments}
\label{sec:experiments}
We evaluate \methodname{} through two research questions:
\textbf{(RQ1)} Does \methodname{} improve downstream task performance compared with existing memory-augmented and RL-based methods?
\textbf{(RQ2)} Can online memory evolution improve performance over time while maintaining stable SID addressing?
\subsection{Experimental Setup}
\label{sec:exp_setup}

\paragraph{\ding{182} Datasets and Metrics.}
We evaluate \methodname{} on interactive decision-making and search-augmented question answering. A separate GPQA experiment examines the contribution of symbolic SID information (Section~\ref{sec:symbolic_experiments}).
\begin{itemize}[nosep,leftmargin=*]
  \item \textbf{ALFWorld}~\citep{shridhar2021alfworld}: a household environment that evaluates sequential decision-making under natural-language instructions across Pick, Look, Clean, Heat, Cool, and Pick2 tasks. For ALFWorld, we report per-subtask and overall success rates (\%).
  \item \textbf{WebShop}~\citep{yao2022webshop}: a simulated e-commerce environment in which agents search for, inspect, and purchase products that satisfy user-specified requirements. For WebShop, we report task score and success rate (\%).
  \item \textbf{Search-augmented QA}: following SkillRL~\citep{xia2026skillrl}, we evaluate on six question-answering benchmarks, \reviewOne{using 2WikiMultiHopQA and HotpotQA as task-model training sources. Bamboogle, MuSiQue, NQ, and TriviaQA are held out from task-model training and contribute seed experiences; see Appendix~\ref{sec:app_dataset_stats}.} For search-augmented QA, we report F1 (\%).
  \item \textbf{Deep Research, Code, and SQL}: we further evaluate whether experience acquired in the source domains transfers to \textbf{out-of-domain tasks} involving long-horizon information seeking over web and document sources, code generation, and SQL query generation; see Appendix~\ref{sec:app_dataset_stats}.
\end{itemize}
Results for ALFWorld and WebShop are in Table~\ref{tab:main_results}, and search-augmented QA results in Table~\ref{tab:qa_results}. Benchmark coverage, comparison settings, and metric definitions are in Appendix~\ref{sec:app_dataset_stats}--\ref{sec:app_metrics}.

\paragraph{\ding{183} Baselines.}
We compare against the methods that appear in the two main result tables:
\begin{itemize}[nosep,leftmargin=*]
  \item \textbf{Closed-source LLMs}: GPT-4o and Gemini-2.5-Pro, representing frontier proprietary capabilities.
  \item \textbf{Prompt-based agentic / memory methods}: ReAct~\citep{yao2023react}, Reflexion~\citep{shinn2023reflexion}, Mem0~\citep{chhikara2025mem0}, MemP~\citep{fang2025memp}, ExpeL~\citep{zhao2024expel},  SimpleMem~\citep{wang2025simplemem}, 
  IRCOT~\citep{trivedi2023interleaving}, and
  TCRAG~\citep{jiang2025tc}.
  \item \textbf{RL-based methods}: RLOO~\citep{ahmadian2024backtobasics} and GRPO~\citep{shao2024deepseekmath} for the embodied and web-task experiments; ReSearch~\citep{chen2025research}, Search-R1~\citep{jin2025searchr1}, AEPO~\citep{dong2025aepo}, ARPO~\citep{dong2025arpo}, MEM1~\citep{zhou2025mem1}, ZeroSearch~\citep{sun2025zerosearch}, and AgenticRAG-R1~\citep{jiang2026agenticragr1} for search-augmented QA.
  \item \textbf{Memory-augmented RL methods}: MemRL~\citep{Zhang2026MemRL},
  EvolveR~\citep{wu2025evolver},
  Mem0+GRPO~\citep{chhikara2025mem0}, 
  SimpleMem+GRPO~\citep{wang2025simplemem}, 
  SkillRL~\citep{xia2026skillrl}, 
  SkillGraph~\citep{li2026skillgraph}, 
  AgentOCR~\citep{feng2026agentocr}, and 
  Skill0~\citep{lu2026skill0}, representing the current state of the art included in our comparison tables.
  \item \textbf{Skill-based Search methods}:
  EvolveR, SkillRL, SkillGraph, AgentOCR, Skill0, 
  Skill-SD~\citep{wang2026skill}, 
  SAPO~\citep{zhang2026co}, and
  SDAR~\citep{lu2026self}
  for search-augmented QA.
\end{itemize}
\paragraph{\ding{184} Implementation and Training.}
StackPlanner, MemRetriever, and MemEvolver all use
\textit{Qwen2.5-7B-Instruct}~\citep{bai2023qwentechnicalreport}
as the backbone LLM.
We construct four SID codebooks using context-aware RQ-KMeans
with sizes $(48, 16, 8, 8)$, yielding $49{,}152$ possible leaf addresses.
The memory modules undergo SID alignment pretraining for $17{,}465$
steps with a batch size of 48, followed by supervised mid-training
on ReAct-style CoT traces for 157 steps with an effective batch size of 32.
Both stages use cross-entropy loss and AdamW with cosine learning-rate
decay, with initial learning rates of $1\times10^{-5}$ and
$2\times10^{-6}$, respectively.
We then jointly optimize MemRetriever and MemEvolver with GRPO
for 130 steps \reviewOne{ with eight rollouts per prompt}.
For StackPlanner, we train the policy model by adopting
Harness-RL~\citep{jiang2026harnessrlblackboxreinforcementlearning}
to train the planning policy with GRPO in a blackbox-Harness manner \reviewOne{ for 200 steps with four rollouts per prompt}.
Memory-module training uses eight NVIDIA A100 80GB GPUs
with DeepSpeed ZeRO-3, and rollouts are generated using vLLM.
Further training details are provided in the appendix, including
oracle-assisted StackPlanner diagnostics in
Appendix~\ref{sec:app_central_rl}.
\subsection{Main Results (RQ1)}
\label{sec:main_results}

\paragraph{\ding{182} Experience Retrieval.}
\Cref{tab:evolution_retrieval} shows that \methodname{} achieves
the highest candidate and final Top-1/Top-5 hit rates.
Under the same candidate budget and BGE reranking protocol,
its final Top-5 hit rate reaches 13.15\%, compared with 10.28\% for SkillRouter,
12.52\% for TF-IDF, and 2.89\% for Qwen3-Embedding-0.6B.
These results highlight the limitations of similarity-based skill retrieval: useful experience must \textbf{match a task's strategic} needs, which textual similarity alone may not capture. \reviewOne{In this comparison, \methodname{}'s generative retriever achieves a higher reference-SID hit rate than SkillRouter.}

\begin{table*}[!t]
\centering
\fontsize{6.5pt}{7.5pt}\selectfont
\captionsetup{font=small,skip=4pt}
\setlength{\tabcolsep}{3.4pt}
\setlength{\aboverulesep}{1.5pt}
\setlength{\belowrulesep}{1.5pt}
\renewcommand{\arraystretch}{1.0}
\caption{\textbf{Main results on ALFWorld and WebShop.} ALFWorld reports per-subtask and overall success rates (\%); WebShop reports task score and success rate (\%).}
\label{tab:main_results}
\begin{tabular*}{\textwidth}{@{\extracolsep{\fill}}l*{9}{c}@{}}
\toprule
\noalign{\begingroup\color{ResultHeader}\hrule height 7.5pt\vskip-7.5pt\endgroup}
& \multicolumn{7}{c}{\textbf{ALFWorld Success Rate (\%)}} & \multicolumn{2}{c}{\textbf{WebShop}} \\
\cmidrule(lr){2-8}\cmidrule(l){9-10}
\textbf{Method} & \textbf{Pick} & \textbf{Look} & \textbf{Clean} & \textbf{Heat} & \textbf{Cool} & \textbf{Pick2} & \textbf{All} & \textbf{Score} & \textbf{Succ.} \\
\midrule
\rowcolor{ResultGroup}
\multicolumn{10}{c}{\textit{Closed-source LLMs}\quad\textnormal{(absolute scores)}} \\
GPT-4o         & 75.3 & 60.8 & 31.2 & 56.7 & 21.6 & 49.8 & 48.0 & 31.8 & 23.7 \\
Gemini-2.5-Pro & 92.8 & 63.3 & 62.1 & 69.0 & 26.6 & 58.7 & 60.3 & 42.5 & 35.9 \\
\midrule
\rowcolor{ResultGroup}
\multicolumn{10}{c}{\textit{Prompt-based Agentic or Memory-based Methods}\quad\textnormal{($\Delta$ vs. ReAct)}} \\
ReAct & \resultcell{48.5}{\samechg{0.0}} & \resultcell{35.4}{\samechg{0.0}} & \resultcell{34.3}{\samechg{0.0}} & \resultcell{13.2}{\samechg{0.0}} & \resultcell{18.2}{\samechg{0.0}} & \resultcell{17.6}{\samechg{0.0}} & \resultcell{31.2}{\samechg{0.0}} & \resultcell{46.2}{\samechg{0.0}} & \resultcell{19.5}{\samechg{0.0}} \\
Reflexion & \resultcell{62.0}{\upchg{+13.5}} & \resultcell{41.6}{\upchg{+6.2}} & \resultcell{44.9}{\upchg{+10.6}} & \resultcell{30.9}{\upchg{+17.7}} & \resultcell{36.3}{\upchg{+18.1}} & \resultcell{23.8}{\upchg{+6.2}} & \resultcell{42.7}{\upchg{+11.5}} & \resultcell{58.1}{\upchg{+11.9}} & \resultcell{28.8}{\upchg{+9.3}} \\
Mem0 & \resultcell{54.0}{\upchg{+5.5}} & \resultcell{55.0}{\upchg{+19.6}} & \resultcell{26.9}{\downchg{-7.4}} & \resultcell{36.4}{\upchg{+23.2}} & \resultcell{20.8}{\upchg{+2.6}} & \resultcell{7.69}{\downchg{-9.9}} & \resultcell{33.6}{\upchg{+2.4}} & \resultcell{23.9}{\downchg{-22.3}} & \resultcell{2.00}{\downchg{-17.5}} \\
MemP & \resultcell{54.3}{\upchg{+5.8}} & \resultcell{38.5}{\upchg{+3.1}} & \resultcell{48.1}{\upchg{+13.8}} & \resultcell{56.2}{\upchg{+43.0}} & \resultcell{32.0}{\upchg{+13.8}} & \resultcell{16.7}{\downchg{-0.9}} & \resultcell{41.4}{\upchg{+10.2}} & \resultcell{25.3}{\downchg{-20.9}} & \resultcell{6.40}{\downchg{-13.1}} \\
ExpeL & \resultcell{21.0}{\downchg{-27.5}} & \resultcell{67.0}{\upchg{+31.6}} & \resultcell{55.0}{\upchg{+20.7}} & \resultcell{52.0}{\upchg{+38.8}} & \resultcell{11.0}{\downchg{-7.2}} & \resultcell{6.00}{\downchg{-11.6}} & \resultcell{46.3}{\upchg{+15.1}} & \resultcell{30.9}{\downchg{-15.3}} & \resultcell{11.2}{\downchg{-8.3}} \\
SimpleMem & \resultcell{64.5}{\upchg{+16.0}} & \resultcell{33.3}{\downchg{-2.1}} & \resultcell{20.0}{\downchg{-14.3}} & \resultcell{12.5}{\downchg{-0.7}} & \resultcell{33.3}{\upchg{+15.1}} & \resultcell{3.84}{\downchg{-13.8}} & \resultcell{29.7}{\downchg{-1.5}} & \resultcell{33.2}{\downchg{-13.0}} & \resultcell{8.59}{\downchg{-10.9}} \\ \midrule
\methodname{} &
\resultcell{91.7}{\upchg{+43.2}} &
\resultcell{83.3}{\upchg{+47.9}} &
\resultcell{71.0}{\upchg{+36.7}} &
\resultcell{56.5}{\upchg{+43.3}} &
\resultcell{85.7}{\upchg{+67.5}} &
\resultcell{58.8}{\upchg{+41.2}} &
\resultcell{74.6}{\upchg{+43.4}} &
\resultcell{52.0}{\upchg{+5.8}} &
\resultcell{27.0}{\upchg{+7.5}} \\
\midrule
\rowcolor{ResultGroup}
\multicolumn{10}{c}{\textit{RL-based Methods}\quad\textnormal{($\Delta$ vs. MemRL)}} \\
RLOO & \resultcell{87.6}{\upchg{+24.8}} & \resultcell{78.2}{\upchg{+39.7}} & \resultcell{87.3}{\upchg{+65.1}} & \resultcell{81.3}{\upchg{+68.8}} & \resultcell{71.9}{\upchg{+63.9}} & \resultcell{48.9}{\upchg{+48.9}} & \resultcell{75.5}{\upchg{+54.1}} & \resultcell{80.3}{\upchg{+50.8}} & \resultcell{65.7}{\upchg{+56.5}} \\
GRPO & \resultcell{90.8}{\upchg{+28.0}} & \resultcell{66.1}{\upchg{+27.6}} & \resultcell{89.3}{\upchg{+67.1}} & \resultcell{74.7}{\upchg{+62.2}} & \resultcell{72.5}{\upchg{+64.5}} & \resultcell{64.7}{\upchg{+64.7}} & \resultcell{77.6}{\upchg{+56.2}} & \resultcell{79.3}{\upchg{+49.8}} & \resultcell{66.1}{\upchg{+56.9}} \\
\midrule
\rowcolor{ResultGroup}
\multicolumn{10}{c}{\textit{Memory-Augmented RL-based Methods}\quad\textnormal{($\Delta$ vs. MemRL)}} \\
MemRL & \resultcell{62.8}{\samechg{0.0}} & \resultcell{38.5}{\samechg{0.0}} & \resultcell{22.2}{\samechg{0.0}} & \resultcell{12.5}{\samechg{0.0}} & \resultcell{8.00}{\samechg{0.0}} & \resultcell{0.00}{\samechg{0.0}} & \resultcell{21.4}{\samechg{0.0}} & \resultcell{29.5}{\samechg{0.0}} & \resultcell{9.2}{\samechg{0.0}} \\
EvolveR & \resultcell{64.9}{\upchg{+2.1}} & \resultcell{33.3}{\downchg{-5.2}} & \resultcell{46.4}{\upchg{+24.2}} & \resultcell{13.3}{\upchg{+0.8}} & \resultcell{33.3}{\upchg{+25.3}} & \resultcell{33.3}{\upchg{+33.3}} & \resultcell{43.8}{\upchg{+22.4}} & \resultcell{42.5}{\upchg{+13.0}} & \resultcell{17.6}{\upchg{+8.4}} \\
Mem0+GRPO & \resultcell{78.1}{\upchg{+15.3}} & \resultcell{54.8}{\upchg{+16.3}} & \resultcell{56.1}{\upchg{+33.9}} & \resultcell{31.0}{\upchg{+18.5}} & \resultcell{65.0}{\upchg{+57.0}} & \resultcell{26.9}{\upchg{+26.9}} & \resultcell{54.7}{\upchg{+33.3}} & \resultcell{58.1}{\upchg{+28.6}} & \resultcell{37.5}{\upchg{+28.3}} \\
SimpleMem+GRPO & \resultcell{89.5}{\upchg{+26.7}} & \resultcell{63.6}{\upchg{+25.1}} & \resultcell{60.0}{\upchg{+37.8}} & \resultcell{50.0}{\upchg{+37.5}} & \resultcell{64.9}{\upchg{+56.9}} & \resultcell{26.3}{\upchg{+26.3}} & \resultcell{62.5}{\upchg{+41.1}} & \resultcell{67.8}{\upchg{+38.3}} & \resultcell{46.9}{\upchg{+37.7}} \\
SkillRL &
\resultcell{\underline{97.9}}{\upchg{+35.1}} &
\resultcell{71.4}{\upchg{+32.9}} & 
\resultcell{\underline{90.0}}{\upchg{+67.8}} &
\resultcell{\textbf{90.0}}{\upchg{+77.5}} &
\resultcell{\textbf{95.5}}{\upchg{+87.5}} &
\resultcell{\underline{87.5}}{\upchg{+87.5}} &
\resultcell{\underline{89.9}}{\upchg{+68.5}} &
\resultcell{85.2}{\upchg{+55.7}} &
\resultcell{\underline{72.7}}{\upchg{+63.5}} \\

AgentOCR &
\resultcell{95.6}{\upchg{+32.8}} &
\resultcell{\underline{96.2}}{\upchg{+57.7}} &
\resultcell{78.1}{\upchg{+55.9}} &
\resultcell{73.2}{\upchg{+60.7}} &
\resultcell{72.4}{\upchg{+64.4}} &
\resultcell{72.0}{\upchg{+72.0}} &
\resultcell{81.2}{\upchg{+59.8}} &
\resultcell{78.6}{\upchg{+49.1}} &
\resultcell{59.3}{\upchg{+50.1}} \\


Skill0 &
\resultcell{\textbf{100.0}}{\upchg{+37.2}} &
\resultcell{85.8}{\upchg{+47.3}} &
\resultcell{94.6}{\upchg{+72.4}} &
\resultcell{81.9}{\upchg{+69.4}} &
\resultcell{85.7}{\upchg{+77.7}} &
\resultcell{80.1}{\upchg{+80.1}} &
\resultcell{89.8}{\upchg{+68.4}} &
\resultcell{\underline{85.3}}{\upchg{+55.8}} &
\resultcell{71.9}{\upchg{+62.7}} \\

\midrule
\textbf{\methodname{} (RL)} &
\resultcell{95.8}{\upchg{+33.0}} &
\resultcell{\textbf{100}}{\upchg{+61.5}} &
\resultcell{\textbf{96.8}}{\upchg{+74.6}} &
\resultcell{\underline{88.7}}{\upchg{+76.2}} &
\resultcell{\underline{90.5}}{\upchg{+82.5}} &
\resultcell{\textbf{88.2}}{\upchg{+88.2}} &
\cellcolor{ResultBest}\resultcell{\textbf{93.6}}{\upchg{+72.2}} &
\resultcell{\textbf{90.8}}{\upchg{+61.3}} &
\resultcell{\textbf{82.8}}{\upchg{+73.6}} \\
\bottomrule
\end{tabular*}
\end{table*}

\begin{table*}[!t]
\centering
\fontsize{6.5pt}{7.5pt}\selectfont
\captionsetup{font=small,skip=4pt}
\setlength{\tabcolsep}{3.4pt}
\setlength{\aboverulesep}{1.5pt}
\setlength{\belowrulesep}{1.5pt}
\renewcommand{\arraystretch}{1.0}
\caption{\textbf{Search-based QA results (Qwen2.5-7B; F1, \%).} \reviewOne{The column groups refer to StackPlanner training. }}
\label{tab:qa_results}
\begin{tabular*}{\textwidth}{@{\extracolsep{\fill}}l*{7}{c}@{}}
\toprule
\noalign{\begingroup\color{ResultHeader}\hrule height 7.5pt\vskip-7.5pt\endgroup}
& \multicolumn{2}{c}{\reviewOne{\textbf{Task-model training sources}}}
& \multicolumn{4}{c}{\reviewOne{\textbf{Held out from task-model training}}}
& \textbf{Overall} \\
\cmidrule(lr){2-3}\cmidrule(lr){4-7}\cmidrule(l){8-8}
\textbf{Method} & \textbf{2Wiki} & \textbf{HotpotQA}
& \textbf{Bamboogle} & \textbf{MuSiQue} & \textbf{NQ} & \textbf{TriviaQA}
& \textbf{Avg.} \\
\midrule
\rowcolor{ResultGroup}
\multicolumn{8}{c}{\textit{Prompt-based Agentic or Memory-based Methods}\quad\textnormal{($\Delta$ vs. ReAct)}} \\
Base & \resultcell{25.41}{\downchg{-2.10}} & \resultcell{26.63}{\downchg{-16.18}} & \resultcell{17.86}{\downchg{-9.77}} & \resultcell{12.15}{\downchg{-7.19}} & \resultcell{19.72}{\downchg{-10.29}} & \resultcell{49.08}{\downchg{-5.47}} & \resultcell{25.1}{\downchg{-8.5}} \\
CoT & \resultcell{23.55}{\downchg{-3.96}} & \resultcell{29.10}{\downchg{-13.71}} & \resultcell{37.56}{\upchg{+9.93}} & \resultcell{14.35}{\downchg{-4.99}} & \resultcell{22.47}{\downchg{-7.54}} & \resultcell{49.33}{\downchg{-5.22}} & \resultcell{29.4}{\downchg{-4.2}} \\
\addlinespace[2pt]
FS-RAG & \resultcell{17.71}{\downchg{-9.80}} & \resultcell{29.21}{\downchg{-13.60}} & \resultcell{16.86}{\downchg{-10.77}} & \resultcell{10.74}{\downchg{-8.60}} & \resultcell{16.82}{\downchg{-13.19}} & \resultcell{35.02}{\downchg{-19.53}} & \resultcell{21.1}{\downchg{-12.5}} \\
FL-RAG & \resultcell{19.78}{\downchg{-7.73}} & \resultcell{34.42}{\downchg{-8.39}} & \resultcell{24.10}{\downchg{-3.53}} & \resultcell{12.46}{\downchg{-6.88}} & \resultcell{19.72}{\downchg{-10.29}} & \resultcell{42.66}{\downchg{-11.89}} & \resultcell{25.5}{\downchg{-8.1}} \\
\addlinespace[2pt]
ReAct & \resultcell{27.51}{\samechg{0.00}} & \resultcell{42.81}{\samechg{0.00}} & \resultcell{27.63}{\samechg{0.00}} & \resultcell{19.34}{\samechg{0.00}} & \resultcell{30.01}{\samechg{0.00}} & \resultcell{54.55}{\samechg{0.00}} & \resultcell{33.6}{\samechg{0.0}} \\
IRCoT & \resultcell{36.45}{\upchg{+8.94}} & \resultcell{26.29}{\downchg{-16.52}} & \resultcell{21.90}{\downchg{-5.73}} & \resultcell{8.39}{\downchg{-10.95}} & \resultcell{19.63}{\downchg{-10.38}} & \resultcell{49.43}{\downchg{-5.12}} & \resultcell{27.0}{\downchg{-6.6}} \\
TCRAG & \resultcell{29.70}{\upchg{+2.19}} & \resultcell{40.83}{\downchg{-1.98}} & \resultcell{25.13}{\downchg{-2.50}} & \resultcell{17.56}{\downchg{-1.78}} & \resultcell{29.01}{\downchg{-1.00}} & \resultcell{54.78}{\upchg{+0.23}} & \resultcell{32.8}{\downchg{-0.8}} \\
\midrule
\methodname{} &
\resultcell{42.91}{\upchg{+15.40}} &
\resultcell{35.83}{\downchg{-6.98}} &
\resultcell{22.21}{\downchg{-5.42}} &
\resultcell{19.85}{\upchg{+0.51}} &
\resultcell{33.88}{\upchg{+3.87}} &
\resultcell{47.62}{\downchg{-6.93}} &
\resultcell{33.7}{\upchg{+0.1}} \\
\midrule
\rowcolor{ResultGroup}
\multicolumn{8}{c}{\textit{RL-based Methods}\quad\textnormal{($\Delta$ vs. ReSearch)}} \\
ReSearch & \resultcell{30.03}{\samechg{0.00}} & \resultcell{30.39}{\samechg{0.00}} & \resultcell{30.42}{\samechg{0.00}} & \resultcell{12.58}{\samechg{0.00}} & \resultcell{23.69}{\samechg{0.00}} & \resultcell{48.25}{\samechg{0.00}} & \resultcell{29.2}{\samechg{0.0}} \\
Search-R1 & \resultcell{35.03}{\upchg{+5.00}} & \resultcell{38.89}{\upchg{+8.50}} & \resultcell{42.04}{\upchg{+11.62}} & \resultcell{19.08}{\upchg{+6.50}} & \resultcell{29.59}{\upchg{+5.90}} & \resultcell{55.91}{\upchg{+7.66}} & \resultcell{36.8}{\upchg{+7.6}} \\
AEPO & \resultcell{19.88}{\downchg{-10.15}} & \resultcell{13.85}{\downchg{-16.54}} & \resultcell{13.24}{\downchg{-17.18}} & \resultcell{5.85}{\downchg{-6.73}} & \resultcell{9.93}{\downchg{-13.76}} & \resultcell{17.53}{\downchg{-30.72}} & \resultcell{13.4}{\downchg{-15.8}} \\
ARPO & \resultcell{30.71}{\upchg{+0.68}} & \resultcell{25.20}{\downchg{-5.19}} & \resultcell{32.94}{\upchg{+2.52}} & \resultcell{12.71}{\upchg{+0.13}} & \resultcell{17.80}{\downchg{-5.89}} & \resultcell{40.16}{\downchg{-8.09}} & \resultcell{26.6}{\downchg{-2.6}} \\
Mem1 & \resultcell{25.29}{\downchg{-4.74}} & \resultcell{29.98}{\downchg{-0.41}} & \resultcell{36.50}{\upchg{+6.08}} & \resultcell{14.13}{\upchg{+1.55}} & \resultcell{26.38}{\upchg{+2.69}} & \resultcell{51.04}{\upchg{+2.79}} & \resultcell{30.6}{\upchg{+1.4}} \\
ZeroSearch & \resultcell{35.2}{\upchg{+5.17}} & \resultcell{34.6}{\upchg{+4.21}} & \resultcell{27.8}{\downchg{-2.62}} & \resultcell{18.4}{\upchg{+5.82}} & \resultcell{43.6}{\upchg{+19.91}} & \resultcell{61.8}{\upchg{+13.55}} & \resultcell{36.9}{\upchg{+7.7}} \\
AgenticRAG-R1 & \resultcell{38.34}{\upchg{+8.31}} & \resultcell{\underline{45.15}}{\upchg{+14.76}} & \resultcell{49.21}{\upchg{+18.79}} & \resultcell{\underline{22.01}}{\upchg{+9.43}} & \resultcell{23.60}{\downchg{-0.09}} & \resultcell{58.45}{\upchg{+10.20}} & \resultcell{39.5}{\upchg{+10.3}} \\
\midrule
\rowcolor{ResultGroup}
\multicolumn{8}{c}{\textit{Memory-Augmented RL-based Methods}\quad\textnormal{($\Delta$ vs. EvolveR)}} \\
EvolveR & \resultcell{42.0}{\samechg{0.0}} & \resultcell{38.2}{\samechg{0.0}} & \resultcell{54.4}{\samechg{0.0}} & \resultcell{15.6}{\samechg{0.0}} & \resultcell{43.5}{\samechg{0.0}} & \resultcell{63.4}{\samechg{0.0}} & \resultcell{42.9}{\samechg{0.0}} \\
SkillRL & \resultcell{40.3}{\downchg{-1.7}} & \resultcell{43.2}{\upchg{+5.0}} & \resultcell{\textbf{73.8}}{\upchg{+19.4}} & \resultcell{20.2}{\upchg{+4.6}} & \resultcell{45.9}{\upchg{+2.4}} & \resultcell{63.3}{\downchg{-0.1}} & \resultcell{47.8}{\upchg{+4.9}} \\
SkillGraph & \resultcell{43.4}{\upchg{+1.4}} & \resultcell{44.7}{\upchg{+6.5}} & \resultcell{72.6}{\upchg{+18.2}} & \resultcell{19.5}{\upchg{+3.9}} & \resultcell{48.0}{\upchg{+4.5}} & \resultcell{63.8}{\upchg{+0.4}} & \resultcell{48.7}{\upchg{+5.8}} \\
AgentOCR & \resultcell{38.3}{\downchg{-3.7}} & \resultcell{40.8}{\upchg{+2.6}} & \resultcell{36.8}{\downchg{-17.6}} & \resultcell{15.7}{\upchg{+0.1}} & \resultcell{43.1}{\downchg{-0.4}} & \resultcell{61.0}{\downchg{-2.4}} & \resultcell{39.3}{\downchg{-3.6}} \\
Skill0 & \resultcell{38.3}{\downchg{-3.7}} & \resultcell{40.0}{\upchg{+1.8}} & \resultcell{66.9}{\upchg{+12.5}} & \resultcell{16.4}{\upchg{+0.8}} & \resultcell{42.7}{\downchg{-0.8}} & \resultcell{61.1}{\downchg{-2.3}} & \resultcell{44.2}{\upchg{+1.3}} \\
Skill-SD & \resultcell{42.1}{\upchg{+0.1}} & \resultcell{44.3}{\upchg{+6.1}} & \resultcell{69.0}{\upchg{+14.6}} & \resultcell{20.2}{\upchg{+4.6}} & \resultcell{47.1}{\upchg{+3.6}} & \resultcell{64.5}{\upchg{+1.1}} & \resultcell{47.9}{\upchg{+5.0}} \\
SAPO & \resultcell{45.2}{\upchg{+3.2}} & \resultcell{45.0}{\upchg{+6.8}} & \resultcell{46.4}{\downchg{-8.0}} & \resultcell{18.3}{\upchg{+2.7}} & \resultcell{\underline{48.4}}{\upchg{+4.9}} & \resultcell{\underline{68.9}}{\upchg{+5.5}} & \resultcell{45.4}{\upchg{+2.5}} \\
SDAR & \resultcell{\underline{48.4}}{\upchg{+6.4}} & \resultcell{43.8}{\upchg{+5.6}} & \resultcell{\underline{73.0}}{\upchg{+18.6}} & \resultcell{19.6}{\upchg{+4.0}} & \resultcell{46.3}{\upchg{+2.8}} & \resultcell{63.5}{\upchg{+0.1}} & \resultcell{\underline{49.1}}{\upchg{+6.2}} \\
\midrule
\textbf{\methodname{} (RL)} & \resultcell{\textbf{54.8}}{\upchg{+12.8}} & \resultcell{\textbf{52.9}}{\upchg{+14.7}} & \resultcell{64.6}{\upchg{+10.2}} & \resultcell{\textbf{23.5}}{\upchg{+7.9}} & \resultcell{\textbf{50.1}}{\upchg{+6.6}} & \resultcell{\textbf{70.7}}{\upchg{+7.3}} & \cellcolor{ResultBest}\resultcell{\textbf{52.8}}{\upchg{+9.9}} \\
\bottomrule
\end{tabular*}
\end{table*}

\begin{table*}[!t]
\centering
\fontsize{5.8pt}{6.8pt}\selectfont
\captionsetup{font=small,skip=4pt}
\setlength{\tabcolsep}{1.5pt}
\setlength{\aboverulesep}{1.5pt}
\setlength{\belowrulesep}{1.5pt}
\renewcommand{\arraystretch}{1.0}
\caption{\textbf{Skill transfer across held-out benchmarks and agent harnesses under DeepSeek-V4-Flash.} Scores and parenthesized changes over the skill-free settings are in \%; \textbf{bold} marks the better result within each harness.}
\label{tab:skill_transfer}
\begin{tabular*}{\textwidth}{@{\extracolsep{\fill}}l*{12}{c}@{}}
\toprule
\noalign{\begingroup\color{ResultHeader}\hrule height 7.5pt\vskip-7.5pt\endgroup}
& \multicolumn{4}{c}{\textbf{StackPlanner}}
& \multicolumn{4}{c}{\textbf{TCRAG}}
& \multicolumn{4}{c}{\textbf{OpenHands}} \\
\cmidrule(lr){2-5}\cmidrule(lr){6-9}\cmidrule(l){10-13}
\textbf{Method}
& \textbf{Search} & \textbf{Research} & \textbf{SQL} & \textbf{Code}
& \textbf{Search} & \textbf{Research} & \textbf{SQL} & \textbf{Code}
& \textbf{Search} & \textbf{Research} & \textbf{SQL} & \textbf{Code} \\
\midrule
Skill-free
& \textbf{58.0} & 49.96 & 78.0 & 74.0
& \textbf{56.0} & 48.88 & 80.0 & 62.0
& 57.0 & 51.48 & 79.0 & 72.0 \\
\rowcolor[HTML]{F0F6FF}
\textbf{+ \methodname{} Skills}
& \resultcell{\textbf{58.0}}{\samechg{0.0}}
& \resultcell{\textbf{52.03}}{\upchg{+2.07}}
& \resultcell{\textbf{83.0}}{\upchg{+5.0}}
& \resultcell{\textbf{75.0}}{\upchg{+1.0}}
& \resultcell{55.0}{\downchg{-1.0}}
& \resultcell{\textbf{50.92}}{\upchg{+2.04}}
& \resultcell{\textbf{81.0}}{\upchg{+1.0}}
& \resultcell{\textbf{63.0}}{\upchg{+1.0}}
& \resultcell{\textbf{58.0}}{\upchg{+1.0}}
& \resultcell{\textbf{53.60}}{\upchg{+2.12}}
& \resultcell{\textbf{81.0}}{\upchg{+2.0}}
& \resultcell{\textbf{74.0}}{\upchg{+2.0}} \\
\bottomrule
\end{tabular*}
\end{table*}


\paragraph{\ding{183} Downstream Task Performance.}
\Cref{tab:main_results,tab:qa_results} show that \methodname{}
improves task performance on ALFWorld, WebShop, and
search-augmented QA both before and after Harness-RL training
of StackPlanner.
Compared with the corresponding memory-free baselines,
\methodname{} yields average gains of 17.0 and 15.3 percentage points before and after training, respectively.
It also outperforms memory-augmented baselines, exceeding SDAR, the second-best method, by 3.7 percentage points on average token F1.
These improvements suggest that the retrieved experience
provides useful support for downstream task execution
and complements the gains from planning-policy optimization.
\textbf{Generalization to unseen domains and harness:}
\Cref{tab:skill_transfer} applies \methodname{} to four benchmarks and three agent harnesses. \textbf{Cross-benchmark transfer.} On unseen datasets from related task families, Research yields consistent gains of 2\%. Search ranges from a 1\% decrease under TCRAG to a 1\% gain under OpenHands Harness~\citep{wang2025openhandsopenplatformai}.

\subsection{Online Self-Evolution (RQ2)}
\label{sec:self_evolution}
We evaluate continual improvement on a held-out query stream disjoint from the initial memory bank,  with model parameters, SID encoder, and codebooks frozen throughout. For each query, the system retrieves, executes, and scores the answer against the reference \emph{before} passing the trajectory to the update policy; the resulting \texttt{insert} or \texttt{revise} is applied before the next query arrives. The reference is used for evaluation only and never enters the update prompt. The stream is partitioned into consecutive blocks; a fixed probe set is evaluated after each block without writing to the bank during probing. \reviewOne{We use the PT-4L MemRetriever checkpoint from Table~\ref{tab:app_full_retrieval}.} Table~\ref{tab:evolution_retrieval} reports retrieval hit rates and downstream task accuracy for four methods as the memory bank evolves over $T$ steps.

\noindent \ding{182} \textit{\textbf{Addressing stability and retrieval quality across evolution.}}
As evolution progresses, \methodname{}'s retrieval hit rates and task accuracy \reviewOne{\emph{improve overall}}. Candidate Hit improves from $20.25\%$ at $t{=}0$ to $25.58\%$ at $t{=}9$, and Task ACC climbs from $30.56\%$ to $36.20\%$, demonstrating that the refined payloads translate into better downstream support. The three content-coupled baselines rebuild their dense-embedding or TF-IDF indices from the latest payloads at each checkpoint and show \reviewOne{\emph{limited gains or declines in task accuracy}. Rebuilding an index after a payload edit can change which entries a query retrieves, offsetting some of the benefit of improved content.}
\reviewThree{\methodname{} achieves these gains while preserving \textbf{addressing invariance} (Challenge~2). Each revised entry retains its SID, so the memory agents can continue to access it at the same address as its payload evolves. A separate control in Appendix~\ref{sec:app_frozen_key} keeps the dense lookup keys fixed while updating the payloads. In that comparison, the content-relevance hit rate falls from $59.0\%$ to $52.0\%$ for the frozen-key baseline and rises from $59.0\%$ to $67.0\%$ for \methodname{}.}

\begin{figure}[t]
\centering   
\includegraphics[width=\linewidth]{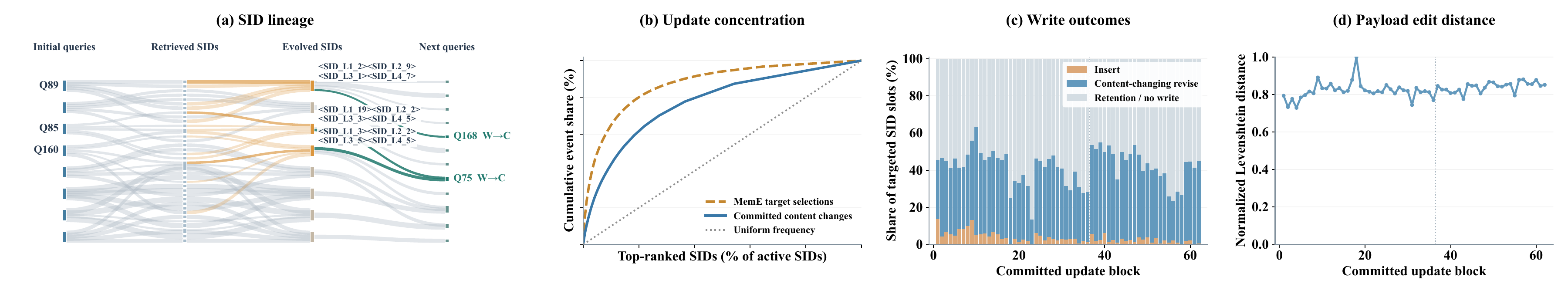}
\caption{Memory bank evolution dynamics over the query stream: SID lineage, SID update concentration, operation-type distribution, and payload edit distance.}
\label{fig:evolution_stats}
\end{figure}

\noindent \ding{183} \textit{\textbf{Evolution dynamics.}}
Figure~\ref{fig:evolution_stats} visualizes how the memory bank changes over the evolution stream: (a) SID lineage traces memory retrieval, revision, and reuse across successive queries, highlighting cross-case evolution: memories revised from one case are subsequently reused to help solve another; (b) update concentration, showing the cumulative share of events against the fraction of active SIDs. Curves closer to the upper-left corner indicate stronger concentration; 
(c) per-block breakdown of insert vs.\ content-changing revise vs.\ retention, showing that MemEvolver writes selectively rather than indiscriminately; and (d) average payload edit distance, quantifying how much existing entries are rewritten. Together with Table~\ref{tab:evolution_retrieval}, these statistics show that substantial content churn occurs throughout evolution---yet only \methodname{} translates this churn into consistent performance gains.
\begin{table}[t]
\centering
\caption{Retrieval hit rates (\%) and downstream task accuracy (\%) across evolution checkpoints. Dense methods rebuild indices at each step; SID addresses are frozen by construction.}
\label{tab:evolution_retrieval}
\scriptsize
\setlength{\tabcolsep}{3.2pt}
\renewcommand{\arraystretch}{0.8}
\begin{tabularx}{\linewidth}{
  @{}
  >{\raggedright\arraybackslash}p{1.6cm}
  >{\raggedright\arraybackslash}p{1.15cm}
  *{10}{Y}
  @{}
}
\toprule
\textbf{Method} & \textbf{Metric} & $t{=}0$ & $t{=}1$ & $t{=}2$ & $t{=}3$ & $t{=}4$ & $t{=}5$ & $t{=}6$ & $t{=}7$ & $t{=}8$ & $t{=}9$ \\
\midrule
\multirow{4}{*}{Qwen3-Emb}
& Cand.\ Hit
& 8.41 & 8.79 & 8.32 & 9.07 & 8.60
& 8.04 & 8.51 & 7.85 & 8.13 & 7.66 \\
& Hit@1
& 0.19 & 0.28 & 0.19 & 0.37 & 0.28
& 0.19 & 0.28 & 0.19 & 0.28 & 0.19 \\
& Hit@5
& 2.89 & 3.17 & 2.80 & 3.36 & 3.08
& 2.71 & 2.99 & 2.61 & 2.80 & 2.52 \\
& Task ACC
& 28.53 & 28.91 & 28.62 & 29.10 & 28.74
& 28.31 & 28.65 & 28.12 & 28.46 & 27.98 \\
\midrule
\multirow{4}{*}{TF-IDF}
& Cand.\ Hit
& 18.47 & 19.03 & 20.06 & 19.31 & 19.78
& 20.34 & 19.59 & 19.97 & 19.41 & 19.69 \\
& Hit@1
& 2.36 & 2.64 & 3.01 & 2.73 & 2.92
& 3.20 & 2.83 & 3.01 & 2.73 & 2.92 \\
& Hit@5
& 12.52 & 12.99 & \textbf{13.74} & 13.18 & 13.55
& \textbf{14.02} & 11.27 & 13.64 & 13.08 & 13.36 \\
& Task ACC
& 30.02 & 30.58 & \textbf{31.46} & 30.91 & 31.18
& \textbf{31.72} & 29.30 & 31.26 & 30.73 & 31.08 \\
\midrule
\multirow{4}{*}{SkillRouter}
& Cand.\ Hit
& 15.79 & 16.26 & 15.98 & 16.54 & 16.17
& 15.61 & 16.08 & 15.33 & 15.70 & 15.14 \\
& Hit@1
& 6.62 & 6.81 & 6.53 & 7.00 & 6.72
& 6.34 & 6.62 & 6.16 & 6.44 & 6.06 \\
& Hit@5
& 10.28 & 10.65 & 10.37 & 10.93 & 10.56
& 10.09 & 12.47 & 9.91 & 10.19 & 9.72 \\
& Task ACC
& 28.74 & 29.12 & 28.93 & 29.48 & 29.21
& 28.86 & 30.84 & 28.79 & 29.07 & 28.95 \\
\midrule
\multirow{4}{*}{\textbf{\methodname{}}}
& Cand.\ Hit
& \textbf{20.25} & \textbf{20.91} & \textbf{20.63}
& \textbf{21.84} & \textbf{22.68} & \textbf{22.21}
& \textbf{23.52} & \textbf{24.36} & \textbf{23.99}
& \textbf{25.58} \\
& Hit@1
& \textbf{7.00} & \textbf{7.37} & \textbf{7.19}
& \textbf{7.93} & \textbf{8.40} & \textbf{8.12}
& \textbf{9.05} & \textbf{9.52} & \textbf{9.24}
& \textbf{10.08} \\
& Hit@5
& \textbf{13.15} & \textbf{13.71} & 13.43
& \textbf{14.55} & \textbf{15.20} & 13.92
& \textbf{16.04} & \textbf{16.79} & \textbf{16.51}
& \textbf{17.63} \\
& Task ACC
& \textbf{30.56} & \textbf{31.10} & 30.88
& \textbf{32.14} & \textbf{32.83} & 31.49
& \textbf{33.68} & \textbf{34.76} & \textbf{34.39}
& \textbf{36.20} \\
\bottomrule
\end{tabularx}
\vspace{-0.2cm}
\end{table}

\section{SIDs as Symbolic Language: Composition\& Emergent Reasoning}
\label{sec:symbolic_experiments}
\label{sec:symbolic_language}
\label{sec:compose_fallback}
A central claim of \methodname{} is that SIDs are
\emph{informative symbolic tokens} that carry task-relevant structure \reviewTwo{even when used without the associated experience text}. We test this on GPQA \citep{rein2023gpqagraduatelevelgoogleproofqa} by varying the depth of SID prefixes supplied without experience text, then comparing complete SIDs with rewritten and directly retrieved experience. Appendix~\ref{sec:app_sid_language_protocol} specifies the input conditions.
Using MemRetriever, we generate $k$ SIDs per query with \reviewTwo{raw beam size} $k\in\{1,5,10\}$ and make two comparisons:
\begin{itemize}[leftmargin=*,noitemsep,topsep=2pt]
    \item \textbf{Symbolic information scaling}: the model receives the raw query alone, then progressively deeper SID prefixes---$L_1$, $L_1$--$L_2$, $L_1$--$L_3$, and the complete $L_1$--$L_4$ address, appended as token sequences. No experience text is provided; model must reason from symbolic addresses alone.
    \item \textbf{Experience representation}: the complete SIDs are used to retrieve stored experience, which is either rewritten into task-adaptive support or supplied directly. Both conditions access the memory bank, testing the contribution of rewriting rather than experience generation from SID tokens alone.
\end{itemize}

\noindent\begin{minipage}{\textwidth}
\begin{minipage}[t]{0.57\linewidth}
\vspace{0pt}
\paragraph{Observations.}
Figure~\ref{fig:symbolic_composition} shows that accuracy increases with SID depth at each beam setting, from $48.32\%$ for the raw query to $54.06$--$54.66\%$ with complete addresses. This gain of $5.74$--$6.34$ percentage points occurs without experience text, supporting the usefulness of symbolic input in this evaluation. Rewritten experience reaches $55.65$--$56.91\%$, while directly retrieved experience reaches $56.09$--$56.63\%$. Rewriting scores higher only at $k=10$, so neither text condition consistently dominates. The additional gains suggest that experience text supplies useful detail beyond the addresses, without establishing generalization to unseen SID combinations.
\end{minipage}\hfill
\hspace{-0.376cm}
\begin{minipage}[t]{0.41\linewidth}
\vspace{0pt}
\centering
\includegraphics[width=\linewidth]{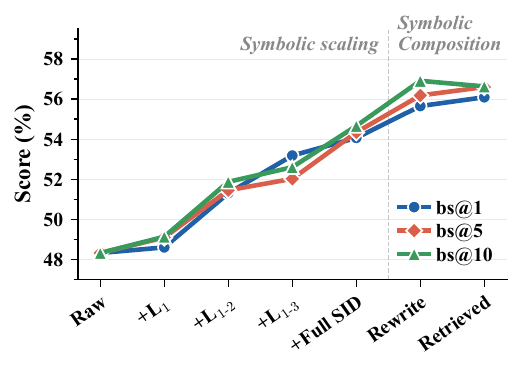}
{\setlength{\abovecaptionskip}{-6pt}%
\captionof{figure}{GPQA accuracy with SID prefixes and experience text. }%
\label{fig:symbolic_composition}%
}
\end{minipage}
\end{minipage}\par


\section{Conclusion, Limitations and Future Work}
\methodname{} demonstrates that reformulating memory as generative symbolic addressing \reviewThree{can address experience sparsity and the invariance--evolution tension}. SID generation over a hierarchical address space keeps addresses stable while content is refined, enabling long-horizon self-evolution. Experiments across embodied planning, web navigation, and multi-hop QA evaluate this design against strong memory-augmented agents, and we further observe that \reviewTwo{combining multiple reranked SIDs improves answer accuracy over using a single SID, even without the associated experience text}.

\textbf{Limitations and Future Directions.}
\reviewTwo{Three} practical limitations stand out. \ding{182} Some leaf codes attract disproportionate traffic, creating uneven utilization across the codebook. \ding{183} Cold-start noise in early trajectories can propagate into the SID space. \reviewTwo{\ding{184} It remains unclear how well the system generalizes to new skills and task domains when the required distinctions exceed the effective capacity of the learned SID space.} \reviewTwo{These limitations motivate \emph{adaptive codebook allocation} to add addresses for new skills while preserving existing SIDs. A background revisit policy could refine experiences derived from noisy early trajectories and those stored in under-exercised subtrees.} A second direction concerns \emph{SID--text-token interleaving}. Because SIDs share the decoder substrate with natural-language tokens, the compositional behaviors observed in Section~\ref{sec:symbolic_language} suggest that SIDs could serve as first-class orchestration tokens in long, multi-tool workflows, extending their role as lookup indices and turning the memory bank into a programmable symbolic layer for complex task planning.

%

\clearpage
\subsection*{AI use statement}
Large language models were used for language polishing, literature discovery, synthetic trajectory construction, figure-design assistance, and AI-assisted programming. LLMs helped generate candidate agent trajectories from public benchmark tasks, compare successful and failed executions, and distill reusable strategies and failure-recovery lessons into the experience memory; they also provided visual drafts, code suggestions, and debugging support. All AI-assisted trajectories, distilled experiences, figures, citations, and code were reviewed and verified by the authors, with benchmark verifiers, environment feedback, automated tests, and human inspection used as appropriate to validate generated artifacts and the qualitative cases reported in the paper. The research questions, methodology, experimental design, analyses, and conclusions were determined by the authors.

\subsection*{Ethics statement}
All experiments use publicly available benchmarks, such as ALFWorld, WebShop, GPQA, BIRD, SWE-bench Verified, and DeepResearch Bench, under their respective licenses and terms. The study involves no human or animal subjects and uses no private user conversations. Agent trajectories and reusable experiences are generated from benchmark tasks using author-designed protocols and LLM assistance rather than collected from real users. Interactive and software-engineering tasks are executed in benchmark environments or isolated workspaces to limit unintended effects on external systems.

\subsection*{Reproducibility statement}
The appendix reports dataset versions and splits, memory-construction statistics, prompts, model and decoding configurations, tool and token budgets, training settings, baseline implementations, and evaluation procedures. The method and objectives are described in Sections~\ref{sec:method}--\ref{sec:memory_optimization}; the general experimental protocol is documented in Appendix~\ref{sec:app_experiments}; the cross-domain memory split and overlap checks are reported in Appendix; and the complete prompts are provided in Appendix~\ref{sec:app_full_prompts}.




\bibliographystyle{iclr2027_conference}
\bibliography{example_paper}

@misc{wang2025openhandsopenplatformai,
      title={OpenHands: An Open Platform for AI Software Developers as Generalist Agents}, 
      author={Xingyao Wang and Boxuan Li and Yufan Song and Frank F. Xu and Xiangru Tang and Mingchen Zhuge and Jiayi Pan and Yueqi Song and Bowen Li and Jaskirat Singh and Hoang H. Tran and Fuqiang Li and Ren Ma and Mingzhang Zheng and Bill Qian and Yanjun Shao and Niklas Muennighoff and Yizhe Zhang and Binyuan Hui and Junyang Lin and Robert Brennan and Hao Peng and Heng Ji and Graham Neubig},
      year={2025},
      eprint={2407.16741},
      archivePrefix={arXiv},
      primaryClass={cs.SE},
      url={https://arxiv.org/abs/2407.16741}, 
}

@misc{jiang2026harnessrlblackboxreinforcementlearning,
      title={Harness-RL: Black-Box Reinforcement Learning with Action-Args Decoupling for Central-Agent Multi-Agent Harnesses}, 
      author={Xinke Jiang and Zhixin Zhang and Zhibang Yang and Jiaran Gao and Rihong Qiu and Shijin Chen and Xu Chu and Junfeng Zhao and Yasha Wang},
      year={2026},
      eprint={2608.29641},
      archivePrefix={arXiv},
      primaryClass={cs.MA},
      url={https://arxiv.org/abs/2608.29641}, 
}

@misc{rein2023gpqagraduatelevelgoogleproofqa,
      title={GPQA: A Graduate-Level Google-Proof Q\&A Benchmark}, 
      author={David Rein and Betty Li Hou and Asa Cooper Stickland and Jackson Petty and Richard Yuanzhe Pang and Julien Dirani and Julian Michael and Samuel R. Bowman},
      year={2023},
      eprint={2311.12022},
      archivePrefix={arXiv},
      primaryClass={cs.AI},
      url={https://arxiv.org/abs/2311.12022}, 
}

@misc{stackplanner,
      title={StackPlanner: A Centralized Hierarchical Multi-Agent System with Task-Experience Memory Management}, 
      author={Ruizhe Zhang and Xinke Jiang and Zhibang Yang and Zhixin Zhang and Jiaran Gao and Yuzhen Xiao and Tao Feng and Yue Fang and Yuxuan Liu and Ruiqing Li and Hongbin Lai and Huheng Huang and Xu Chu and Junfeng Zhao and Yasha Wang},
      year={2026},
      eprint={2601.05890},
      archivePrefix={arXiv},
      primaryClass={cs.AI},
      url={https://arxiv.org/abs/2601.05890}, 
}

@article{Wang2023Voyager,
  title={Voyager: An Open-Ended Embodied Agent with Large Language Models},
  author={Wang, Guanzhi and Xie, Yuqi and Jiang, Yunfan and Mandlekar, Ajay and Xiao, Chaowei and Zhu, Yuke and Fan, Linxi and Anandkumar, Anima},
  journal={arXiv preprint arXiv:2305.16291},
  year={2023},
  eprint={2305.16291},
  archivePrefix={arXiv},
  primaryClass={cs.AI}
}

@article{Hong2023MetaGPT,
  title={{MetaGPT}: Meta Programming for A Multi-Agent Collaborative Framework},
  author={Hong, Sirui and Zhuge, Mingchen and Chen, Jiaqi and Zheng, Xiawu and Cheng, Yuheng and Zhang, Ceyao and Wang, Jinlin and Wang, Zili and Yau, Steven Ka Shing and Lin, Zijuan and Zhou, Liyang and Ran, Chenyu and Xiao, Lingfeng and Wu, Chenglin and Schmidhuber, J{\"u}rgen},
  journal={arXiv preprint arXiv:2308.00352},
  year={2023},
  eprint={2308.00352},
  archivePrefix={arXiv},
  primaryClass={cs.AI}
}

@article{wang2026skill,
  title={Skill-sd: Skill-conditioned self-distillation for multi-turn llm agents},
  author={Wang, Hao and Wang, Guozhi and Xiao, Han and Zhou, Yufeng and Pan, Yue and Wang, Jichao and Xu, Ke and Wen, Yafei and Ruan, Xiaohu and Chen, Xiaoxin and others},
  journal={arXiv preprint arXiv:2604.10674},
  year={2026}
}

@article{lu2026self,
  title={Self-distilled agentic reinforcement learning},
  author={Lu, Zhengxi and Yao, Zhiyuan and Han, Zhuowen and Wang, Zi-Han and Wu, Jinyang and Gu, Qi and Cai, Xunliang and Lu, Weiming and Xiao, Jun and Zhuang, Yueting and others},
  journal={arXiv preprint arXiv:2605.15155},
  year={2026}
}

@article{zhang2026co,
  title={Co-Evolving Skill Generation and Policy Optimization},
  author={Zhang, Zhiwei and Lin, Yudi and Kuang, Nikki Lijing and Wu, Linlin and Li, Xiaomin and Liu, Songtao and Ma, Fenglong},
  journal={arXiv preprint arXiv:2606.08755},
  year={2026}
}

@article{Zhang2026MemRL,
  title={MemRL: Self-Evolving Agents via Runtime Reinforcement Learning on Episodic Memory},
  author={Zhang, Shengtao and Wang, Jiaqian and Zhou, Ruiwen and Liao, Junwei and Feng, Yuchen and Li, Zhuo and Zheng, Yujie and Zhang, Weinan and Wen, Ying and Li, Zhiyu and Xiong, Feiyu and Qi, Yutao and Tang, Bo and Wen, Muning},
  journal={arXiv preprint arXiv:2601.03192},
  year={2026},
  eprint={2601.03192},
  archivePrefix={arXiv},
  primaryClass={cs.CL}
}

@article{Zhang2024MemorySurvey,
  title={A Survey on the Memory Mechanism of Large Language Model based Agents},
  author={Zhang, Zeyu and Bo, Xiaohe and Ma, Chen and Li, Rui and Chen, Xu and Dai, Quanyu and Zhu, Jieming and Dong, Zhenhua and Wen, Ji-Rong},
  journal={arXiv preprint arXiv:2404.13501},
  year={2024},
  eprint={2404.13501},
  archivePrefix={arXiv},
  primaryClass={cs.AI}
}

@article{Ma2026FineMem,
  title={Fine-Mem: Fine-Grained Feedback Alignment for Long-Horizon Memory Management},
  author={Ma, Weitao and Feng, Xiaocheng and Huang, Lei and Feng, Xiachong and Ma, Zhanyu and Xu, Jun and Gao, Jiuchong and Hao, Jinghua and He, Renqing and Qin, Bing},
  journal={arXiv preprint arXiv:2601.08435},
  year={2026},
  eprint={2601.08435},
  archivePrefix={arXiv},
  primaryClass={cs.CL}
}

@article{hu2022lora,
  title={{LoRA}: Low-Rank Adaptation of Large Language Models},
  author={Hu, Edward J. and Shen, Yelong and Wallis, Phillip and Allen-Zhu, Zeyuan and Li, Yuanzhi and Wang, Shean and Wang, Lu and Chen, Weizhu},
  journal={arXiv preprint arXiv:2106.09685},
  year={2022},
  eprint={2106.09685},
  archivePrefix={arXiv},
  primaryClass={cs.CL}
}

@article{fedus2022switch,
  title={Switch Transformers: Scaling to Trillion Parameter Models with Simple and Efficient Sparsity},
  author={Fedus, William and Zoph, Barret and Shazeer, Noam},
  journal={Journal of Machine Learning Research},
  volume={23},
  number={120},
  pages={1--39},
  year={2022}
}

@article{shao2024deepseekmath,
  title={{DeepSeek-Math}: Pushing the Limits of Mathematical Reasoning in Open Language Models},
  author={Shao, Zhihong and Wang, Peiyi and Zhu, Qihao and Xu, Runxin and Song, Junxiao and Bi, Xiao and Zhang, Haowei and Zhang, Mingchuan and Li, Y.K. and Wu, Y. and Guo, Daya},
  journal={arXiv preprint arXiv:2402.03300},
  year={2024},
  eprint={2402.03300},
  archivePrefix={arXiv},
  primaryClass={cs.CL}
}

@inproceedings{rajput2023recommender,
  title={Recommender Systems with Generative Retrieval},
  author={Rajput, Shashank and Mehta, Nikhil and Singh, Anima and Keshavan, Raghunandan H. and Vu, Trung and Heldt, Lukasz and Hong, Lichan and Tay, Yi and Tran, Vinh Q. and Samost, Jonah and Kula, Maciej and Chi, Ed H. and Sathiamoorthy, Maheswaran},
  booktitle={Advances in Neural Information Processing Systems (NeurIPS)},
  year={2023}, eprint={2305.05065}, archivePrefix={arXiv}
}

@article{jin2025onerec,
  title={{OneRec}: Unifying Retrieve and Rank with Generative Recommender and Iterative Preference Alignment},
  author={Deng, Jiaxin and Wang, Shiyao and Cai, Kuo and Ren, Lejian and Hu, Qigen and Ding, Weifeng and Luo, Qiang and Zhou, Guorui},
  journal={arXiv preprint arXiv:2502.18965}, year={2025}, eprint={2502.18965}, archivePrefix={arXiv}, primaryClass={cs.IR}
}

@misc{trivedi2023interleaving,
      title={Interleaving Retrieval with Chain-of-Thought Reasoning for Knowledge-Intensive Multi-Step Questions}, 
      author={Harsh Trivedi and Niranjan Balasubramanian and Tushar Khot and Ashish Sabharwal},
      year={2023},
      eprint={2212.10509},
      archivePrefix={arXiv},
      primaryClass={cs.CL}
}

@article{jiang2025tc,
  title={TC-RAG: Turing-Complete RAG's Case study on Medical LLM Systems},
  author={Jiang , Xinke and Fang , Yue and Qiu , Rihong and Zhang, Haoyu and Xu, Yongxin and Chen, Hao and Zhang, Wentao and Zhang, Ruizhe and Fang, Yuchen and Chu, Xu and others},
  journal={ACL oral 2025},
  year={2025}
}

@inproceedings{clip,
  title={Learning transferable visual models from natural language supervision},
  author={Radford, Alec and Kim, Jong Wook and Hallacy, Chris and Ramesh, Aditya and Goh, Gabriel and Agarwal, Sandhini and Sastry, Girish and Askell, Amanda and Mishkin, Pamela and Clark, Jack and others},
  booktitle={International conference on machine learning},
  pages={8748--8763},
  year={2021},
  organization={PMLR}
}

@misc{bai2023qwentechnicalreport,
      title={Qwen Technical Report}, 
      author={Jinze Bai and Shuai Bai and Yunfei Chu and Zeyu Cui and Kai Dang and Xiaodong Deng and Yang Fan and Wenbin Ge and Yu Han and Fei Huang and Binyuan Hui and Luo Ji and Mei Li and Junyang Lin and Runji Lin and Dayiheng Liu and Gao Liu and Chengqiang Lu and Keming Lu and Jianxin Ma and Rui Men and Xingzhang Ren and Xuancheng Ren and Chuanqi Tan and Sinan Tan and Jianhong Tu and Peng Wang and Shijie Wang and Wei Wang and Shengguang Wu and Benfeng Xu and Jin Xu and An Yang and Hao Yang and Jian Yang and Shusheng Yang and Yang Yao and Bowen Yu and Hongyi Yuan and Zheng Yuan and Jianwei Zhang and Xingxuan Zhang and Yichang Zhang and Zhenru Zhang and Chang Zhou and Jingren Zhou and Xiaohuan Zhou and Tianhang Zhu},
      year={2023},
      eprint={2309.16609},
      archivePrefix={arXiv},
      primaryClass={cs.CL},
      url={https://arxiv.org/abs/2309.16609}, 
}

@article{zhang2025memgen,
  title={Memgen: Weaving generative latent memory for self-evolving agents},
  author={Zhang, Guibin and Fu, Muxin and Yan, Shuicheng},
  journal={arXiv preprint arXiv:2509.24704},
  year={2025}
}

@article{chhikara2025mem0,
  title={Mem0: Building production-ready ai agents with scalable long-term memory},
  author={Chhikara, Prateek and Khant, Dev and Aryan, Saket and Singh, Taranjeet and Yadav, Deshraj},
  journal={arXiv preprint arXiv:2504.19413},
  year={2025}
}

@misc{zhang2026memoryactionautonomouscontext,
      title={Memory as Action: Autonomous Context Curation for Long-Horizon Agentic Tasks}, 
      author={Yuxiang Zhang and Jiangming Shu and Ye Ma and Xueyuan Lin and Shangxi Wu and Jitao Sang},
      year={2026},
      eprint={2510.12635},
      archivePrefix={arXiv},
      primaryClass={cs.AI},
      url={https://arxiv.org/abs/2510.12635}, 
}

@misc{xu2025amemagenticmemoryllm,
      title={A-MEM: Agentic Memory for LLM Agents}, 
      author={Wujiang Xu and Zujie Liang and Kai Mei and Hang Gao and Juntao Tan and Yongfeng Zhang},
      year={2025},
      eprint={2502.12110},
      archivePrefix={arXiv},
      primaryClass={cs.CL},
      url={https://arxiv.org/abs/2502.12110}, 
}

@inproceedings{tay2022transformer,
  title={Transformer Memory as a Differentiable Search Index},
  author={Tay, Yi and Tran, Vinh Q. and Dehghani, Mostafa and Ni, Jianmo and Bahri, Dara and Mehta, Harsh and Metzler, Donald and others},
  booktitle={Advances in Neural Information Processing Systems (NeurIPS)},
  year={2022},
  eprint={2202.06991},
  archivePrefix={arXiv}
}

@inproceedings{decao2021autoregressive,
  title={Autoregressive Entity Retrieval},
  author={de Cao, Nicola and Izacard, Gautier and Riedel, Sebastian and Petroni, Fabio},
  booktitle={Proceedings of the 2021 Conference of the North American Chapter of the Association for Computational Linguistics (NAACL)},
  year={2021},
  eprint={2010.00904},
  archivePrefix={arXiv}
}

@inproceedings{oord2017neural,
  title={Neural Discrete Representation Learning},
  author={van den Oord, Aaron and Vinyals, Oriol and Kavukcuoglu, Koray},
  booktitle={Advances in Neural Information Processing Systems (NeurIPS)},
  year={2017},
  eprint={1711.00937},
  archivePrefix={arXiv}
}

@inproceedings{lee2022autoregressive,
  title={Autoregressive Image Generation using Residual Quantization},
  author={Lee, Doyup and Kim, Chiheon and Kim, Saehoon and Cho, Minsu and Han, Wook-Shin},
  booktitle={Proceedings of the IEEE/CVF Conference on Computer Vision and Pattern Recognition (CVPR)},
  year={2022},
  eprint={2203.01941},
  archivePrefix={arXiv}
}

@article{sumers2024cognitive,
  title={Cognitive Architectures for Language Agents},
  author={Sumers, Theodore R. and Yao, Shunyu and Narasimhan, Karthik and Griffiths, Thomas L.},
  journal={Transactions on Machine Learning Research},
  year={2024},
  eprint={2309.02427},
  archivePrefix={arXiv}
}

@inproceedings{shinn2023reflexion,
  title={Reflexion: Language Agents with Verbal Reinforcement Learning},
  author={Shinn, Noah and Cassano, Federico and Labash, Beck and Gopinath, Ashwin and Narasimhan, Karthik and Yao, Shunyu},
  booktitle={Advances in Neural Information Processing Systems (NeurIPS)},
  year={2023},
  eprint={2303.11366},
  archivePrefix={arXiv}
}

@inproceedings{zhao2024expel,
  title={{ExpeL}: {LLM} Agents Are Experiential Learners},
  author={Zhao, Andrew and Huang, Daniel and Xu, Quentin and Lin, Matthieu and Liu, Yong-Jin and Huang, Gao},
  booktitle={Proceedings of the AAAI Conference on Artificial Intelligence (AAAI)},
  year={2024}, eprint={2308.10144}, archivePrefix={arXiv}
}

@article{jin2025searchr1,
  title={Search-R1: Training {LLMs} to Reason and Leverage Search Engines with Reinforcement Learning},
  author={Jin, Bowen and Zeng, Hansi and Yue, Zhenrui and Yoon, Jinsung and Arik, Sercan and Wang, Dong and Zamani, Hamed and Han, Jiawei},
  journal={arXiv preprint arXiv:2503.09516}, year={2025}, eprint={2503.09516}, archivePrefix={arXiv}, primaryClass={cs.CL}
}

@article{ahmadian2024backtobasics,
  title={Back to Basics: Revisiting {REINFORCE} Style Optimization for Learning from Human Feedback in {LLMs}},
  author={Ahmadian, Arash and Cremer, Chris and Gall{\'e}, Matthias and Fadaee, Marzieh and Kreutzer, Julia and Pietquin, Olivier and Ustun, Ahmet and Hooker, Sara},
  journal={arXiv preprint arXiv:2402.14740}, year={2024}, eprint={2402.14740}, archivePrefix={arXiv}, primaryClass={cs.LG}
}

@article{chen2025research,
  title={ReSearch: Learning to Reason with Search for {LLMs} via Reinforcement Learning},
  author={Chen, Mingyang and Sun, Linzhuang and Li, Tianpeng and Sun, Haoze and Zhou, Yijie and Zhu, Chenzheng and Wang, Haofen and Pan, Jeff Z. and Zhang, Wen and Chen, Huajun and Yang, Fan and Zhou, Zenan and Chen, Weipeng},
  journal={arXiv preprint arXiv:2503.19470}, year={2025}, eprint={2503.19470}, archivePrefix={arXiv}, primaryClass={cs.AI}
}

@article{dong2025aepo,
  title={Agentic Entropy-Balanced Policy Optimization},
  author={Dong, Guanting and Bao, Licheng and Wang, Zhongyuan and Zhao, Kangzhi and Li, Xiaoxi and Jin, Jiajie and Yang, Jinghan and Mao, Hangyu and Zhang, Fuzheng and Gai, Kun and Zhou, Guorui and Zhu, Yutao and Wen, Ji-Rong and Dou, Zhicheng},
  journal={arXiv preprint arXiv:2510.14545}, year={2025}, eprint={2510.14545}, archivePrefix={arXiv}, primaryClass={cs.AI}
}

@article{dong2025arpo,
  title={Agentic Reinforced Policy Optimization},
  author={Dong, Guanting and Mao, Hangyu and Ma, Kai and Bao, Licheng and Chen, Yifei and Wang, Zhongyuan and Chen, Zhongxia and Du, Jiazhen and Wang, Huiyang and Zhou, Fuzheng and Zhou, Guorui and Zhu, Yutao and Wen, Ji-Rong and Dou, Zhicheng},
  journal={arXiv preprint arXiv:2507.19849}, year={2025}, eprint={2507.19849}, archivePrefix={arXiv}, primaryClass={cs.AI}
}

@article{zhou2025mem1,
  title={{MEM1}: Learning to Synergize Memory and Reasoning for Efficient Long-Horizon Agents},
  author={Zhou, Zijian and Qu, Ao and Wu, Zhaoxuan and Kim, Sunghwan and Prakash, Alok and Rus, Daniela and Zhao, Jinhua and Low, Bryan Kian Hsiang and Liang, Paul Pu},
  journal={arXiv preprint arXiv:2506.15841}, year={2025}, eprint={2506.15841}, archivePrefix={arXiv}, primaryClass={cs.AI}
}

@article{sun2025zerosearch,
  title={ZeroSearch: Incentivize the Search Capability of {LLMs} without Searching},
  author={Sun, Hao and Qiao, Zile and Guo, Jiayan and Fan, Xuanbo and Hou, Yingyan and Jiang, Yong and Xie, Pengjun and Huang, Fei and Zhang, Yan},
  journal={arXiv preprint arXiv:2505.04588}, year={2025}, eprint={2505.04588}, archivePrefix={arXiv}, primaryClass={cs.CL}
}

@article{jiang2026agenticragr1,
  title={AgenticRag-R1: Agentic Reinforcement Learning with Stack Memory for Multi-Step Reasoning, Retrieval and Memorizing},
  author={Jiang, Xinke and Fang, Yue and Yang, Zhibang and Gao, Jiaran and Zhang, Zhixin and Feng, Tao and Qiu, Rihong and Zhang, Wentao and Ding, Hongxin and Zhang, Ruizhe and Xu, Yongxin and Huang, Yuheng and Chu, Xu and Zhao, Junfeng and Wang, Yasha},
  journal={arXiv preprint arXiv:2608.29622}, year={2026}, eprint={2608.29622}, archivePrefix={arXiv}, primaryClass={cs.MA}
}

@article{xia2026skillrl,
  title={{SkillRL}: Evolving Agents via Recursive Skill-Augmented Reinforcement Learning},
  author={Xia, Peng and Chen, Jianwen and Wang, Hanyang and Liu, Jiaqi and Zeng, Kaide and Wang, Yu and Han, Siwei and Zhou, Yiyang and Zhao, Xujiang and Chen, Haifeng and Zheng, Zeyu and Xie, Cihang and Yao, Huaxiu},
  journal={arXiv preprint arXiv:2602.08234}, year={2026}, eprint={2602.08234}, archivePrefix={arXiv}, primaryClass={cs.LG}
}

@article{li2026skillgraph,
  title={{SkillGraph}: Skill-Augmented Reinforcement Learning for Agents via Evolving Skill Graphs},
  author={Li, Xiaoyuan and Li, Moxin and Bao, Keqin and Ma, Yubo and Wang, Wenjie and Liu, Dayiheng and Feng, Fuli},
  journal={arXiv preprint arXiv:2605.12039}, year={2026}, eprint={2605.12039}, archivePrefix={arXiv}, primaryClass={cs.CL}
}

@article{wu2025evolver,
  title={{EvolveR}: Self-Evolving {LLM} Agents through an Experience-Driven Lifecycle},
  author={Wu, Rong and Wang, Xiaoman and Mei, Jianbiao and Cai, Pinlong and Fu, Daocheng and Yang, Cheng and Wen, Licheng and Yang, Xuemeng and Shen, Yufan and Wang, Yuxin and Shi, Botian},
  journal={arXiv preprint arXiv:2510.16079}, year={2025}, eprint={2510.16079}, archivePrefix={arXiv}, primaryClass={cs.CL}, note={Accepted by ICML 2026}
}

@article{wang2025simplemem,
  title={{SimpleMem}: Efficient Lifelong Memory for {LLM} Agents},
  author={Liu, Jiaqi and Su, Yaofeng and Xia, Peng and Han, Siwei and Zheng, Zeyu and Xie, Cihang and Ding, Mingyu and Yao, Huaxiu},
  journal={arXiv preprint arXiv:2601.02553}, year={2026}, eprint={2601.02553}, archivePrefix={arXiv}, primaryClass={cs.AI}
}

@article{fang2025memp,
  title={{Memp}: Exploring Agent Procedural Memory},
  author={Fang, Runnan and Liang, Yuan and Wang, Xiaobin and Wu, Jialong and Qiao, Shuofei and Xie, Pengjun and Huang, Fei and Chen, Huajun and Zhang, Ningyu},
  journal={arXiv preprint arXiv:2508.06433}, year={2025}, eprint={2508.06433}, archivePrefix={arXiv}, primaryClass={cs.CL}, note={ACL 2026 Findings}
}

@inproceedings{shridhar2021alfworld,
  title={{ALFWorld}: Aligning Text and Embodied Environments for Interactive Learning},
  author={Shridhar, Mohit and Yuan, Xingdi and C{\^o}t{\'e}, Marc-Alexandre and Bisk, Yonatan and Trischler, Adam and Hausknecht, Matthew},
  booktitle={International Conference on Learning Representations (ICLR)},
  year={2021}, eprint={2010.03768}, archivePrefix={arXiv}
}

@inproceedings{yao2022webshop,
  title={{WebShop}: Towards Scalable Real-World Web Interaction with Grounded Language Agents},
  author={Yao, Shunyu and Chen, Howard and Yang, John and Narasimhan, Karthik},
  booktitle={Advances in Neural Information Processing Systems (NeurIPS)},
  year={2022}, eprint={2207.01206}, archivePrefix={arXiv}
}

@inproceedings{yao2023react,
  title={{ReAct}: Synergizing Reasoning and Acting in Language Models},
  author={Yao, Shunyu and Zhao, Jeffrey and Yu, Dian and Du, Nan and Shafran, Izhak and Narasimhan, Karthik and Cao, Yuan},
  booktitle={International Conference on Learning Representations (ICLR)},
  year={2023}, eprint={2210.03629}, archivePrefix={arXiv}
}

@article{cheng2026engram,
  title     = {Conditional Memory via Scalable Lookup: A New Axis of Sparsity for Large Language Models},
  author    = {Xin Cheng and Rui Tian and Wangding Zeng and Damai Dai and Qinyu Chen and Bingxuan Wang and Zhenda Xie and Kezhao Huang and Xingkai Yu and Chengqi Deng and Shangyan Zhou and Chenggang Zhao and Zhewen Hao and Yukun Li and Han Zhang and Zhengyan Zhang and Yixu Wei and M.Y. Xu and Huishuai Zhang and Dongyan Zhao and Wenfeng Liang},
  journal   = {arXiv preprint arXiv:2601.07372},
  year      = {2026},
  url       = {https://arxiv.org/abs/2601.07372}
}

@article{ni2026trace2skill,
  title={Trace2Skill: Distill Trajectory-Local Lessons into Transferable Agent Skills},
  author={Ni, Jingwei and Liu, Yihao and Liu, Xinpeng and Sun, Yutao and Zhou, Mengyu and Cheng, Pengyu and Wang, Dexin and Zhao, Erchao and Jiang, Xiaoxi and Jiang, Guanjun},
  journal={arXiv preprint arXiv:2603.25158},
  year={2026}
}

@article{zheng2026skillrouter,
  title={{SkillRouter}: Skill Routing for {LLM} Agents at Scale},
  author={Zheng, YanZhao and Zhang, ZhenTao and Ma, Chao and Yu, YuanQiang and Zhu, JiHuai and Wu, Yong and Xu, Tianze and Dong, Baohua and Zhu, Hangcheng and Huang, Ruohui and Yu, Gang},
  journal={arXiv preprint arXiv:2603.22455},
  year={2026}
}

@inproceedings{wang2026skillx,
  title={{SkillX}: Automatically Constructing Skill Knowledge Bases for Agents},
  author={Wang, Chenxi and Yu, Zhuoyun and Xie, Xin and Yao, Wuguannan and Fang, Runnan and Qiao, Shuofei and Cao, Kexin and Zheng, Guozhou and Qi, Xiang and Zhang, Peng and Deng, Shumin},
  booktitle={Proceedings of the 43rd International Conference on Machine Learning (ICML)},
  year={2026}
}

@article{lu2026skill0,
  title={{SKILL0}: In-Context Agentic Reinforcement Learning for Skill Internalization},
  author={Lu, Zhengxi and Yao, Zhiyuan and Wu, Jinyang and Han, Chengcheng and Gu, Qi and Cai, Xunliang and Lu, Weiming and Xiao, Jun and Zhuang, Yueting and Shen, Yongliang},
  journal={arXiv preprint arXiv:2604.02268},
  year={2026}
}

@inproceedings{feng2026agentocr,
  title={Agentocr: Reimagining agent history via optical self-compression},
  author={Feng, Lang and Yang, Fuchao and Chen, Feng and Cheng, Xin and Xu, Haiyang and Wan, Zhenglin and Yan, Ming and An, Bo},
  booktitle={Proceedings of the 64th Annual Meeting of the Association for Computational Linguistics (Volume 1: Long Papers)},
  pages={5067--5086},
  year={2026}
}

@inproceedings{jimenez2024swe,
  title={Swe-bench: Can language models resolve real-world github issues?},
  author={Jimenez, Carlos E and Yang, John and Wettig, Alexander and Yao, Shunyu and Pei, Kexin and Press, Ofir and Narasimhan, Karthik},
  booktitle={International Conference on Learning Representations},
  volume={2024},
  pages={54107--54157},
  year={2024}
}

@inproceedings{du2026deepresearch,
  title={Deepresearch bench: A comprehensive benchmark for deep research agents},
  author={Du, Mingxuan and Xu, Benfeng and Zhu, Chiwei and Zhang, Licheng and Wang, Xiaorui and Mao, Zhendong},
  booktitle={International Conference on Learning Representations},
  volume={2026},
  pages={42414--42448},
  year={2026}
}

@article{li2023can,
  title={Can llm already serve as a database interface? a big bench for large-scale database grounded text-to-sqls},
  author={Li, Jinyang and Hui, Binyuan and Qu, Ge and Yang, Jiaxi and Li, Binhua and Li, Bowen and Wang, Bailin and Qin, Bowen and Geng, Ruiying and Huo, Nan and others},
  journal={Advances in Neural Information Processing Systems},
  volume={36},
  pages={42330--42357},
  year={2023}
}

@article{chen2025browsecomp,
  title={Browsecomp-plus: A more fair and transparent evaluation benchmark of deep-research agent},
  author={Chen, Zijian and Ma, Xueguang and Zhuang, Shengyao and Nie, Ping and Zou, Kai and Liu, Andrew and Green, Joshua and Patel, Kshama and Meng, Ruoxi and Su, Mingyi and others},
  journal={arXiv preprint arXiv:2508.06600},
  year={2025}
}

\clearpage
\appendix
\section{Related Work}

\paragraph{Memory-Augmented LLM Agents.}
Memory-augmented LLM agents accumulate and reuse experience across tasks to support long-horizon reasoning and continual adaptation~\citep{Zhang2024MemorySurvey,sumers2024cognitive}.
Existing work explores both how experience is organized and how memory is managed.
Reflexion~\citep{shinn2023reflexion} and ExpeL~\citep{zhao2024expel} distill past interactions into verbal reflections and reusable insights without gradient updates, while SimpleMem~\citep{wang2025simplemem} and MemP~\citep{fang2025memp} curate compact memory representations.
Mem0~\citep{chhikara2025mem0} and A-MEM~\citep{xu2025amemagenticmemoryllm} further investigate structured memory management through graph-based organization or associative linking.
At the skill level, SkillRL~\citep{xia2026skillrl} and SkillGraph~\citep{li2026skillgraph} organize reusable knowledge into hierarchical or graph-structured banks.
Trace2Skill~\citep{ni2026trace2skill} consolidates execution trajectories into portable skill directories, while SkillX~\citep{wang2026skillx} constructs three-tiered hierarchies of strategic plans, functional skills, and atomic operations.
Beyond memory organization, MemRL~\citep{Zhang2026MemRL}, Memory-as-Action~\citep{zhang2026memoryactionautonomouscontext}, and MemGen~\citep{zhang2025memgen} introduce learned mechanisms for memory utilization, control, or generation.
SKILL0~\citep{lu2026skill0} takes a complementary approach by internalizing skills into model parameters.
Despite these advances, \textbf{richer memory organization does not by itself provide a model with a structured address space that it can learn to navigate: similarity-based access remains sensitive to how experiences are represented, and semantic resemblance need not reflect their strategic utility in the current context.}
SkillRouter~\citep{zheng2026skillrouter} further highlights the sensitivity of skill selection to the information exposed during retrieval.
\textbf{Learning which memories to access and refine also remains challenging when supervision is derived primarily from delayed task outcomes, which provide limited credit assignment to individual memory operations.} 

\paragraph{Generative Retrieval and Discrete Representations.}
Generative retrieval formulates information access as autoregressive identifier generation, allowing a model to directly predict the target of retrieval.
DSI~\citep{tay2022transformer} generates document identifiers from queries, while GENRE~\citep{decao2021autoregressive} retrieves entities by generating their names.
TIGER~\citep{rajput2023recommender} extends identifier generation to recommendation using semantic IDs constructed through residual quantization.
These methods build on discrete representation learning: VQ-VAE~\citep{oord2017neural} maps continuous representations to learned codebook entries, and RQ-VAE~\citep{lee2022autoregressive} progressively quantizes residual information into sequences of discrete codes.
Such representations provide compact identifiers with coarse-to-fine semantic structure, enabling autoregressive access through shared code prefixes.
\textbf{However, generating identifiers for indexed content does not by itself resolve the requirements of an evolving experience bank, where entries must be selectively inserted, revised, consolidated, or removed.}
\textbf{When semantic identifiers depend on content representations, revisions can create tension between preserving learned addresses and maintaining their semantic coherence. Existing retrieval formulations therefore leave open how to jointly support stable context-to-address mappings and targeted content updates through a shared generative interface.}


\section{Formal Foundations and SID Representation}
\label{sec:app_prelim}

This appendix collects formal definitions and analysis underlying the design choices in the main text.

\subsection{The Sparsity Structure of Experience Space}
\label{sec:app_sparsity}

A central insight of this work is that experience space is \emph{inherently sparse}---a property that existing dense memory systems fail to exploit. We identify four distinct sparsity dimensions that collectively motivate generative symbolic addressing.

\paragraph{Value Sparsity.}
Let $\mathcal{T}_q = \{\tau^{(1)}, \dots, \tau^{(K)}\}$ denote a set of $K$ trajectories generated for query $q$. Define the \emph{experience value} of a trajectory as its contribution to the generalizable experience base:
\[
v(\tau^{(k)}) = \mathbb{E}_{q' \sim \mathcal{D}}\bigl[\Delta r(q',\, \mathcal{M} \cup \mathcal{E}(\tau^{(k)})) - \Delta r(q',\, \mathcal{M})\bigr],
\]
where $\mathcal{E}(\tau)$ denotes experience distilled from $\tau$ and $\Delta r$ denotes task performance improvement. In practice, $v(\tau^{(k)})$ is high only for a small fraction of trajectories---those that exhibit novel, generalizable strategies or capture failure patterns not previously recorded.

\paragraph{Structural Sparsity.}
Even among high-value experiences, the experience space exhibits strong structural regularity. Formally, let $\{e_i\}_{i=1}^N$ be the semantic embeddings of all experience entries. These embeddings lie approximately on a low-dimensional manifold:
\[
\dim\bigl(\mathrm{span}\{e_i\}_{i=1}^N\bigr) \;\ll\; N,
\]
because many experiences share latent structural components---task categories, sub-task types, tool-use patterns---even when their surface-level descriptions differ. This \emph{structural sparsity} means that the effective address space of experience is far smaller than its raw cardinality. This is precisely the principle behind RQ-KMeans-based SID construction: the multi-level codebook exploits structural sparsity to represent $N_1 \times N_2 \times \cdots \times N_L$ distinct experience regions using only $N_1 + N_2 + \cdots + N_L$ codebook entries.

\paragraph{Access Sparsity.}
Semantic similarity does not by itself determine whether a memory will help an agent solve a task. Let $c=(q,h_t,\text{env})$ denote the execution context, comprising the query, the agent's working state, and the environmental state. We define the \emph{execution utility} of a memory $m_i$ as $u(c,m_i)=\mathbb{E}_{\tau\sim\pi(\cdot\mid c,m_i)}[r(\tau,q)]$, the expected task reward when the execution policy $\pi$ uses that memory. Access sparsity is the assumption that only a small subset of the memory bank exceeds a given utility threshold for any particular context. Formally, $|\mathcal{R}^*(c)|\ll|\mathcal{M}|$, where $\mathcal{R}^*(c)=\{m_i\in\mathcal{M}:u(c,m_i)>\epsilon\}$ and $\epsilon>0$.

\paragraph{Evolution sparsity (update).}
Only a small subset of experiences---those repeatedly exercised by frequent queries---warrant refinement; blanket updates waste compute and blur useful entries. Effective memory systems must therefore learn selective evolution, deciding not only \emph{how} to update an entry but also \emph{whether} to leave it untouched through a content-preserving \texttt{revise} operation (\S\ref{sec:memory_operations}).

These four sparsity dimensions together motivate designing experience memory around \emph{sparse, structured, generative addressing} rather than dense global retrieval---realized concretely through the SID framework (\S\ref{sec:method}).

\subsection{Retrieval Utility and Address Stability}
\label{sec:app_disc_limits}

\paragraph{Retrieval utility.}
A similarity score measures agreement between representations; it does not by itself specify the value of a memory for task execution. For example, two experiences may describe similar tasks but prescribe different actions. Ranking memories by $\mathrm{sim}(f_q(q),f_m(x_i))$ therefore need not produce the same ordering as ranking them by the execution utility $u(c,m_i)$ defined in Appendix~\ref{sec:app_sparsity}. This is a distinction between objectives, not an inherent limitation of discriminative retrieval. A scoring model can also condition on the execution context and be trained with task feedback. In \methodname{}, process and outcome rewards train the memory policies to select and rewrite experience for downstream use (Section~\ref{sec:posttraining}); symbolic addressing alone does not ensure useful retrieval.

\paragraph{Address stability.}
If a retrieval score depends on an embedding of the current payload, revising that payload and recomputing its embedding can change its rank for the same query, even when its role is unchanged. \methodname{} separates the address from the payload: a revision changes $(x_i,z_i)$ to $(x_i',z_i)$, preserving the SID used for lookup and update targeting (Section~\ref{sec:memory_operations}). With the addressing policy and codebooks fixed, a payload revision does not by itself change the SID generated for an unchanged input context. It can still change how useful the retrieved content is. This design does not imply that vector-based stores cannot support stable identifiers or targeted updates; GenMem uses a shared, generatable SID interface for both reading and writing memory.

\subsection{SID as a Generative Token Sequence}

An SID $\mathbf{z}=(s^1,\ldots,s^L)$ consists of $L$ level-specific tokens, each identifying an entry in the corresponding codebook. Token identities are distinct across levels, even when their numerical indices coincide. The four-level configuration in Section~\ref{sec:sid_construction} therefore represents each address with four tokens, from \texttt{<SID\_L1\_X>} to \texttt{<SID\_L4\_W>}.

Conditioned on context $\mathbf{c}$, the addressing policy generates the sequence autoregressively,
\begin{equation}
    p_{\theta}(\mathbf{z}\mid\mathbf{c})
    =\prod_{\ell=1}^{L}
      p_{\theta}(s^{\ell}\mid\mathbf{c},s^1,\ldots,s^{\ell-1}).
    \label{eq:app_sid_generation}
\end{equation}
MemRetriever conditions address generation on the query and available reasoning context, whereas MemEvolver uses the execution trajectory to select an update address. In both cases, the generated SID identifies a memory slot whose payload is accessed by lookup. Offline quantization constructs the initial addresses; the learned policy subsequently generates addresses for retrieval or revision. Summary generation and payload revision are separate operations performed after lookup.

\begin{figure}[ht]
\centering
\includegraphics[width=0.95\textwidth]{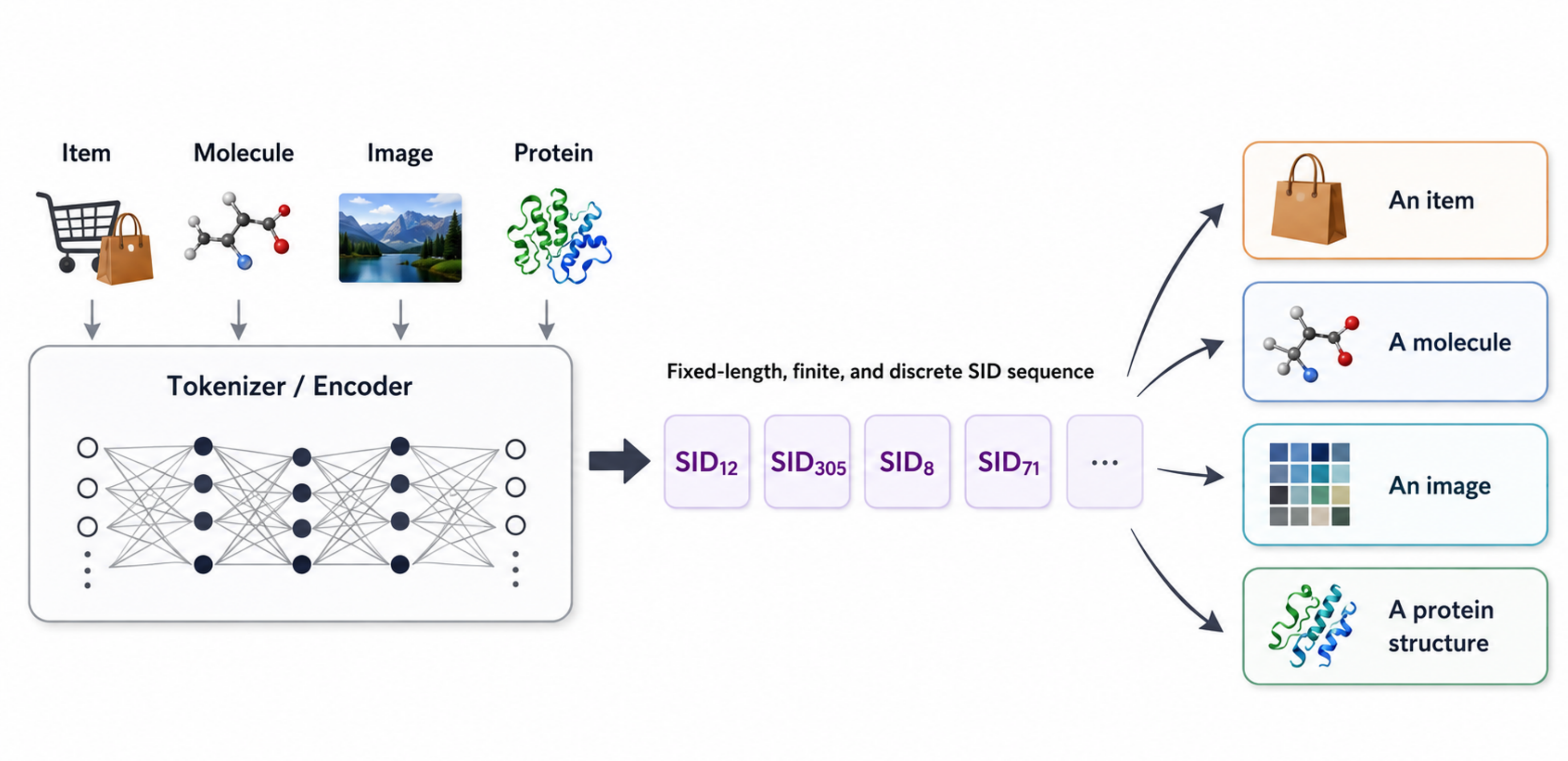}
\caption{Discrete token sequences as object representations. In \methodname{}, a text encoder and RQ-KMeans map seed experiences to SID addresses (Section~\ref{sec:sid_construction}). Other object types illustrate the general concept and are not evaluated in this work.}
\label{fig:sid_concept}
\end{figure}

Figure~\ref{fig:sid_concept} illustrates discrete sequence representations at a conceptual level. \methodname{} applies this principle to textual experiences, using the sequence as an address rather than a lossless encoding of the associated text.

\subsection{Embedding Reconstruction from SIDs}
\label{sec:app_rqkmeans_context}

The SID $\mathbf{z}_i=(s_i^1,\ldots,s_i^L)$ constructed in Section~\ref{sec:sid_construction} selects one centroid from each codebook. Their sum approximates the original experience embedding $\mathbf{e}_i$; it does not reconstruct the experience text. Using the residual recurrence in Eq.~\ref{eq:rq_recurrence}, the reconstruction after $\ell$ levels is
\begin{equation}
    \hat{\mathbf{e}}_i^{(\ell)}
    =\sum_{h=1}^{\ell}\mathbf{c}_{s_i^h}^{(h)},
    \qquad
    \mathbf{r}_i^{(\ell+1)}
    =\mathbf{e}_i-\hat{\mathbf{e}}_i^{(\ell)},
    \quad \ell=1,\ldots,L.
    \label{eq:app_partial_reconstruction}
\end{equation}
Writing $\hat{\mathbf{e}}_i=\hat{\mathbf{e}}_i^{(L)}$, the per-entry Euclidean reconstruction error is
\begin{equation}
    \varepsilon_i
    =\bigl\|\mathbf{e}_i-\hat{\mathbf{e}}_i\bigr\|_2
    =\bigl\|\mathbf{r}_i^{(L+1)}\bigr\|_2.
    \label{eq:app_reconstruction}
\end{equation}
Each successive codebook is fitted to the residuals left by the preceding levels. The Euclidean fitting objective minimizes mean squared residual error over the fitting set, but nearest-centroid assignment alone does not guarantee a decrease for every entry at every level. In particular, the selected centroid need not be closer to an incoming residual than the zero vector is. We therefore do not assume per-entry monotonic error reduction. Equation~\ref{eq:app_reconstruction} defines an unnormalized Euclidean norm; its relationship to the reported reconstruction loss depends on the aggregation and normalization settings noted in Appendix~\ref{sec:app_rqkmeans}.

Reconstruction characterizes the offline quantization of seed embeddings, not online memory evolution. MemEvolver generates the target SID for an update, and the experience text at that address can subsequently change without re-encoding it or changing the SID (Appendix~\ref{sec:app_online_updates}).

\subsection{SID Alignment Mappings}
\label{sec:app_sid_mapping}

SID alignment pretraining (Section~\ref{sec:pretraining}) connects queries $q$, execution trajectories $\tau$, and experience texts $\mathbf{x}$ to the constructed address space. Each target SID is a complete sequence $\mathbf{z}=(s^1,\ldots,s^L)$. The five mappings comprise three address-generation tasks and two text-generation tasks:
\begin{enumerate}[leftmargin=*,noitemsep,topsep=2pt]
    \item \textbf{Query $\rightarrow$ SID} ($q\mapsto\mathbf{z}$). Predict the SID of the experience associated with the query in the training pair.
    \item \textbf{Trajectory $\rightarrow$ SID} ($\tau\mapsto\mathbf{z}$). Predict the SID associated with the experience distilled from the trajectory. This task learns an address association, not an \texttt{insert} or \texttt{revise} decision.
    \item \textbf{Experience $\rightarrow$ SID} ($\mathbf{x}\mapsto\mathbf{z}$). Predict the SID assigned to an experience by offline construction.
    \item \textbf{SID $\rightarrow$ Experience} ($\mathbf{z}\mapsto\mathbf{x}$). Generate the experience text associated with the SID in the training pair.
    \item \textbf{SID $\rightarrow$ Description} ($\mathbf{z}\mapsto\mathbf{d}$). Generate a natural-language semantic description $\mathbf{d}$ associated with the SID.
\end{enumerate}
The reverse mappings train text generation conditioned on an address. They do not invoke \texttt{memory\_lookup}; tool-mediated retrieval and memory evolution are taught during memory-operation mid-training (Section~\ref{sec:midtraining}).

\paragraph{SID compositionality.}
Shared SID prefixes identify common sequences of selected centroids, while later symbols encode the remaining residuals (Appendix~\ref{sec:app_rqkmeans}). The alignment tasks associate these structured addresses with task contexts and experience content. Neither the codebook construction nor the alignment objective assigns a fixed semantic role, such as task domain or tool type, to an individual level. Whether the learned SID structure supports reasoning is examined empirically in Section~\ref{sec:symbolic_experiments}, rather than assumed from the mapping tasks alone.

\section{Additional Method Details}
\label{sec:app_method_details}

\subsection{RQ-KMeans Details}
\label{sec:app_rqkmeans}

\reviewThree{This section describes the distance settings and residual
updates used in SID construction
(Section~\ref{sec:sid_construction}).}

\paragraph{Prefix-guided encoding.}
Let $N_{\mathrm{seed}}=|\mathcal{X}_{\mathrm{seed}}|$ denote the number of experience summaries before SID-based consolidation. We encode each summary as $\mathbf{e}_i=f_{\mathrm{enc}}(p_{\mathrm{enc}}\oplus\mathbf{x}_i)\in\mathbb{R}^d$, where $f_{\mathrm{enc}}$ is frozen, $d$ is the embedding dimension, and $\oplus$ denotes text concatenation. \reviewThree{In the main configuration, $f_{\mathrm{enc}}$ returns a 1,024-dimensional, $\ell_2$-normalized embedding.} The instruction prefix $p_{\mathrm{enc}}$ is reproduced in Appendix~\ref{sec:app_embedding_prompt}; its \texttt{Query} field is filled with the extracted \texttt{rich\_summary}. It emphasizes reusable procedures, applicability conditions, verification, and failure recovery. Thus, context enters through the experience summary, rather than through a separate concatenation of raw queries and trajectories.

\paragraph{Codebook fitting and assignment.}
At level $\ell\in\{1,\ldots,L\}$, the codebook $\mathcal{C}^{(\ell)}=\{\mathbf{c}_j^{(\ell)}\}_{j=1}^{N_\ell}$ contains $N_\ell$ centroids in $\mathbb{R}^d$. Starting with $\mathbf{r}_i^{(1)}=\mathbf{e}_i$, the Euclidean formulation fits each codebook to the incoming residuals using the K-Means objective
\begin{equation}
    \mathcal{L}_{\ell}(\mathcal{C}^{(\ell)})
    =\frac{1}{N_{\mathrm{seed}}}
      \sum_{i=1}^{N_{\mathrm{seed}}}\min_{j\in[N_\ell]}
      \bigl\|\mathbf{r}_i^{(\ell)}-\mathbf{c}_j^{(\ell)}\bigr\|_2^2,
    \label{eq:kmeans_obj}
\end{equation}
where $[N_\ell]=\{1,\ldots,N_\ell\}$. The fitted centroids determine a code and the residual passed to the next level,
\begin{equation}
\begin{aligned}
    s_i^\ell &= \operatorname*{arg\,min}_{j\in[N_\ell]}
        \bigl\|\mathbf{r}_i^{(\ell)}-\mathbf{c}_j^{(\ell)}\bigr\|_2^2,\\
    \mathbf{r}_i^{(\ell+1)}
        &=\mathbf{r}_i^{(\ell)}-\mathbf{c}_{s_i^\ell}^{(\ell)}.
\end{aligned}
\label{eq:rq_recurrence}
\end{equation}
The resulting SID is $\mathbf{z}_i=(s_i^1,\ldots,s_i^L)$. Codebooks are fitted sequentially and then held fixed during online operation. \reviewThree{Equations~\ref{eq:kmeans_obj}--\ref{eq:rq_recurrence} define the Euclidean baseline. Table~\ref{tab:app_distance_summary} compares it with the Spherical and Weighted ($w=0.5$) settings.}

{\color{ReviewThree}%
\paragraph{Weighted distance.}
The weighted distance at level $\ell$ combines squared Euclidean distance
and cosine distance between residual $\mathbf{r}$ and candidate centroid
$\mathbf{c}$,
\begin{equation}
    d_{\mathrm{weighted}}^{(\ell)}(\mathbf{r},\mathbf{c})
    = (1-w_\ell)\lVert\mathbf{r}-\mathbf{c}\rVert_2^2
      + w_\ell\left(1-
        \frac{\mathbf{r}^{\top}\mathbf{c}}
        {\lVert\mathbf{r}\rVert_2\lVert\mathbf{c}\rVert_2}\right).
    \label{eq:rq_weighted_distance}
\end{equation}
For the $48\times16\times8\times8$ configuration, we set $w_\ell=0.5$
at all four levels. At each level we select the centroid that minimizes
Eq.~\ref{eq:rq_weighted_distance} and subtract it to obtain the next
residual, as in Eq.~\ref{eq:rq_assignment}. The initial residual
$\mathbf{r}_i^{(1)}=\mathbf{e}_i$ has unit norm. Later residuals retain
the norms produced by the subtraction.
}

\paragraph{Capacity and utilization.}
\label{sec:app_rqkmeans_definitions}
An $L$-level configuration has address capacity $K_{\mathrm{addr}}$ and level-specific SID vocabulary size $V_{\mathrm{SID}}$ given by
\begin{equation}
    K_{\mathrm{addr}}=\prod_{\ell=1}^{L}N_\ell,
    \qquad
    V_{\mathrm{SID}}=\sum_{\ell=1}^{L}N_\ell.
    \label{eq:app_sid_capacity}
\end{equation}
Capacity counts possible addresses, not populated memory entries. For $N_{\mathrm{eval}}$ encoded summaries, let $\mathcal{P}_\ell=\{(s_i^1,\ldots,s_i^\ell):1\leq i\leq N_{\mathrm{eval}}\}$ be the set of observed prefixes. Single-code utilization and cumulative prefix utilization are
\begin{equation}
    U_\ell=\frac{|\{s_i^\ell:1\leq i\leq N_{\mathrm{eval}}\}|}{N_\ell},
    \qquad
    U_{1:\ell}=\frac{|\mathcal{P}_\ell|}{\prod_{h=1}^{\ell}N_h}.
    \label{eq:app_sid_utilization}
\end{equation}
The used-leaf count is $|\mathcal{P}_L|$, and leaf utilization is $U_{1:L}$; utilization values are reported as percentages. The evaluation set need not be the full fitting set, so its size must be specified separately from $N_{\mathrm{seed}}$. Per-entry Euclidean reconstruction error is defined in Appendix~\ref{sec:app_rqkmeans_context}.

\paragraph{Unique-code ratio.}
For a non-empty evaluation set, let $n_{\mathbf{z}}=\sum_{i=1}^{N_{\mathrm{eval}}}\mathbb{I}[\mathbf{z}_i=\mathbf{z}]$ count the summaries assigned to a complete SID $\mathbf{z}$. We use \emph{unique-code ratio} (UCR) to denote the number of distinct complete SIDs per encoded summary,
\begin{equation}
    \mathrm{UCR}
    =\frac{|\mathcal{P}_L|}{N_{\mathrm{eval}}}
    =\frac{\sum_{\mathbf{z}\in\mathcal{Z}}\mathbb{I}[n_{\mathbf{z}}>0]}{N_{\mathrm{eval}}},
    \qquad
    \mathcal{Z}=\prod_{\ell=1}^{L}[N_\ell].
    \label{eq:app_sid_ucr}
\end{equation}
UCR characterizes address sharing across experiences. Unlike leaf utilization, its denominator is the sample count rather than address capacity: $\mathrm{UCR}=(K_{\mathrm{addr}}/N_{\mathrm{eval}})U_{1:L}$. It is not the fraction of summaries assigned to singleton leaves, which is $N_{\mathrm{eval}}^{-1}\sum_{\mathbf{z}}\mathbb{I}[n_{\mathbf{z}}=1]$.

\paragraph{Assignment balance.}
Utilization measures whether codes are used, not how evenly assignments are distributed. For level $\ell$, define the empirical code probabilities and entropy by
\begin{equation}
    p_{\ell j}=\frac{1}{N_{\mathrm{eval}}}\sum_{i=1}^{N_{\mathrm{eval}}}\mathbb{I}[s_i^\ell=j],
    \qquad
    H_\ell=-\sum_{j=1}^{N_\ell}p_{\ell j}\log p_{\ell j},
    \label{eq:app_sid_entropy}
\end{equation}
with $0\log 0=0$ and natural logarithms. The normalized entropy and effective code count reported in Figure~\ref{fig:codebook_dist} are
\begin{equation}
    H_\ell^{\mathrm{norm}}=\frac{H_\ell}{\log N_\ell},
    \qquad
    N_\ell^{\mathrm{eff}}=\exp(H_\ell).
    \label{eq:app_sid_effective_codes}
\end{equation}
The normalization assumes $N_\ell>1$ and accounts for differences in codebook size. The effective code count expresses assignment entropy as an equivalent number of equally frequent codes.

\paragraph{Distribution uniformity index (DUI).}
An entropy-based distribution uniformity index measures assignment balance by averaging normalized entropy across quantization levels,
\begin{equation}
    \mathrm{DUI}_{\mathrm{entropy}}
    =\frac{1}{L}\sum_{\ell=1}^{L}H_\ell^{\mathrm{norm}}
    =\frac{1}{L}\sum_{\ell=1}^{L}
      \frac{-\sum_{j=1}^{N_\ell}p_{\ell j}\log p_{\ell j}}{\log N_\ell}.
    \label{eq:app_sid_dui_entropy}
\end{equation}
Each level contributes equally, irrespective of its codebook size. This definition measures marginal assignment balance, not statistical independence or semantic orthogonality between SID levels.

\paragraph{Inter-level code independence (ICR).}
To quantify statistical independence between SID levels, let $p(\mathbf{z})=n_{\mathbf{z}}/N_{\mathrm{eval}}$ be the empirical joint distribution of complete SIDs. Its joint entropy and total correlation are
\begin{equation}
    H_{\mathrm{joint}}=-\sum_{\mathbf{z}\in\mathcal{Z}}p(\mathbf{z})\log p(\mathbf{z}),
    \qquad
    T=\sum_{\ell=1}^{L}H_\ell-H_{\mathrm{joint}}.
    \label{eq:app_sid_total_correlation}
\end{equation}
Total correlation measures the discrepancy between the joint distribution and the product of its marginals. We propose an entropy-based independence ratio,
\begin{equation}
    \mathrm{ICR}_{\mathrm{entropy}}
    =\frac{H_{\mathrm{joint}}}{\sum_{\ell=1}^{L}H_\ell}
    =1-\frac{T}{\sum_{\ell=1}^{L}H_\ell},
    \qquad \sum_{\ell=1}^{L}H_\ell>0.
    \label{eq:app_sid_icr_entropy}
\end{equation}
Higher values indicate less cross-level dependence relative to the total marginal entropy. If all marginal entropies vanish, we leave the ratio undefined and report the collapsed assignment separately. This statistic concerns dependence among code assignments, not semantic orthogonality. It is computed from the empirical joint distribution and can be sensitive to sparse sampling of the address space. The entropy-based ratio is a proposed diagnostic, distinct from the original ICR scores reported in the configuration comparison.

\paragraph{Offline quantization cost.}
\label{sec:app_rqkmeans_complexity}
For the Euclidean formulation in Appendix~\ref{sec:app_rqkmeans}, one assignment pass at level $\ell$ compares each of the $N_{\mathrm{seed}}$ residuals with $N_\ell$ centroids in $\mathbb{R}^d$. With $T_{\mathrm{KM}}$ K-Means iterations per level, the total assignment cost during fitting is
\begin{equation}
    C_{\mathrm{fit,assign}}
    =O\!\left(N_{\mathrm{seed}}dT_{\mathrm{KM}}
       \sum_{\ell=1}^{L}N_\ell\right).
    \label{eq:app_quantization_fit_cost}
\end{equation}
Once the codebooks are fitted, assigning a SID to one already computed embedding costs
\begin{equation}
    C_{\mathrm{quantize}}
    =O\!\left(d\sum_{\ell=1}^{L}N_\ell\right).
    \label{eq:app_quantization_assignment_cost}
\end{equation}
The codebooks store $d\sum_{\ell=1}^{L}N_\ell$ scalar coordinates. These counts exclude text encoding, experience extraction, and LLM decoding; they are not end-to-end latency estimates. In particular, Eq.~\ref{eq:app_quantization_assignment_cost} does not describe online SID generation, which is performed by the learned memory policy.

\subsection{Online Memory Self-Evolution}
\label{sec:app_online_updates}

RQ-KMeans assigns addresses to the seed experiences during offline construction. During online memory self-evolution, MemEvolver generates the target SID for both \texttt{insert} and \texttt{revise}. The updated payload is not re-encoded and quantized to determine its update address.

\paragraph{Memory updates.}
Let $\mathcal{M}_t$ be the memory bank before online update $t$, and $\tau_t$ the newly completed execution trajectory. With model parameters and codebooks fixed, MemEvolver generates an update address, consults the associated entry, and returns an update tuple. Using the interface in Section~\ref{sec:memory_operations}, we write
\begin{equation}
    (o_t,\mathbf{z}_{e,t},\tilde{\mathbf{x}}_{e,t})
    =\textsc{\textbf{MemEvolver}}(\tau_t,\mathcal{M}_t),
    \qquad
    o_t\in\{\texttt{insert},\texttt{revise}\}.
    \label{eq:app_online_proposal}
\end{equation}
Here $o_t$ is the operation, $\mathbf{z}_{e,t}\in\mathcal{Z}$ is the target SID, and $\tilde{\mathbf{x}}_{e,t}$ is the updated payload. These are the quantities $(o,\mathbf{z}_e,\tilde{\mathbf{x}}_e)$ defined in the main text, indexed by update step $t$. An \texttt{insert} populates an unoccupied SID slot; a \texttt{revise} updates an existing entry. Neither operation reassigns the SID of an existing entry.

\paragraph{Addressing invariance.}
Write $\mathcal{M}_t[\mathbf{z}]$ for the payload at address $\mathbf{z}$, using $\varnothing$ when no experience is stored. Applying the update changes only the addressed payload,
\begin{equation}
    \mathcal{M}_{t+1}[\mathbf{z}]
    =\begin{cases}
        \tilde{\mathbf{x}}_{e,t}, & \mathbf{z}=\mathbf{z}_{e,t},\\
        \mathcal{M}_t[\mathbf{z}], & \mathbf{z}\neq\mathbf{z}_{e,t}.
    \end{cases}
    \label{eq:app_online_payload_update}
\end{equation}
Revision supports retention, consolidation, and deletion through unchanged, merged, or empty payloads, respectively. These are outcomes of \texttt{revise}, not additional operation labels. Offline consolidation of seed experiences sharing an SID uses the SID-leaf synthesis prompt (Appendix~\ref{sec:app_sid_synthesis_prompts}); online consolidation is handled by MemEvolver's revision at the target SID. Addressing invariance means that content changes do not reassign existing SIDs. It does not guarantee that the revised content remains useful for every query that previously retrieved it.

{\color{ReviewThree}%
\subsection{Reward Design Details}
\label{sec:app_reward}
Let $m\in\{0,1,2,3,4\}$ count the levels that match the reference SID and let $s,u\in[0,1]$ denote normalized judge scores for relevance and output utility. The process terms score the intermediate memory operations. The outcome term scores the downstream answer given the experience supplied by the memory policy.

\subsubsection{MemRetriever}
\label{sec:app_reward_process}
For an eligible retrieval, the SID reward is $R_{\mathrm{SID}}^{\mathrm{R}}=0.05+0.0125m$. SID validity, lookup success, and output format gate the associated reward terms. A Qwen3-32B judge scores the relevance of the retrieved experience, giving $R_{\mathrm{rel}}=0.1s$. The rewrite reward $R_{\mathrm{rewrite}}=0.1u$ scores answerability and strategic utility. Section~\ref{sec:posttraining} describes reward aggregation and prompt-group normalization.

%
%
%
%
\subsubsection{MemEvolver}
\label{sec:app_reward_evolution}
The process reward assigns $R_{\mathrm{act}}=0.005$ for a single SID action, $R_{\mathrm{proto}}=0.005$ for protocol validity, and $R_{\mathrm{SID}}^{\mathrm{E}}=0.0225m$ for SID agreement under a valid protocol. Process scores are assigned to the corresponding output token spans, and the outcome score is placed at termination. A lookup of an existing experience earns $R_{\mathrm{lookup}}=0.01+0.05s$. The evolution reward is $R_{\mathrm{evol}}=0.05u$ for an update that passes checks for reusability, substantive improvement, and faithfulness to the trajectory. A draft that passes the protocol and length checks earns $R_{\mathrm{draft}}=0.01$. The scalar reward record sums these terms and the outcome reward.

\subsubsection{Outcome Reward}
\label{sec:app_reward_outcome}
For MemRetriever, task execution uses the rewritten experience and falls back to the retrieved payload when a rewrite is unavailable. For MemEvolver, it uses the evolved payload. Under the reward schedule used here, a correct answer earns $R_{\mathrm{out}}=2$ and an incorrect answer earns $0$. Appendix~\ref{sec:app_outcome_prompts} shows the task-execution prompts.
}

\subsection{Design Choices and Rationale}
\label{sec:app_design_choices}
\label{sec:method_discussion} 

\paragraph{Fixed SID addressing and editable memory content.}
RQ-KMeans constructs the initial SID space from seed embeddings, after which the codebooks remain fixed (Section~\ref{sec:sid_construction}). Online adaptation changes the memory bank rather than the codebooks or model parameters. Addressing invariance additionally requires that existing SIDs are retained when their payloads change, as specified by Eq.~\ref{eq:app_online_payload_update}. MemEvolver generates the target address; revised text is not re-quantized to assign a replacement SID. This separation permits experience to be updated without invalidating its address. It does not ensure that every revision remains relevant to earlier queries, or that the fixed address space represents all future experience equally well.

\paragraph{Memory-operation mid-training and joint policy post-training.}
Mid-training teaches the policies how to use SIDs within complete memory interactions, using teacher-provided reasoning traces and role-specific outputs under a next-token objective (Section~\ref{sec:midtraining}). Post-training evaluates the resulting behaviour through SID, strategic-quality, and outcome rewards (Section~\ref{sec:posttraining}). Demonstrations specify how to perform an operation; task feedback assesses whether the resulting support helps StackPlanner. The KL term in Eq.~\ref{eq:grpo} regularizes deviation from the reference policy and is distinct from the supervised teacher targets used in mid-training.

\paragraph{Insertion and revision.}
The two-operation interface separates adding an entry from modifying an existing one (Section~\ref{sec:memory_operations}). Revision includes retention: if the trajectory provides no useful change, MemEvolver can emit \texttt{revise} with $\tilde{\mathbf{x}}_{e,t}=\mathcal{M}_t[\mathbf{z}_{e,t}]$, leaving the addressed payload unchanged. Consolidation and deletion likewise use revised or empty payloads rather than separate action labels. This interface allows selective updates without requiring a distinct \texttt{keep} action, but does not guarantee that the policy identifies the appropriate update. That behaviour must be assessed through update traces and downstream evaluation, not inferred from the operation vocabulary alone.

\subsection{Stable Addresses and Editable Memory Content}

Let $\mathbf{x}_t=\mathcal{M}_t[\mathbf{z}]$ denote the payload stored at an occupied address $\mathbf{z}$ at time $t$. During cold-start construction, experiences assigned to the same complete SID are consolidated into a single entry. During online evolution, MemEvolver selects a target SID and updates its payload according to Equation~\ref{eq:app_online_payload_update}. Each occupied address therefore holds one current experience text.

\begin{figure}[ht]
\centering
\includegraphics[width=0.95\textwidth]{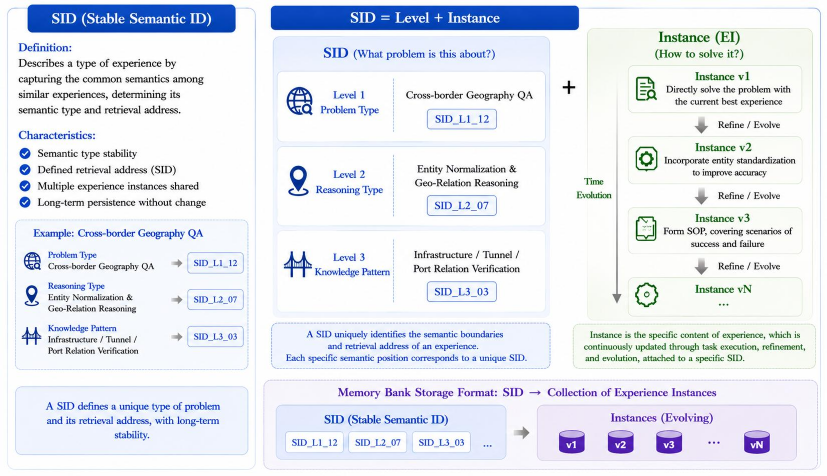}
\caption{Conceptual illustration of SID-addressed memory and experience evolution. The illustrated versions represent successive contents at the same address. Semantic categories and the three-level SID are illustrative; the implemented system uses four residual-quantization levels without predefined semantic roles.}
\label{fig:sid_memory}
\end{figure}

An \texttt{insert} operation populates an empty address, while \texttt{revise} updates an existing entry. Revision encompasses retention, merging, and deletion through unchanged, merged, or empty payloads, respectively. As illustrated in Figure~\ref{fig:sid_memory}, revisions can incorporate new procedures, correct earlier errors, or refine applicability conditions without changing the SID or the codebooks.

This invariance concerns the address, not the meaning or utility of its contents. SID levels likewise have no predefined interpretation as domains, tools, or strategies: they arise from successive residual quantization. Shared prefixes indicate common codebook assignments, but do not establish independent semantic axes. The GPQA evaluation examines whether these prefixes provide useful task information.

\section{Memory Bank Construction and Statistics}
 \label{sec:app_memory_bank}

\subsection{Cold-Start Experience Generation}
\label{sec:app_experience_generation}

The cold-start stage constructs the seed experience bank before memory-policy training (Section~\ref{sec:coldstart}). Let $\mathcal{D}_{\mathrm{train}}$ denote the set of training queries and $y_q^\star$ the reference answer for query $q$, with $y_q^\star=\varnothing$ when no reference answer is available. The notation below describes the information passed through trajectory collection, assessment, and extraction; these stages do not update model parameters.

\paragraph{Trajectory collection.}
For each query $q$, StackPlanner generates $K_q$ attempts. We represent each recorded attempt and the resulting trajectory group as
\begin{equation}
    \tau_q^{(k)}=(\xi_q^{(k)},\hat{y}_q^{(k)},g_q^{(k)}),
    \qquad
    \mathcal{T}(q)=\{\tau_q^{(k)}\}_{k=1}^{K_q},
    \label{eq:app_trajectory_group}
\end{equation}
where $\xi_q^{(k)}$ is the recorded sequence of reasoning steps, actions, and observations, $\hat{y}_q^{(k)}$ is the final answer, and $g_q^{(k)}$ contains any environment reward or completion signals. Missing feedback is represented by $g_q^{(k)}=\varnothing$, not by a failure label. The generation prompts request supporting evidence and verification steps so that subsequent extraction can refer to what the agent actually did.

\paragraph{Outcome and evidence assessment.}
The summarization prompts assess each attempt against the available reference answer or environment feedback. We denote retained assessment information by
\begin{equation}
    \mathcal{A}(q)
    =\bigl\{(\tau_q^{(k)},d_q^{(k)},b_q^{(k)})\bigr\}_{k=1}^{K_q},
    \qquad
    d_q^{(k)}\in\{\mathrm{success},\mathrm{failure},\mathrm{unverified}\},
    \label{eq:app_trajectory_assessment}
\end{equation}
where $d_q^{(k)}$ records whether the available evaluation evidence establishes success, establishes failure, or leaves the outcome unverified. For interactive tasks, a success claim requires an observed environment signal. The textual record $b_q^{(k)}$ captures decisive steps, supporting evidence, failure indications, and low-evidence caveats from the single-trajectory summary. It is not a scalar quality score. Outcome and evidence quality are distinct: a correct final answer without a documented procedure can still be marked as low-evidence. Failed and unverified attempts are retained for comparison, without inventing missing steps or treating uncertain outcomes as verified ones.

\paragraph{Experience extraction.}
An LLM extractor $\mathcal{E}$ uses the query, its available reference, and the assessed attempts to produce a set of textual experience summaries,
\begin{equation}
    \mathcal{X}(q)
    =\mathcal{E}\bigl(q,y_q^\star,\mathcal{A}(q)\bigr).
    \label{eq:app_experience_extraction}
\end{equation}
Each $x\in\mathcal{X}(q)$ describes a reusable procedure or lesson, retaining its applicability conditions, decision rules, verification checks, and failure-recovery strategies. Successful attempts supply supported procedures; failed attempts reveal breakdowns and possible remedies; unverified attempts retain their uncertainty. This comparison is what we mean by \emph{contrastive distillation}: text extraction from contrasting outcomes, not optimization of a contrastive loss. Aggregating the extracted summaries gives
\begin{equation}
    \mathcal{X}_{\mathrm{seed}}
    =\bigcup_{q\in\mathcal{D}_{\mathrm{train}}}
      \mathcal{X}(q).
    \label{eq:app_seed_bank}
\end{equation}
These summaries are the inputs $x_i$ to the encoder and SID construction in Section~\ref{sec:sid_construction}. The single-trajectory and multi-trajectory prompts defining the assessment and extraction content are reproduced in Appendix~\ref{sec:app_trajectory_prompts}.

\subsection{Memory Bank Data Statistics}
\label{sec:app_data_stats}

This section reports the composition, scale, and compression statistics of the memory bank.

\paragraph{Overall pipeline.}
\reviewOne{Raw skill crawling yields 171{,}820 entries. Cluster-level deduplication reduces this to 16{,}384 skill clusters. Together with SOP and factual experiences, the canonical skills form the 138{,}243-entry pre-summary bank. After SID-based consolidation, the earlier $48\times24\times24$ codebook has 16{,}416 occupied leaf addresses. The initial bank used in the main experiments has 30{,}577 occupied leaf addresses under the four-level $48\times16\times8\times8$ codebook. Each occupied address stores one experience payload.}



We first examine the composition of the pre-summary memory bank. As shown in Table~\ref{tab:exp_composition}, SOP and factual experience constitute the majority of the retained entries, while skill experience provides a smaller but complementary source.

\noindent\begin{minipage}{\linewidth}
\centering\small
\captionsetup{font=small,skip=4pt}
\setlength{\tabcolsep}{5pt}
\renewcommand{\arraystretch}{1.08}
\captionof{table}{Composition of the 138{,}243 pre-summary experience entries.}
\label{tab:exp_composition}
\begin{tabularx}{\linewidth}{>{\raggedright\arraybackslash}Xrr}
\toprule
\rowcolor{gray!30}
\textbf{Type} & \textbf{Count} & \textbf{Proportion} \\
\midrule
SOP experience & 62{,}967 & 45.55\% \\
Factual experience & 58{,}892 & 42.60\% \\
Skills experience & 16{,}384 & 11.85\% \\
\bottomrule
\end{tabularx}
\end{minipage}\par\medskip


We next report the source distribution of the retained experience entries. Table~\ref{tab:data_sources} shows that the memory bank draws from a broad collection of interactive, reasoning, and search-oriented datasets rather than being dominated by a single source.

\noindent\begin{minipage}{\linewidth}
\centering\small
\captionsetup{font=small,skip=4pt}
\setlength{\tabcolsep}{5pt}
\renewcommand{\arraystretch}{1.08}
\captionof{table}{\reviewOne{Number of retained experience entries by source in the 138{,}243-entry pre-summary bank.}}
\label{tab:data_sources}
\begin{tabularx}{\linewidth}{>{\raggedright\arraybackslash}Xr>{\raggedright\arraybackslash}Xr}
\toprule
\rowcolor{gray!30}
\textbf{Source} & \textbf{Count} & \textbf{Source} & \textbf{Count} \\
\midrule
WebShop & 30{,}000 & HotpotQA & 9{,}044 \\
Skills & 16{,}384 & NQ & 8{,}080 \\
ALFWorld & 12{,}007 & GSM8K & 7{,}188 \\
2WikiMultiHopQA & 11{,}336 & MATH & 6{,}892 \\
TriviaQA & 9{,}892 & HealthBench & 5{,}216 \\
PopQA & 9{,}596 & GPQA & 1{,}664 \\
MuSiQue & 9{,}312 & \reviewOne{DeepResearch} & \reviewOne{1{,}132} \\
Bamboogle & 500 & & \\
\bottomrule
\end{tabularx}
\end{minipage}\par\medskip

To characterize the effect of compression, Table~\ref{tab:text_length} reports entry-length statistics before and after clustering and summarization. The pipeline substantially reduces both typical and extreme entry lengths, together with the overall text volume.

\noindent\begin{minipage}{\linewidth}
\centering\small
\captionsetup{font=small,skip=4pt}
\setlength{\tabcolsep}{5pt}
\renewcommand{\arraystretch}{1.08}
\captionof{table}{Percentile statistics of entry length (characters) at each pipeline stage.}
\label{tab:text_length}
\begin{tabularx}{\linewidth}{>{\raggedright\arraybackslash}Xrrrr}
\toprule
\rowcolor{gray!30}
\textbf{Stage} & \textbf{P50} & \textbf{P95} & \textbf{P99} & \textbf{Max} \\
\midrule
Skills (raw crawl) & 4{,}748 & 20{,}720 & 40{,}253 & 5{,}425{,}092 \\
Skills (post-cluster) & 2{,}298 & 2{,}942 & 3{,}412 & 5{,}415 \\
Memory (pre-summary) & 8{,}131 & 20{,}208 & 26{,}071 & 123{,}653 \\
Memory (post-summary) & 3{,}116 & 10{,}274 & 19{,}075 & 50{,}204 \\
\bottomrule
\end{tabularx}
\par\vspace{4pt}\raggedright
Total text volume: ${\sim}1.075\times10^9$ characters pre-summary, ${\sim}1.18\times10^8$ post-summary (${\approx}9.1\times$ compression).
\end{minipage}\par\medskip

\reviewOne{Tables~\ref{tab:merge_stats} and~\ref{tab:long_summary} describe the 30{,}577 summaries in the four-level initial bank. Each occupied SID consolidates 4.52 raw experiences on average.}

\noindent\begin{minipage}{\linewidth}
\centering\small
\captionsetup{font=small,skip=4pt}
\setlength{\tabcolsep}{5pt}
\renewcommand{\arraystretch}{1.08}
\captionof{table}{Distribution of raw entries merged per occupied SID in the four-level bank.}
\label{tab:merge_stats}
\begin{tabularx}{\linewidth}{*{6}{>{\raggedleft\arraybackslash}X}}
\toprule
\rowcolor{gray!30}
\textbf{Mean} & \textbf{P50} & \textbf{P90} & \textbf{P95} & \textbf{P99} & \textbf{Max} \\
\midrule
4.52 & 3 & 10 & 14 & 28 & 316 \\
\bottomrule
\end{tabularx}
\end{minipage}\par\medskip

\noindent\begin{minipage}{\linewidth}
\centering\small
\captionsetup{font=small,skip=4pt}
\setlength{\tabcolsep}{5pt}
\renewcommand{\arraystretch}{1.08}
\captionof{table}{Fraction of the 30{,}577 summaries in the four-level bank at or above each character threshold.}
\label{tab:long_summary}
\begin{tabularx}{\linewidth}{>{\raggedright\arraybackslash}Xrr}
\toprule
\rowcolor{gray!30}
\textbf{Threshold} & \textbf{Count} & \textbf{Fraction} \\
\midrule
$\geq$4{,}096 chars & 5{,}370 & 17.56\% \\
$\geq$8{,}192 chars & 2{,}368 & 7.74\% \\
$\geq$16{,}384 chars & 601 & 1.97\% \\
$\geq$32{,}768 chars & 12 & 0.039\% \\
$\geq$65{,}536 chars & 0 & 0.00\% \\
\bottomrule
\end{tabularx}
\end{minipage}\par\medskip

\section{Experimental Datasets and Evaluation Protocols}
\label{sec:app_experiments}

\subsection{Evaluation Benchmarks}
\label{sec:app_dataset_stats}


\begin{table}[ht]
\centering\small
\setlength{\tabcolsep}{5pt}
\renewcommand{\arraystretch}{1.08}
\caption{Evaluation benchmarks and metric correspondence. \reviewOne{QA designations refer to task-model training.} GPQA is used separately for the symbolic reasoning analysis.}
\label{tab:app_dataset}
\begin{tabularx}{\linewidth}{l l >{\raggedright\arraybackslash}X}
\toprule
\rowcolor{gray!30}
\textbf{Benchmark} & \textbf{Evaluation role} & \textbf{Reported metric} \\
\midrule
ALFWorld
& interactive task
& Subtask and overall success rate \\
WebShop
& interactive task
& Task score and success rate \\
\midrule
2WikiMultiHopQA (2Wiki)
& \reviewOne{Task-model training source}
& Answer F1 \\
HotpotQA
& \reviewOne{Task-model training source}
& Answer F1 \\
Bamboogle
& \reviewOne{Held out from task-model training}
& Answer F1 \\
MuSiQue
& \reviewOne{Held out from task-model training}
& Answer F1 \\
Natural Questions (NQ)
& \reviewOne{Held out from task-model training}
& Answer F1 \\
TriviaQA
& \reviewOne{Held out from task-model training}
& Answer F1 \\
\midrule
BrowseComp-Plus
& Out-of-domain search
& Exact-match accuracy \\
DeepResearch Bench
& Out-of-domain research
& RACE \\
BIRD
& Out-of-domain SQL
& Execution accuracy \\
SWE-bench Verified
& Out-of-domain code
& Resolved rate \\
\midrule
GPQA
& Symbolic reasoning
& Answer accuracy \\
\bottomrule
\end{tabularx}
\end{table}
Table~\ref{tab:app_dataset} maps the evaluation benchmarks to the main-text results and their reported metrics. ALFWorld and WebShop are in-domain interactive tasks. \reviewOne{For search-augmented QA, 2WikiMultiHopQA and HotpotQA are task-model training sources; all six QA benchmarks contribute seed experiences.} We further evaluate cross-domain generalization on BrowseComp-Plus~\citep{chen2025browsecomp} for search, DeepResearch Bench~\citep{du2026deepresearch} for research, BIRD~\citep{li2023can} for SQL generation, and SWE-bench Verified~\citep{jimenez2024swe} for software engineering. GPQA is used separately for the symbolic reasoning experiment in Section~\ref{sec:symbolic_experiments}.

The SID retrieval evaluation is distinct from these downstream benchmarks. Table~\ref{tab:app_full_retrieval} reports 2{,}000 evaluation examples for each MemRetriever configuration and 735 for MemEvolver. The latter comprise 590 \texttt{merge\_existing} and 145 \texttt{insert\_new} examples (Table~\ref{tab:app_meme_breakdown}); these are evaluation categories, not additional memory operations. Memory-bank construction counts are reported separately in Appendix~\ref{sec:app_data_stats} and should not be interpreted as evaluation-set sizes.

%
%
{\color{ReviewThree}%
\paragraph{Seed and evaluation overlap.}
\label{sec:app_seed_overlap}
The audit matches source IDs after removing rollout suffixes and checks normalized question text for query--SID alignment. MuSiQue (2{,}328 train questions / 9{,}312 entries), NQ (2{,}020 / 8{,}080), and TriviaQA (2{,}473 / 9{,}892) have zero seed overlap with their dev/validation evaluations (2{,}417, 3{,}610, and 4{,}000 questions, respectively). Bamboogle uses the same 125 test questions for seed construction (500 entries) and evaluation. GPQA contributes 416 training-split seed questions (1{,}664 entries); the perturbation diagnostic selects 100 in-bank questions from the 448-question GPQA pool. None of these five sources is used for task-model training.
}

\subsection{Evaluation Metrics}
\label{sec:app_metrics}

\paragraph{Downstream task performance.}
For an evaluation set of $n$ episodes, let $b_j\in\{0,1\}$ denote benchmark-verified task success. The success rate is
\begin{equation}
    \mathrm{SR}=\frac{100}{n}\sum_{j=1}^{n}b_j.
    \label{eq:app_success_rate}
\end{equation}
For ALFWorld, success is determined by the environment, not by matching an answer string. The Pick, Look, Clean, Heat, Cool, and Pick2 columns apply this measure to the corresponding task subsets; the All column reports overall success. WebShop reports both task score and success rate. The score averages graded task rewards, whereas success counts completed tasks satisfying the benchmark's success criterion. The exact score normalization and overall aggregation used by each baseline source remain part of the reproduction checks in Appendix~\ref{sec:app_baselines}.

\paragraph{QA F1 and symbolic reasoning accuracy.}
For answer token precision $P_j$ and recall $Q_j$ on example $j$, token F1 is $F_{1,j}=2P_jQ_j/(P_j+Q_j)$, with zero assigned when there is no token overlap. The QA tables report the mean example-level F1 as a percentage. Answer normalization, multiple-reference handling, and empty-answer conventions must follow the evaluation script used for each dataset and remain to be recorded. The Avg.\ column in Table~\ref{tab:qa_results} summarizes the six dataset-level scores, rather than pooling all questions into a single evaluation set. The GPQA experiment instead reports answer accuracy, computed as in Eq.~\ref{eq:app_success_rate} with $b_j$ indicating a correct answer choice; it is not a retrieval hit rate.

\paragraph{Candidate and final retrieval hit rates.}
Let $\mathbf{z}_j^\star$ be the reference SID for evaluation example $j$, and let $\mathcal{A}_{j,k}$ contain the first $k$ predicted addresses under a specified ranking. Define
\begin{equation}
    H@k(\mathcal{A})=\frac{100}{n}\sum_{j=1}^{n}
        \mathbb{I}[\mathbf{z}_j^\star\in\mathcal{A}_{j,k}].
    \label{eq:app_retrieval_hit}
\end{equation}
Candidate Hit applies this criterion to the full candidate set before reranking. R@$k$ in Table~\ref{tab:app_full_retrieval} uses the beam-search ranking, whereas HR@$k$ uses the reranked list. In Table~\ref{tab:evolution_retrieval}, final Hit@$k$ is computed on the reranked list for all methods. These single-reference hit rates measure whether the target is present, not the fraction of returned memories judged relevant; they are not precision@$k$.

\paragraph{Independent level and cumulative prefix accuracy.}
For $\mathbf{z}=(s^1,\ldots,s^L)$, write $\mathbf{z}_{1:\ell}=(s^1,\ldots,s^\ell)$. The two hierarchical metrics in Tables~\ref{tab:app_beam_hitrates} and~\ref{tab:app_rerank_hitrates} are
\begin{equation}
\begin{aligned}
    H_{\ell}@k
    &=\frac{100}{n}\sum_{j=1}^{n}
       \mathbb{I}[\exists\,\mathbf{z}\in\mathcal{A}_{j,k}:s^\ell=s_j^{\star,\ell}],\\
    H_{1:\ell}@k
    &=\frac{100}{n}\sum_{j=1}^{n}
       \mathbb{I}[\exists\,\mathbf{z}\in\mathcal{A}_{j,k}:\mathbf{z}_{1:\ell}=\mathbf{z}_{j,1:\ell}^\star].
\end{aligned}
    \label{eq:app_hierarchical_hit}
\end{equation}
Independent accuracy requires a match only at level $\ell$. Cumulative accuracy requires all first $\ell$ symbols to match within the same candidate. At $\ell=L$, cumulative accuracy equals full-SID hit rate. Table~\ref{tab:sid_alignment} reports cumulative Top-5 accuracy after reranking, matching the corresponding rows of Table~\ref{tab:app_rerank_hitrates}; its ROUGE-L and LLM-Judge values assess generated experience text separately and are not percentages of correct SIDs. The judge rubric and score normalization remain to be documented.

\paragraph{Reporting scope.}
The MemEvolver update-type breakdown reports SID retrieval performance within each operation category, not accuracy of choosing \texttt{insert} versus \texttt{revise}. Training rewards are also distinct from held-out task metrics. No statistical-significance claim is made here without repeated-run results and a specified testing procedure.

\paragraph{Cross-domain evaluation metrics.}
BrowseComp-Plus is evaluated using normalized exact-match accuracy. DeepResearch Bench is evaluated using RACE. BIRD is evaluated using execution accuracy, obtained by executing the predicted and reference SQL queries on the same database and comparing their returned results. SWE-bench Verified is evaluated using resolved rate under the official evaluation harness.

\subsection{Online Evaluation Protocol}
\label{sec:app_online_eval}

\reviewOne{All methods process the same query stream in a fixed order and use the read-only probe protocol below. Appendix~\ref{sec:app_seed_overlap} reports overlap between seed sources and downstream evaluation sets. The evolving-bank and frozen-bank conditions start with the same bank, model weights, encoder, and codebooks. During memory updates, the model weights and codebooks remain fixed, and the update policy sees the execution trajectory without the reference answer.}

For each incoming query $q_t$, the system generates one SID, retrieves the addressed support, and executes StackPlanner to obtain an answer $\hat y_t$ and trajectory $\tau_t$. We score $(q_t,\hat y_t)$ immediately, then run the evolution policy on $\tau_t$ and apply its emitted \texttt{insert} or \texttt{revise} operation to obtain $\mathcal{M}_{t+1}$.
\reviewOne{After each fixed-size stream block, we evaluate on the same read-only probe set selected before the run.} We store the bank snapshot and all operation logs at each probe point.

\reviewOne{Table~\ref{tab:evolution_retrieval} reports candidate hit rate, reranked Hit@1 and Hit@5, and task accuracy across checkpoints. Figure~\ref{fig:evolution_stats} shows memory-update statistics. Table~\ref{tab:ablation} also includes a frozen-bank control with memory writes disabled.}

\section{Baselines, Implementation, and Training Details}
\label{sec:app_implementation}

\subsection{Comparison Methods and Settings}
\label{sec:app_baselines}

\paragraph{ALFWorld and WebShop.}
Table~\ref{tab:main_results} compares GPT-4o and Gemini-2.5-Pro; the prompt-based or memory-based methods ReAct, Reflexion, Mem0, MemP, ExpeL, and SimpleMem; the RL baselines RLOO and GRPO; and the memory-augmented RL methods MemRL, EvolveR, Mem0+GRPO, SimpleMem+GRPO, SkillRL, SkillGraph, AgentOCR, SIRI, and Skill0. The \methodname{} and \methodname{} (RL) rows distinguish the configurations before and after Harness-RL training of StackPlanner, as described in Section~\ref{sec:main_results}.

\paragraph{Search-augmented QA.}
Table~\ref{tab:qa_results} includes Base and CoT prompting, FS-RAG, FL-RAG, ReAct, IRCoT, and TCRAG; the RL methods ReSearch, Search-R1, AEPO, ARPO, KBPO, Mem1, ZeroSearch, and AgenticRAG-R1; and EvolveR, SkillRL, SkillGraph, AgentOCR, Skill0, Skill-SD, SAPO, and SDAR. Method citations are provided in Section~\ref{sec:exp_setup}. Entries marked -- are unavailable results, not zero scores, and should not enter performance rankings or averages.

\paragraph{Experience retrieval.}
The comparison in Table~\ref{tab:evolution_retrieval} uses Qwen3-Embedding-0.6B, TF-IDF, SkillRouter-Embedding-0.6B, and \methodname{} with up to 50 candidates followed by BGE reranking to Top-5. All methods therefore share the same candidate budget and reranking protocol, enabling a matched-budget comparison of candidate and final retrieval hit rates.

\paragraph{Reproduction scope.}
%
%
%
{\color{ReviewOne}%
We rerun FS-RAG, FL-RAG, TCRAG, and AgenticRAG-R1 using their native harnesses and take the remaining baseline scores in Tables~\ref{tab:main_results} and~\ref{tab:qa_results} from the original publications. Table~\ref{tab:skill_transfer} includes our OpenHands run and compares each harness with and without \methodname{} skills under a fixed backbone.
}

\subsection{Training Hyperparameters}
\label{sec:app_sid_hyperparams}

%
We use Qwen2.5-7B-Instruct as the backbone for both memory policies and optimize them with GRPO during post-training. The SID codebook is fixed to the four-level configuration $48\times16\times8\times8$. \reviewThree{Table~\ref{tab:app_sid_hyper} separates the two GRPO schedules.}

\begin{table}[h]
\centering
\caption{Hyperparameters for reinforcement learning}
\label{tab:app_sid_hyper}
\begin{tabular}{lcc}
\toprule
\textbf{Hyperparameter} & \reviewThree{\textbf{Memory policies}} & \reviewThree{\textbf{StackPlanner}} \\
\midrule
Backbone model & Qwen2.5-7B-Instruct & \reviewThree{Qwen2.5-7B-Instruct} \\
\reviewThree{Rollouts per prompt} & 8 & \reviewThree{4} \\
\reviewThree{Prompts per training batch} & \reviewThree{16} & -- \\
\reviewThree{Optimization minibatch} & \reviewThree{8} & -- \\
KL coefficient $\beta$ & \reviewThree{0.003} & -- \\
Training steps (GRPO) & \reviewThree{130} & \reviewThree{200} \\

\bottomrule
\end{tabular}
\end{table}

\section{Additional Experiments and Analysis}
\label{sec:app_additional_experiments}
  
\subsection{RQ-KMeans Evaluation Metrics and Configuration Analysis}
\label{sec:app_rqkmeans_metrics}
\paragraph{Prefix-prompt comparison.}
\label{sec:app_prefix_comparison}
Table~\ref{tab:prefix_prompt_ablation} compares encoding with and without the prefix, holding the seed bank, encoder, and reported Weighted setting ($w=0.5$) fixed. This experiment uses a three-level $48\times24\times24$ codebook, not the four-level $48\times16\times8\times8$ configuration used in the main system. Under this three-level setting, the prefix reduces the reported reconstruction loss but also slightly reduces leaf occupancy. It is not a direct ablation of prefix guidance in the final four-level system.

\begin{table}[ht]
\centering\small
\caption{Prefix-prompt comparison with a three-level $48\times24\times24$ codebook and the reported Weighted setting ($w=0.5$). The encoder and seed bank are held fixed. Values and rounded differences are transcribed from the supplied experiment summary. SID uniqueness is computed over 13{,}824 evaluation examples.}
\label{tab:prefix_prompt_ablation}
\begin{tabular}{lrrr}
\toprule
\rowcolor{gray!30}
\textbf{Metric} & \textbf{Prefix} & \textbf{No prefix} & \textbf{$\Delta$} \\
\midrule
Test reconstruction loss $\downarrow$ & 0.000332 & 0.000392 & $-15.27\%$ \\
Single-code utilization (L1/L2/L3) $\uparrow$ & 100/100/100\% & 100/100/100\% & $0$ \\
Cumulative L1+L2 utilization $\uparrow$ & 84.72\% & 83.25\% & $+1.48$ pp \\
Leaf-combination utilization $\uparrow$ & 23.75\% & 23.93\% & $-0.18$ pp \\
Used leaves & 6{,}567 & 6{,}617 & $-50$ \\
DUI uniformity $\uparrow$ & 0.8929 & 0.8730 & $+0.0199$ \\
ICR (independent-code rate) $\uparrow$ & 1.1487 & 1.1422 & $+0.0064$ \\
SID uniqueness $\uparrow$ & 47.50\% & 47.87\% & $-0.36$ pp \\
RQ-KMeans training time & 77.6 s & 73.6 s & $+4.0$ s \\
\bottomrule
\end{tabular}
\end{table}

\paragraph{RQ-KMeans codebook configuration analysis.}
\label{sec:app_rqkmeans_config}

We compare 40 codebook configurations spanning two to four quantization levels and address capacities from 27{,}648 to 147{,}456 (Table~\ref{tab:app_codebook_search_space}). Each configuration is evaluated under three reported distance settings (Euclidean, Spherical, Weighted $w=0.5$), giving 120 combinations fitted on the same 138{,}243 pre-merge embeddings. The comparison examines the trade-off between address capacity, SID vocabulary size, and quantization quality, reporting reconstruction loss, leaf utilization, SID uniqueness, and DUI.

\begin{table}[ht]
\centering\small
\setlength{\tabcolsep}{5pt}
\renewcommand{\arraystretch}{1.08}
\caption{Codebook search space. Each tuple lists the codebook sizes $(N_1,\ldots,N_L)$; counts in parentheses give the number of configurations at each depth.}
\label{tab:app_codebook_search_space}
\begin{tabularx}{\linewidth}{l>{\raggedright\arraybackslash}X}
\toprule
\rowcolor{gray!30}
\textbf{Quantization depth} & \textbf{Codebook configurations} \\
\midrule
Two levels (3) & $256{\times}256$,\quad $320{\times}256$,\quad $384{\times}256$ \\
\midrule
Three levels (30) & $48{\times}24{\times}24$,\quad $48{\times}32{\times}32$,\quad $64{\times}32{\times}24$,\quad $64{\times}32{\times}32$,\quad $64{\times}40{\times}20$ \\
 & $64{\times}48{\times}20$,\quad $64{\times}48{\times}24$,\quad $72{\times}24{\times}16$,\quad $72{\times}32{\times}24$,\quad $80{\times}32{\times}16$ \\
 & $80{\times}32{\times}20$,\quad $80{\times}32{\times}32$,\quad $80{\times}40{\times}24$,\quad $80{\times}48{\times}16$,\quad $88{\times}40{\times}8$ \\
 & $96{\times}24{\times}24$,\quad $96{\times}32{\times}16$,\quad $96{\times}32{\times}24$,\quad $96{\times}40{\times}16$,\quad $96{\times}40{\times}24$ \\
 & $96{\times}48{\times}16$,\quad $96{\times}48{\times}24$,\quad $128{\times}24{\times}16$,\quad $128{\times}24{\times}24$,\quad $128{\times}32{\times}16$ \\
 & $128{\times}32{\times}24$,\quad $128{\times}40{\times}16$,\quad $128{\times}40{\times}24$,\quad $144{\times}40{\times}24$,\quad $192{\times}32{\times}24$ \\
\midrule
Four levels (7) & $48{\times}16{\times}8{\times}8$,\quad $48{\times}24{\times}16{\times}8$,\quad $64{\times}16{\times}8{\times}8$,\quad $64{\times}16{\times}16{\times}8$ \\
 & $64{\times}24{\times}8{\times}8$,\quad $80{\times}16{\times}8{\times}8$,\quad $96{\times}16{\times}8{\times}8$ \\
\bottomrule
\end{tabularx}
\end{table}

\paragraph{Distance Metric Comparison.}
\label{sec:app_distance_comparison}

Table~\ref{tab:app_distance_summary} summarizes each distance setting across the 40 architectures. \reviewThree{The Weighted setting uses the combination of squared Euclidean and cosine distances in Eq.~\ref{eq:rq_weighted_distance}.} The standard deviation describes variation across architectures, not uncertainty across repeated random seeds. Reconstruction loss is reported in units of $\times10^{-4}$.

\begin{table}[ht]
\centering\small
\caption{Distance-setting comparison, with mean $\pm$ standard deviation across 40 architectures. Loss is reported in $\times10^{-4}$.}
\label{tab:app_distance_summary}
\begin{tabular}{lcccc}
\toprule
\rowcolor{gray!30}
\textbf{Distance} & \textbf{Recon.\ Loss ($\times 10^{-4}$)} & \textbf{Leaf Util.\ (\%)} & \textbf{Uniqueness Ratio} & \textbf{Avg.\ DUI (\%)} \\
\midrule
Euclidean & $3.38 \pm 0.09$ & $14.92 \pm 3.68$ & $0.0765 \pm 0.0241$ & $56.53 \pm 3.71$ \\
Spherical & $5.98 \pm 1.60$ & $5.14 \pm 6.28$ & $0.0270 \pm 0.0375$ & $81.90 \pm 1.93$ \\
Weighted ($w{=}0.5$) & $3.32 \pm 0.10$ & $49.15 \pm 8.89$ & $0.2523 \pm 0.0719$ & $88.36 \pm 2.04$ \\
\bottomrule
\end{tabular}
\end{table}

\paragraph{Selected configuration vs.\ competitors.}
\label{sec:app_config_comparison}

Table~\ref{tab:app_config_comparison} compares the selected configuration ($48\times16\times8\times8$, reported Weighted setting $w=0.5$) with five alternatives in the 40K--60K capacity band. This is a trade-off between vocabulary size and the reported quantization metrics, rather than a claim that one configuration is best on every measure.

\begin{table}[ht]
\centering\small
\caption{Configuration comparison within the 40K--60K capacity band. The selected configuration (\textbf{bold}) uses the fewest SID tokens among the configurations shown. Loss is reported in $\times10^{-4}$.}
\label{tab:app_config_comparison}
\begin{tabular}{lccccccc}
\toprule
\rowcolor{gray!30}
\textbf{Codebook} & \textbf{$L$} & \textbf{Capacity} & \textbf{Vocab} & \textbf{Recon.} & \textbf{Leaf Util.} & \textbf{Uniq.} & \textbf{DUI} \\
\rowcolor{gray!30}
 & & & \textbf{Size} & \textbf{($\times 10^{-4}$)} & \textbf{(\%)} & \textbf{Ratio} & \textbf{(\%)} \\
\midrule
\textbf{48$\!\times\!$16$\!\times\!$8$\!\times\!$8} & \textbf{4} & \textbf{49{,}152} & \textbf{80} & \textbf{$3.48$} & \textbf{$62.22$} & \textbf{$0.2212$} & \textbf{$90.70$} \\
\midrule
128$\!\times\!$24$\!\times\!$16 & 3 & 49{,}152 & 168 & $3.31$ & $64.25$ & $0.2284$ & $89.48$ \\
96$\!\times\!$24$\!\times\!$24 & 3 & 55{,}296 & 144 & $3.34$ & $56.71$ & $0.2268$ & $88.74$ \\
72$\!\times\!$32$\!\times\!$24 & 3 & 55{,}296 & 128 & $3.36$ & $51.26$ & $0.2051$ & $88.44$ \\
96$\!\times\!$32$\!\times\!$16 & 3 & 49{,}152 & 144 & $3.34$ & $57.49$ & $0.2044$ & $87.53$ \\
64$\!\times\!$40$\!\times\!$20 & 3 & 51{,}200 & 124 & $3.36$ & $52.93$ & $0.1960$ & $88.13$ \\
\bottomrule
\end{tabular}
\end{table}

\paragraph{Configuration trade-offs.}
\label{sec:app_rqkmeans_findings}

The reported Weighted setting has higher mean leaf utilization than the Euclidean setting ($49.15\%$ versus $14.92\%$), with similar reported reconstruction loss ($3.32$ versus $3.38$, in units of $10^{-4}$). These aggregate results compare distance settings; they do not isolate the effect of hierarchy depth. Within Table~\ref{tab:app_config_comparison}, the selected four-level configuration represents $49{,}152$ possible addresses with 80 SID tokens, compared with 124--168 tokens for the listed three-level alternatives. Its leaf utilization is $62.22\%$, but its reported reconstruction loss is higher than that of every listed alternative. The configuration therefore favors a smaller vocabulary rather than minimum reconstruction error. These measurements do not establish effects on catastrophic forgetting, training convergence, or downstream task accuracy.

\subsection{Pretraining and Mid-Training Retrieval Analysis}
\label{sec:app_pt_mt}

This section systematically compares SID retrieval quality across pretraining (PT) and mid-training (MT) stages for both the legacy 3-level and current 4-level codebooks. We evaluate MemRetriever and MemEvolver under beam search and reranking, reporting recall at multiple cutoffs and per-level hit rates. All retrieval evaluations \reviewOne{in the PT/MT comparison} use uniformly sampled training examples from the six search benchmarks listed in Table~\reviewOne{\ref{tab:qa_results}}, together with ALFWorld and WebShop.

Table~\ref{tab:sid_alignment} gives the SID alignment and experience-reconstruction scores omitted from the main text to conserve space.

\begin{table}[ht]
\centering
\scriptsize
\setlength{\tabcolsep}{2.2pt}
\renewcommand{\arraystretch}{1.0}
\caption{SID alignment accuracy (\%) at each cumulative hierarchical level, measured at Top-5 after reranking. For SID$\to$experience reconstruction, ROUGE-L and LLM-Judge scores are reported separately.}
\label{tab:sid_alignment}
\begin{tabular}{@{}l cccc@{}}
\toprule
\rowcolor{gray!30}
\textbf{Component} & \textbf{$L_1$} & \textbf{$L_1$--$L_2$} & \textbf{$L_1$--$L_3$} & \textbf{$L_1$--$L_4$} \\
\midrule
\makecell[l]{MemRetriever\\(query$\to$SID)} & 40.90 & 13.00 & 6.30 & 3.75 \\
\makecell[l]{MemEvolver\\(traj$\to$SID)} & 40.41 & 11.29 & 3.27 & 1.36 \\
\midrule
\rowcolor{gray!30}
\textbf{SID$\to$experience rewrite} & \multicolumn{2}{c}{\textbf{ROUGE-L}: 0.53} & \multicolumn{2}{c}{\textbf{LLM-Judge}: 0.57} \\
\bottomrule
\end{tabular}
\end{table}

Model abbreviations used throughout:
\begin{itemize}[leftmargin=*,itemsep=1pt]
    \item \textbf{PT-3L}: Pretrained MemRetriever (MemR) with 3-level codebook $(48 \times 24 \times 24)$.
    \item \textbf{PT-4L}: Pretrained MemR with 4-level codebook $(48 \times 16 \times 8 \times 8)$.
    \item \textbf{MT-3L}: Mid-trained MemR with 3-level codebook $(48 \times 24 \times 24)$.
    \item \textbf{MT-4L-Full}: Mid-trained 4-level COT MemR with full-sequence beam decoding.
    \item \textbf{MT-4L-SID}: Mid-trained 4-level COT MemR with SID-only beam decoding.
    \item \textbf{MT-4L-MemE}: Mid-trained 4-level COT MemEvolver (MemE) with training-isomorphic schema.
\end{itemize}

\paragraph{Codebook configuration}
\label{sec:app_codebook_config}

Table~\ref{tab:app_codebook_config} compares the two codebook configurations. The 4-level factorization offers a larger theoretical address space with shorter per-level vocabularies, reducing per-token prediction entropy during autoregressive SID generation.

\begin{table}[ht]
\centering\small
\setlength{\tabcolsep}{4pt}
\renewcommand{\arraystretch}{0.9}
\caption{Codebook configuration: 3-level vs.\ 4-level SID.}
\label{tab:app_codebook_config}
\begin{tabular}{lcc}
\toprule
\rowcolor{gray!30}
\textbf{Property} & \textbf{3-Level} & \textbf{4-Level} \\
\midrule
Number of levels & 3 & 4 \\
L1 size & 48 & 48 \\
L2 size & 24 & 16 \\
L3 size & 24 & 8 \\
L4 size & --- & 8 \\
SID tokens per address & 3 & 4 \\
Theoretical capacity & $48 \times 24 \times 24 = 27{,}648$ & $48 \times 16 \times 8 \times 8 = 49{,}152$ \\
\bottomrule
\end{tabular}
\end{table}

\paragraph{Full SID Retrieval Results}
\label{sec:app_pt_mt_full_retrieval}

Table~\ref{tab:app_full_retrieval} reports end-to-end SID retrieval performance across all six configurations, measuring beam-search recall (R@$k$) and reranking hit rate (HR@$k$) after dense rescoring.

\begin{table}[ht]
\centering\small
\setlength{\tabcolsep}{2.7pt}
\renewcommand{\arraystretch}{0.9}
\caption{Full SID retrieval results across PT and MT configurations. R@$k$: beam recall; HR@$k$: reranking hit rate.}
\label{tab:app_full_retrieval}
\begin{tabular}{lccccccccc}
\toprule
\rowcolor{gray!30}
\textbf{Model} & \textbf{N} & \textbf{Avg Cand.} & \textbf{R@1} & \textbf{R@5} & \textbf{R@10} & \textbf{R@20} & \textbf{R@50} & \textbf{HR@1} & \textbf{HR@5} \\
\midrule
PT-3L (48$\times$24$\times$24)  & 2000 & 49.98 & 1.15 & 4.30 & 7.30 & 11.25 & 20.80 & 2.25 & 4.85 \\
PT-4L (48$\times$16$\times$8$\times$8) & 2000 & 49.57 & 1.40 & 4.95 & 7.90 & 13.40 & 20.25 & 7.00 & 13.15 \\
MT-3L MemR (48$\times$24$\times$24) & 2000 & 29.43 & 0.75 & 1.75 & 2.90 & 5.05 & 10.85 & 2.95 & 5.35 \\
MT-4L COT MemR (Full Beam) & 2000 & 33.72 & 0.40 & 1.05 & 1.50 & 2.20 & 4.90 & 1.85 & 3.10 \\
MT-4L COT MemR (SID-only) & 2000 & 50.00 & 0.35 & 1.15 & 1.80 & 3.35 & 6.80 & 2.25 & 3.75 \\
MT-4L COT MemE & 735 & 50.00 & 0.14 & 0.54 & --- & --- & 5.31 & 0.68 & 1.36 \\
\bottomrule
\end{tabular}
\end{table}

{\color{ReviewOne}%
\paragraph{PT/MT utility and post-GRPO retrieval.}
The PT-4L row supplies the initial retrieval scores in Table~\ref{tab:evolution_retrieval}. Under SID-only decoding, reference HR@5 drops from $13.15\%$ to $3.75\%$ after MT. In contrast, MT raises rewrite quality from 5.600 to 7.885 (/8), winning 76 of 100 blinded judgments (PT wins 1; 23 ties), and raises downstream exact match from $19.57\%$ to $21.74\%$ and token F1 from $28.10\%$ to $31.17\%$. On a separate 200-question diagnostic across nine datasets, MemRetriever v11 (step 50) reaches $50.50\%$ R@50, $14.50\%$ HR@1, and $24.00\%$ HR@5 with constrained SID Beam@50 and BGE reranking.
}

\paragraph{SID Decoding Configurations}

\label{sec:app_sid_decoding}

Table~\ref{tab:app_sid_decoding} compares greedy decoding, deterministic beam search, and beam sampling. Deterministic beam search uses no random sampling. Increasing its beam width from 5 to 20 raises Recall@5 from $7.93\%$ to $8.53\%$, while reported latency increases from 121.0 to 406.1\,ms. None of the sampled configurations exceeds the highest deterministic Recall@5 in this comparison.

\begin{table}[ht]
\centering\small
\setlength{\tabcolsep}{3pt}
\renewcommand{\arraystretch}{1.08}
\caption{SID decoding configurations and retrieval results. Top-1 SID and Recall@5 are percentages; latency is reported in milliseconds. Greedy decoding returns one candidate, so its two retrieval scores coincide. Temperature and top-$p$ entries are transcribed as reported and do not imply sampling in the deterministic rows. Bold denotes the best value in each result column, including ties.}
\label{tab:app_sid_decoding}
\begin{tabularx}{\textwidth}{@{}Xrrrrrrr@{}}
\toprule
\rowcolor{gray!30}
\textbf{Decoding} & \textbf{Beam} & \textbf{Top-$k$} & \textbf{Temp.} & \textbf{Top-$p$} & \textbf{Top-1 SID} & \textbf{Recall@5} & \textbf{Latency} \\
\midrule
Greedy & 1 & 0 & 1.0 & 1.0 & 1.87 & 1.87 & \textbf{35.7} \\
\midrule
Beam deterministic & 5 & 0 & 1.0 & 1.0 & 2.93 & 7.93 & 121.0 \\
Beam deterministic & 10 & 0 & 1.0 & 1.0 & 2.93 & 8.27 & 221.1 \\
Beam deterministic & 20 & 0 & 1.0 & 1.0 & \textbf{3.13} & \textbf{8.53} & 406.1 \\
\midrule
Beam sample & 5 & 0 & 0.7 & 0.9 & 2.53 & 6.53 & 121.5 \\
Beam sample & 5 & 5 & 0.7 & 0.9 & 2.73 & 7.27 & 121.8 \\
Beam sample & 5 & 10 & 0.7 & 0.9 & 2.47 & 7.07 & 121.9 \\
Beam sample & 5 & 20 & 0.7 & 0.9 & 2.93 & 6.87 & 122.0 \\
Beam sample & 5 & 20 & 0.3 & 0.9 & 2.73 & 7.73 & 122.0 \\
Beam sample & 5 & 20 & 0.5 & 0.9 & 2.87 & 7.40 & 122.0 \\
Beam sample & 5 & 20 & 1.0 & 0.9 & 2.60 & 5.67 & 122.1 \\
Beam sample & 10 & 20 & 0.7 & 0.9 & \textbf{3.13} & 7.27 & 224.3 \\
Beam sample & 20 & 20 & 0.7 & 0.9 & 3.07 & 8.27 & 413.6 \\
Beam sample & 5 & 20 & 0.7 & 0.8 & 2.67 & 6.60 & 121.8 \\
Beam sample & 5 & 20 & 0.7 & 0.95 & 2.53 & 6.20 & 122.0 \\
\bottomrule
\end{tabularx}
\end{table}

\paragraph{Beam-search hit rates.}
\label{sec:app_pt_mt_beam}

Table~\ref{tab:app_beam_hitrates} reports per-level independent and cumulative prefix hit rates under beam search. Each level is evaluated independently (left) or cumulatively from L1 through L$\ell$ (right), at cutoffs Top-1/5/50.

\begin{table*}[ht]
\centering
\small
\renewcommand{\arraystretch}{0.85}
\caption{Beam search hit rates (\%) across PT and MT configurations. Left: per-level independent accuracy. Right: cumulative prefix accuracy (L1 through L$\ell$ must all match). L4 is N/A for 3-level models.}
\label{tab:app_beam_hitrates}
{%
\begin{tabular}{ll cccc cccc}
\toprule
 & & \multicolumn{4}{c}{\textbf{Per-Level Independent (\%)}} & \multicolumn{4}{c}{\textbf{Cumulative Prefix (\%)}} \\
\cmidrule(lr){3-6} \cmidrule(lr){7-10}
\textbf{Model} & \textbf{Cutoff} & \textbf{L1} & \textbf{L2} & \textbf{L3} & \textbf{L4} & \textbf{L1} & \textbf{L1+L2} & \textbf{L1\dots L3} & \textbf{L1\dots L4} \\
\midrule
\multirow{3}{*}{PT-3L}
  & Top-1  & 31.85 & 24.80 & 14.00 & ---   & 31.85 &  7.85 &  1.15 & ---   \\
  & Top-5  & 53.85 & 35.40 & 26.60 & ---   & 53.85 & 18.50 &  4.30 & ---   \\
  & Top-50 & 88.20 & 69.55 & 67.05 & ---   & 88.20 & 49.95 & 20.80 & ---   \\
\midrule
\multirow{3}{*}{PT-4L}
  & Top-1  & 22.05 & 26.75 & 31.65 & 33.60 & 22.05 &  7.10 &  2.95 &  1.40 \\
  & Top-5  & 44.25 & 42.30 & 47.50 & 52.05 & 44.25 & 18.80 &  9.05 &  4.95 \\
  & Top-50 & 85.55 & 78.60 & 81.25 & 84.10 & 85.55 & 52.65 & 32.05 & 20.25 \\
\midrule
\multirow{3}{*}{MT-3L}
  & Top-1  & 14.40 & 23.80 & 16.60 & ---   & 14.40 &  3.70 &  0.75 & ---   \\
  & Top-5  & 22.95 & 28.50 & 23.00 & ---   & 22.95 &  7.50 &  1.75 & ---   \\
  & Top-50 & 66.35 & 56.45 & 59.65 & ---   & 66.35 & 29.00 & 10.85 & ---   \\
\midrule
\multirow{3}{*}{MT-4L-Full}
  & Top-1  & 14.15 & 20.20 & 24.80 & 25.30 & 14.15 &  2.90 &  1.05 &  0.40 \\
  & Top-5  & 26.20 & 27.05 & 37.90 & 35.95 & 26.20 &  7.00 &  2.50 &  1.05 \\
  & Top-50 & 65.20 & 57.30 & 69.90 & 70.30 & 65.20 & 25.30 & 11.25 &  4.90 \\
\midrule
\multirow{3}{*}{MT-4L-SID}
  & Top-1  & 14.30 & 20.10 & 25.40 & 25.40 & 14.30 &  3.05 &  1.00 &  0.35 \\
  & Top-5  & 32.20 & 29.10 & 42.10 & 40.40 & 32.20 &  8.70 &  3.10 &  1.15 \\
  & Top-50 & 72.35 & 65.55 & 77.75 & 76.85 & 72.35 & 32.30 & 14.95 &  6.80 \\
\midrule
\multirow{3}{*}{MT-4L-MemE}
  & Top-1  & 16.60 & 19.73 & 23.27 & 25.31 & 16.60 &  3.67 &  0.41 &  0.14 \\
  & Top-5  & 33.88 & 26.26 & 41.09 & 42.99 & 33.88 &  7.62 &  1.77 &  0.54 \\
  & Top-50 & 77.55 & 57.69 & 75.37 & 87.62 & 77.55 & 30.20 & 12.24 &  5.31 \\
\bottomrule
\end{tabular}%
}
\end{table*}

\paragraph{Rerank Hit Rates}
\label{sec:app_pt_mt_rerank}

Table~\ref{tab:app_rerank_hitrates} reports per-level and cumulative metrics after reranking (Beam@50 candidates rescored to Top-1 and Top-5).

\begin{table*}[ht]
\centering
\small
\renewcommand{\arraystretch}{0.85}
\caption{Rerank hit rates (\%) (Beam@50 rescored by dense reranker). Layout matches Table~\ref{tab:app_beam_hitrates}. L4 is N/A for 3-level models.}
\label{tab:app_rerank_hitrates}
{%
\begin{tabular}{ll cccc cccc}
\toprule
 & & \multicolumn{4}{c}{\textbf{Per-Level Independent (\%)}} & \multicolumn{4}{c}{\textbf{Cumulative Prefix (\%)}} \\
\cmidrule(lr){3-6} \cmidrule(lr){7-10}
\textbf{Model} & \textbf{Cutoff} & \textbf{L1} & \textbf{L2} & \textbf{L3} & \textbf{L4} & \textbf{L1} & \textbf{L1+L2} & \textbf{L1\dots L3} & \textbf{L1\dots L4} \\
\midrule
\multirow{2}{*}{PT-3L}
  & Top-1  & 31.80 & 22.05 & 12.60 & ---   & 31.80 &  7.20 &  2.25 & ---   \\
  & Top-5  & 59.55 & 39.25 & 28.80 & ---   & 59.55 & 19.35 &  4.85 & ---   \\
\midrule
\multirow{2}{*}{PT-4L}
  & Top-1  & 26.45 & 28.45 & 32.60 & 32.60 & 26.45 & 11.25 &  8.25 &  7.00 \\
  & Top-5  & 54.45 & 50.25 & 57.35 & 59.85 & 54.45 & 25.70 & 16.40 & 13.15 \\
\midrule
\multirow{2}{*}{MT-3L}
  & Top-1  & 20.05 & 24.55 & 15.85 & ---   & 20.05 &  6.50 &  2.95 & ---   \\
  & Top-5  & 42.80 & 36.60 & 34.15 & ---   & 42.80 & 15.10 &  5.35 & ---   \\
\midrule
\multirow{2}{*}{MT-4L-Full}
  & Top-1  & 16.60 & 21.20 & 25.55 & 24.45 & 16.60 &  4.65 &  2.40 &  1.85 \\
  & Top-5  & 40.05 & 35.45 & 48.00 & 48.65 & 40.05 & 12.10 &  5.35 &  3.10 \\
\midrule
\multirow{2}{*}{MT-4L-SID}
  & Top-1  & 17.60 & 21.00 & 24.95 & 25.00 & 17.60 &  5.10 &  2.85 &  2.25 \\
  & Top-5  & 40.90 & 37.70 & 49.60 & 49.65 & 40.90 & 13.00 &  6.30 &  3.75 \\
\midrule
\multirow{2}{*}{MT-4L-MemE}
  & Top-1  & 16.87 & 19.46 & 23.81 & 22.99 & 16.87 &  3.54 &  1.09 &  0.68 \\
  & Top-5  & 40.41 & 30.88 & 45.71 & 50.88 & 40.41 & 11.29 &  3.27 &  1.36 \\
\bottomrule
\end{tabular}%
}
\end{table*}

\paragraph{MemE Update-Type Breakdown}
\label{sec:app_pt_mt_meme_breakdown}

Table~\ref{tab:app_meme_breakdown} retains the original evaluation labels and values. The label \texttt{merge\_existing} targets already-populated slots and corresponds to a merging outcome of \texttt{revise}, while \texttt{insert\_new} targets unpopulated leaves and corresponds to \texttt{insert}.

\begin{table}[ht]
\centering\small
\setlength{\tabcolsep}{4pt}
\renewcommand{\arraystretch}{0.9}
\caption{MemE retrieval performance by update-operation type.}
\label{tab:app_meme_breakdown}
\begin{tabular}{lccccc}
\toprule
\rowcolor{gray!30}
\textbf{Type} & \textbf{N} & \textbf{R@1} & \textbf{R@5} & \textbf{R@50} & \textbf{Avg Cand.} \\
\midrule
\texttt{merge\_existing} & 590 & 0.169 & 0.339 & 5.763 & 50.00 \\
\texttt{insert\_new}     & 145 & 0.000 & 1.379 & 3.448 & 50.00 \\
\midrule
Total                     & 735 & 0.136 & 0.544 & 5.306 & 50.00 \\
\bottomrule
\end{tabular}
\end{table}

\paragraph{SID Depth Selection: 3-Level vs.\ 4-Level}
\label{sec:app_sid_depth}

We compare a \textbf{3-level} configuration $(N_1, N_2, N_3) = (48, 24, 24)$ against a \textbf{4-level} configuration $(N_1, N_2, N_3, N_4) = (48, 16, 8, 8)$ under identical training conditions (global batch size 48, 500-step moving average).

\begin{figure}[ht]
\centering
\includegraphics[width=0.85\textwidth]{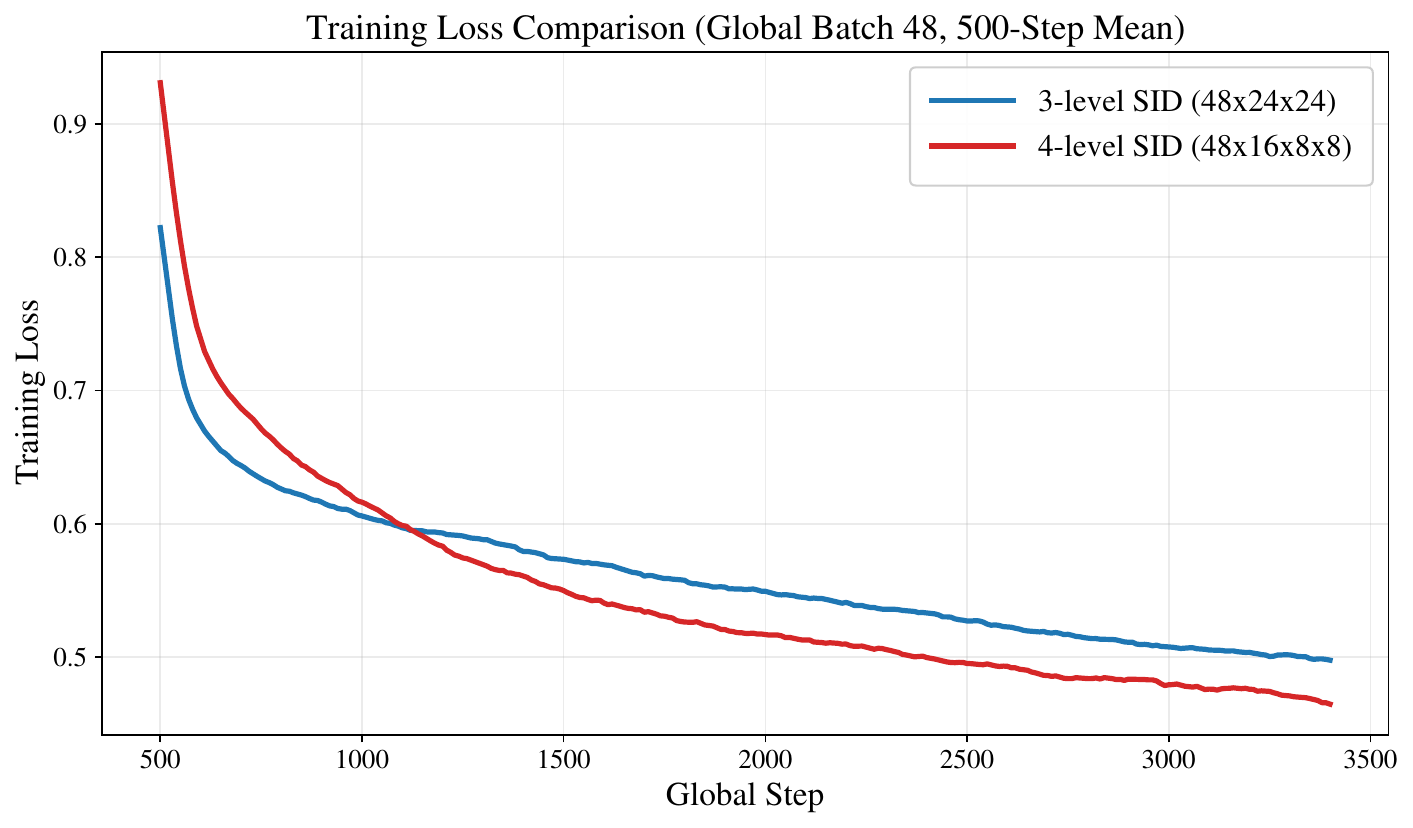}
\caption{Alignment pretraining loss: 3-level vs.\ 4-level SID (500-step moving average, batch size 48). The 4-level SID converges faster and to a lower plateau.}
\label{fig:sid_level_loss}
\end{figure}

As shown in Figure~\ref{fig:sid_level_loss}, the 4-level SID achieves \textbf{consistently lower training loss} throughout alignment pretraining, converging faster and to a lower plateau. Finer factorization provides the autoregressive decoder with shorter per-level prediction targets and sharper categorical distributions, reducing per-token entropy. This motivates the adoption of the 4-level SID as the default configuration.

\paragraph{SID Codebook Utilization Analysis}
\label{sec:app_codebook_util}

\begin{figure}[ht]
\centering
\includegraphics[width=\textwidth]{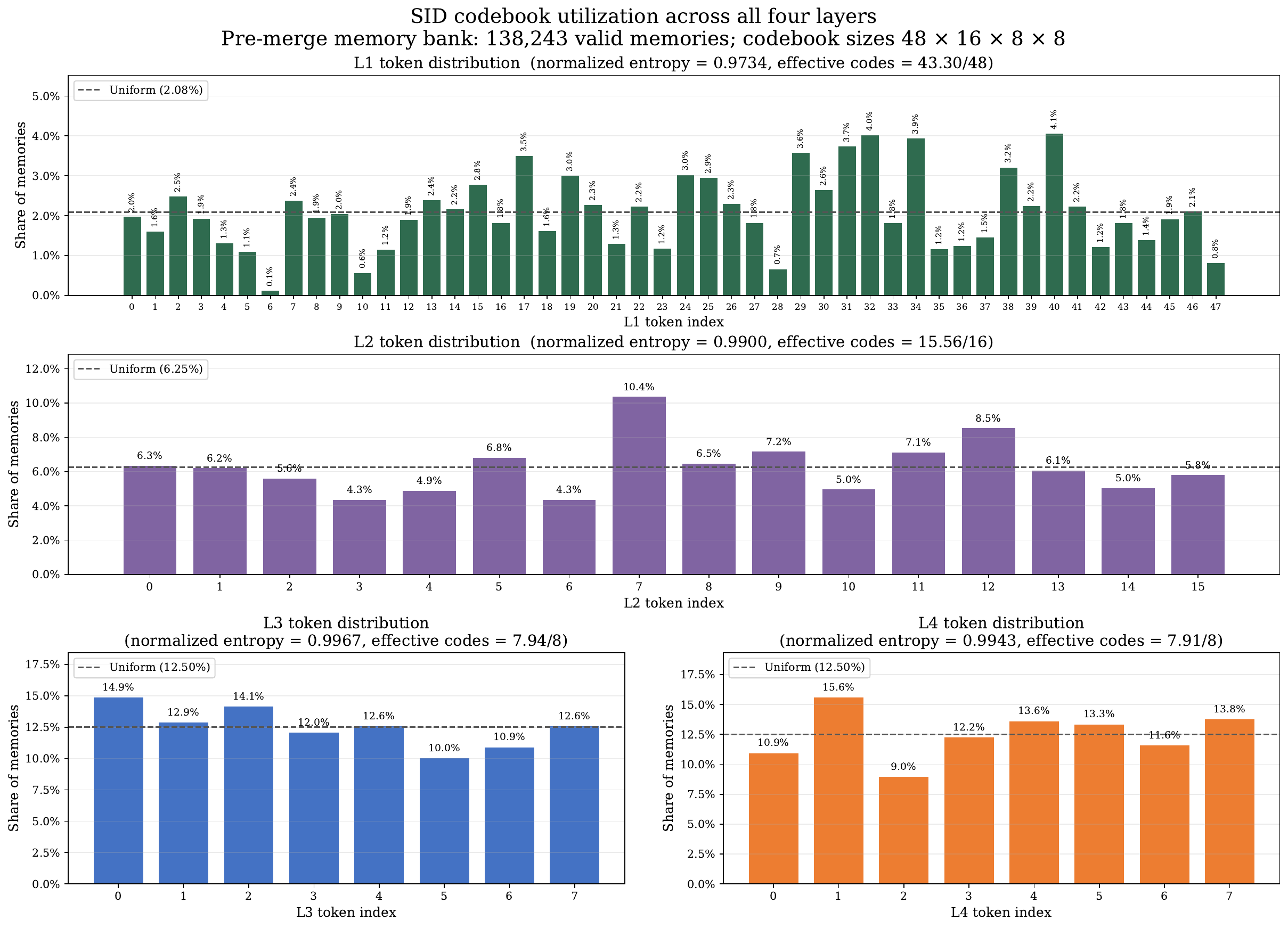}
\caption{SID codebook utilization across four levels (138{,}243 pre-merge entries). Bars show the fraction assigned to each token index; dashed line = uniform baseline. Normalized entropy and effective code counts annotated per level.}
\label{fig:codebook_dist}
\end{figure}

We analyze token-level utilization of the 4-level codebook $(48, 16, 8, 8)$ across the 138{,}243 pre-merge entries. Figure~\ref{fig:codebook_dist} reports the share of memories assigned to each token index per level, with normalized entropy and effective code counts.

Key observations:
\begin{itemize}[leftmargin=*,itemsep=2pt]
\item \textbf{High overall utilization.} All four levels achieve normalized entropy $\geq 0.97$, indicating that the RQ-KMeans construction distributes memories broadly across the codebook rather than collapsing onto a few dominant codes. The effective code counts---43.3/48 at L1, 15.6/16 at L2, 7.9/8 at L3, and 7.9/8 at L4---confirm that nearly all codebook entries are actively used.
\item \textbf{Mild imbalance at L1.} While L2--L4 are close to uniform, L1 exhibits moderate variation (range ${\approx}0.1\%$--$4.1\%$ vs.\ the $2.08\%$ uniform baseline). \reviewThree{Several L1 tokens receive fewer assignments than the uniform baseline, motivating the adaptive codebook allocation discussed in Future Work.}
\item \textbf{Finer levels are more balanced.} As the codebook size decreases from 48 (L1) to 8 (L3/L4), the per-token distribution converges toward uniform, consistent with the residual quantization design: later levels partition progressively smaller residual spaces where the variance is more evenly spread.
\end{itemize}

These results validate the Cartesian-product design: the 4-level factorization ($48 \times 16 \times 8 \times 8 = 49{,}152$ addressable regions) achieves high utilization without balancing regularization during construction.

\subsection{Training Dynamic}
\label{sec:app_midtraining_loss}

\begin{figure}[ht]
\centering
\includegraphics[width=\textwidth, trim=0 0 0 60, clip]{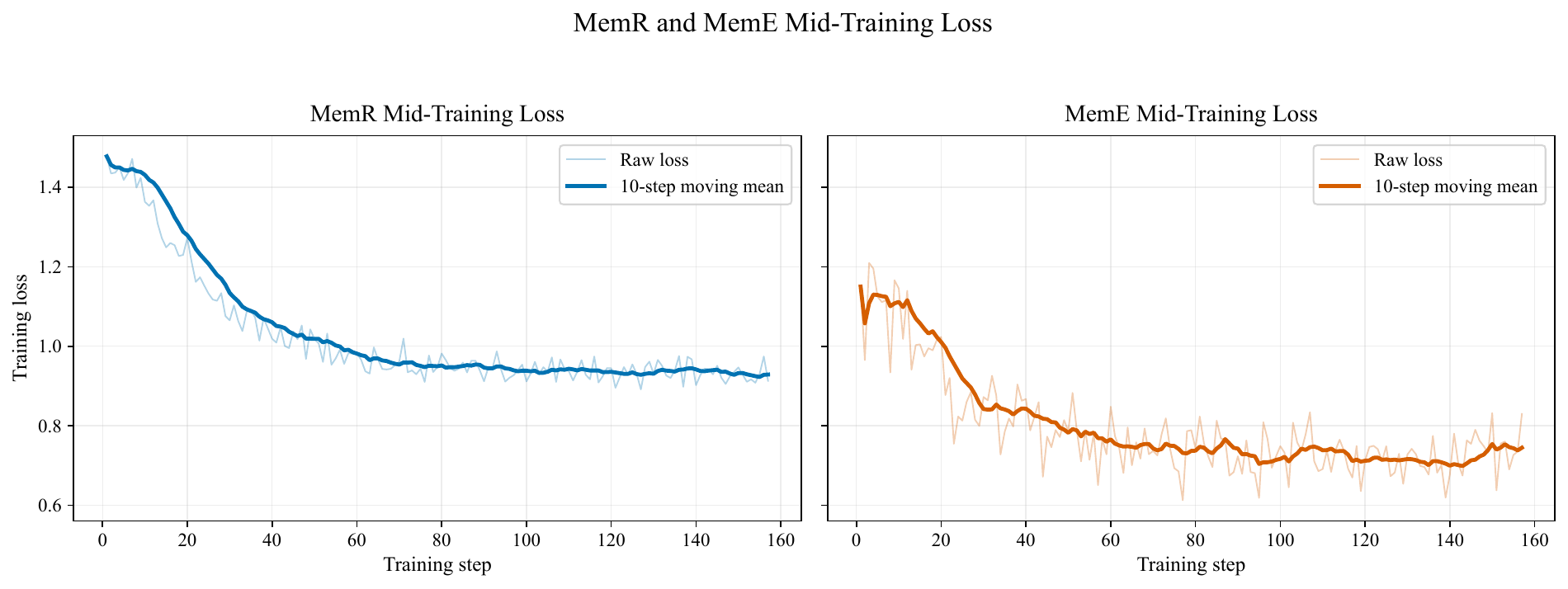}
\caption{Mid-training loss for MemRetriever (left) and MemEvolver (right). Both converge smoothly on ReAct-style traces with embedded SID generation.}
\label{fig:midtraining_loss}
\end{figure}

Figure~\ref{fig:midtraining_loss} shows training loss during mid-training for both MemRetriever and MemEvolver. This stage fine-tunes the alignment checkpoint on ReAct-style chain-of-thought traces with embedded SID generation. Both modules converge smoothly, confirming that mid-training successfully bridges SID alignment and GRPO post-training without instability.

\begin{figure*}[t]
\centering
\includegraphics[width=\textwidth]{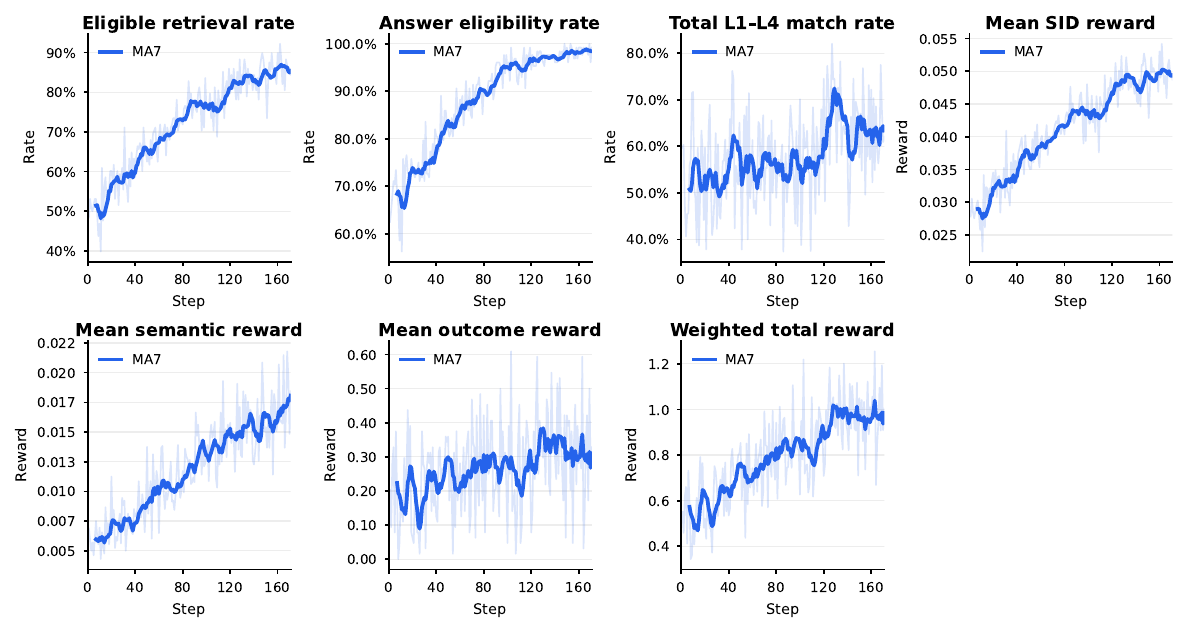}
\caption{\reviewOne{MemRetriever GRPO post-training curves (MA7-smoothed). The panels track retrieval eligibility, answer eligibility, SID matching, and four reward signals. Eligibility and training rewards generally improve over the run.}}
\label{fig:memr_training_curves}
\end{figure*}

Figure~\ref{fig:memr_training_curves} presents the MemRetriever GRPO post-training curves, tracking both eligibility rates and reward signals over the course of training.

\begin{figure*}[t]
\centering
\includegraphics[width=\textwidth]{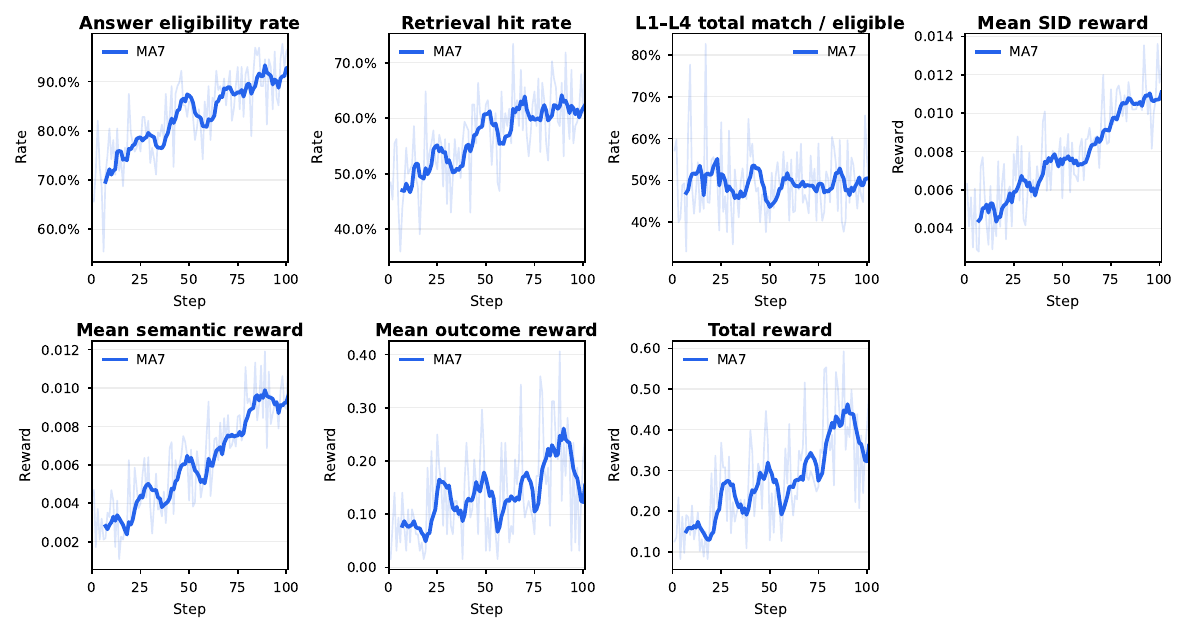}
\caption{MemEvolver GRPO post-training curves (MA7-smoothed). The first two panels report answer eligibility and retrieval hits, respectively, each divided by the total number of rollouts. The remaining panels track SID matching, SID and semantic rewards, outcome reward, and total reward.}
\label{fig:meme_training_curves}
\end{figure*}

Figure~\ref{fig:meme_training_curves} reports the corresponding MemEvolver GRPO diagnostics, covering answer eligibility, retrieval hits, SID matching, and the reward signals used to train the evolution policy.

\paragraph{StackPlanner RL training diagnostics.}
\label{sec:app_central_rl}

We report LoRA training diagnostics for StackPlanner on Search, ALFWorld, and WebShop. The Search and WebShop runs optimize the central agent with oracle memory and oracle actions, respectively. The ALFWorld run instead optimizes the Interactor while keeping the central agent frozen. It uses mixed memory conditions, with four oracle-assisted and four no-memory rollouts per query. These curves describe task-execution training, not MemRetriever/MemEvolver training or held-out performance of the complete \methodname{} system.

The runs contain 159, 65, and 131 training steps, covering 3,776, 1,035, and 1,042 training queries on Search, ALFWorld, and WebShop, respectively. Search and WebShop use four rollouts per query; ALFWorld uses eight, yielding 8,280 rollouts over one complete epoch and 65 optimizer updates. For Search, we retain the committed training lineage and exclude three superseded log records. The horizontal axis denotes the logged training-step counter for each run.

\begin{figure}[t]
\centering
\includegraphics[width=\textwidth]{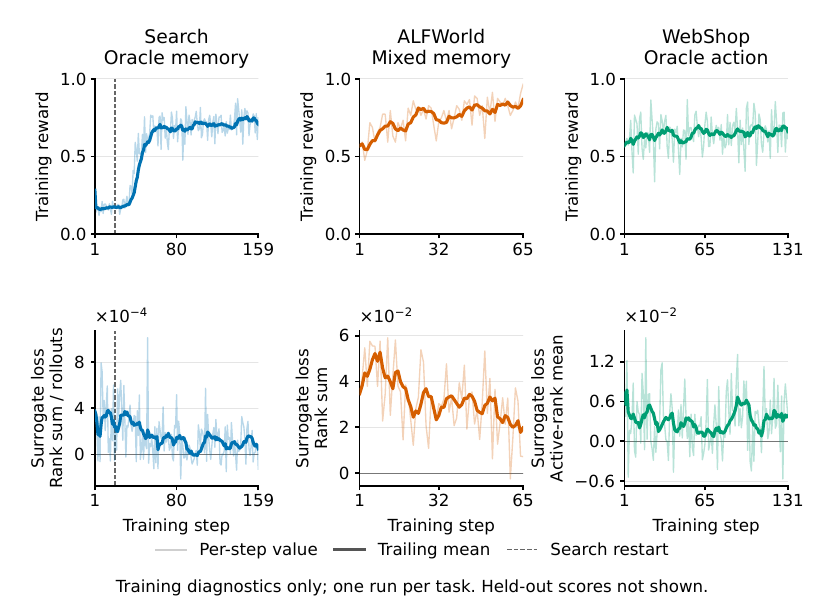}
\caption{StackPlanner RL training diagnostics. Search and WebShop use oracle-assisted central-agent training, while ALFWorld trains the Interactor with a 1:1 mixture of oracle-assisted and no-memory rollouts. Top panels show mean training reward and bottom panels show logged GRPO surrogate loss. The dashed line marks the Search restart after checkpoint 20.}
\label{fig:central_rl_training}
\end{figure}

As shown in Figure~\ref{fig:central_rl_training}, light lines denote per-step values and dark lines trailing means, using task-specific smoothing windows: rollout-weighted 10-step for Search, unweighted 5-step for ALFWorld, and unweighted 10-step for WebShop. Shorter prefixes are used before a full window is available. Loss normalization differs across tasks, so loss magnitudes are not directly comparable. Each column represents a single run without across-seed uncertainty estimates.

Search reward is token F1, whereas the interactive tasks use their logged task rewards. The sampled training queries change across steps, and the GRPO surrogate losses are not supervised cross-entropy losses. Consequently, these curves alone neither establish held-out improvements nor isolate the contribution of learned memory. They are not used to populate the main result tables. De-identified per-step data and the plotting script are retained with the manuscript sources.
\clearpage

\subsection{Ablation Study}
\label{sec:ablation}

Table~\ref{tab:ablation} provides a component-level analysis of \methodname{} under the same self-evolution setting. We evaluate each variant on 200 samples for three complete evolution rounds, \reviewTwo{holding the backbone, harness, planner, and initial bank fixed}.

\begin{table*}[htbp]
\captionof{table}{Ablation of key \methodname{} components.}
\label{tab:ablation}
\centering
\begin{tabular}{l | l c c@{}}
\toprule
\rowcolor{gray!30}
\textbf{Variant} & description & \makecell{Avg.\\Accuracy} & \makecell{Token F1} \\
\midrule
\methodname{} & full & \textbf{49.5} & \textbf{52.1} \\
\midrule
w/o MemEvolver &frozen memory bank & 46.5 & 46.8 \\
\reviewOne{Discriminative retrieval} & TF-IDF + evolution & 40.5 & 43.4 \\
\reviewOne{TF-IDF} & \reviewOne{frozen memory bank} & \reviewOne{43.0} & \reviewOne{--} \\
\reviewOne{Discriminative retrieval} & embedding + evolution & 37.0 & 41.2 \\
\reviewOne{Dense embedding} & \reviewOne{frozen memory bank} & \reviewOne{44.5} & \reviewOne{--} \\
w/o MemRetriever &query only, no memory & 41.0 & 45.2 \\
w/o RL post-training &MT MemRetriever + MT MemEvolver & 38.5 & 42.6 \\
w/o beam search &greedy SID decoding & 42.0 & 45.7 \\
\bottomrule
\end{tabular}
\end{table*}

\Cref{tab:ablation} isolates the contribution of each major component under the same three-round self-evolution protocol. The \emph{w/o MemEvolver} variant freezes the memory bank, removing experience refinement while retaining memory retrieval. The \emph{w/ discriminative retrieval} variant replaces generative SID retrieval with a content-based retriever, instantiated using an embedding model or TF-IDF, while keeping memory evolution unchanged. The \emph{w/o MemRetriever} variant removes memory access entirely and feeds only the original query to the task model. The \emph{w/o RL post-training} variant uses the \reviewOne{mid-trained} MemRetriever and MemEvolver checkpoints without reinforcement-learning optimization. Finally, the \emph{w/o beam search} variant replaces beam-based SID generation with greedy decoding.

Together, these variants separate the effects of experience evolution, symbolic retrieval, policy optimization, and decoding strategy. In particular, full vs.\ \emph{w/o MemEvolver} measures the benefit of iterative memory refinement; full vs.\ \emph{w/ discriminative retrieval} measures the effect of generative SID-based retrieval under the same evolution process; and full vs.\ \emph{w/o MemRetriever} measures the overall contribution of external memory access.

{\color{ReviewThree}%
\paragraph{Frozen dense keys with evolving payloads.}
\label{sec:app_frozen_key}
We fix the 30{,}577-entry initial bank, planner, MemEvolver, and inference budget. The dense baseline retrieves five frozen initial keys by cosine similarity and reads their latest payloads. Over 1{,}500 NQ/Bamboogle update steps, accuracy on the same 200 HotpotQA/2WikiMultiHopQA probes falls from $47.0\%$ to $42.0\%$ for the frozen-key baseline and rises from $58.0\%$ to $61.0\%$ for \methodname{}. Content-relevance hit rate, scored by a Qwen2.5-7B judge on a 0--3 scale, falls from $59.0\%$ to $52.0\%$ for frozen keys and rises from $59.0\%$ to $67.0\%$ for \methodname{}. Both systems preserve existing addresses; the comparison tests how the two routing pipelines use their evolving payloads.
}

\subsection{GPQA Evaluation Protocol}
\label{sec:app_sid_language_protocol}

The GPQA experiment tests whether SID tokens support downstream reasoning
without access to experience text, and whether rewriting retrieved experience
improves its utility. We compare seven input conditions and report answer
accuracy in Figure~\ref{fig:symbolic_composition}.

\paragraph{Experimental conditions.}
For each query, MemRetriever generates $k$ candidate SIDs with beam size
$k\in\{1,5,10\}$ \reviewTwo{without BGE reranking}. Symbolic inputs follow \emph{level-first} order: all
first-level tokens precede the second-level tokens, followed by subsequent
levels where applicable. Each condition supplies the query together with
the following information.
\begin{enumerate}[nosep,leftmargin=*]
  \item \textbf{Raw}: no additional information.
  \item \textbf{+$L_1$}: first-level SID tokens only.
  \item \textbf{+$L_1{..}L_2$}: the first two SID levels.
  \item \textbf{+$L_1{..}L_3$}: the first three SID levels.
  \item \textbf{+Full SID}: complete four-level SIDs, without experience text.
  \item \textbf{+Rewrite}: a task-adaptive rewrite of experience retrieved using the complete SIDs.
  \item \textbf{+Retrieved Memory}: experience text retrieved using the complete SIDs, without rewriting.
\end{enumerate}
Conditions~1--5 vary SID depth without providing a memory payload.
Conditions~5--7 compare complete symbolic addresses with rewritten and
unmodified experience text. Both text conditions access the memory bank.
Rewrite follows the lookup-then-rewrite procedure in
Section~\ref{sec:memory_operations}, so it does not measure reconstruction
of experience from SID tokens alone.

\paragraph{Results and interpretation.}
Complete SIDs improve accuracy from the raw-query baseline of $48.32\%$ to
$54.06$--$54.66\%$ across the three beam settings, corresponding to
$5.74$--$6.34$ percentage points. Rewritten experience reaches
$55.65$--$56.91\%$, compared with $56.09$--$56.63\%$ for directly retrieved
text. Direct retrieval performs better at $k=1$ and $k=5$; rewriting performs
better at $k=10$. The comparison therefore supports the utility of SID input
in this evaluation, with no consistent advantage from rewriting retrieved
text. Independent semantic axes and generalization to unseen SID combinations
remain unestablished. Assessing these properties requires additional controls,
including held-out combinations and shuffled or reassigned SID tokens.

{\color{ReviewTwo}%
\paragraph{Multi-SID and fixed-payload controls.}
On 185 questions spanning nine QA and mathematical reasoning benchmarks, constrained SID beam search generates up to 50 candidates, which BGE reranks. We supply one, five, or ten reranked SIDs without experience text and obtain accuracies of $23.78\%$ (44/185), $25.95\%$ (48/185), and $30.27\%$ (56/185), respectively. In a separate GPQA control with the experience text held fixed, adding SID tokens raises accuracy from $56.0\%$ to $57.0\%$ with retrieved text and from $57.0\%$ to $59.0\%$ with rewritten text.
}

{\color{ReviewThree}%
\paragraph{SID perturbation.}
On 100 GPQA questions, we supply the SID of a successful seed trajectory without experience text. Accuracy is $56.0\%$ with the reference SID, $55.0\%$ after perturbing levels 3--4, and $51.0\%$ after perturbing levels 1--2, averaging each perturbation condition over index swaps, adjacent substitutions, and random substitutions.
}

\section{Case Studies}
\label{sec:app_cases}

\reviewTwo{We present three examples of symbolic reasoning and online evolution, followed by three execution traces. The cases are drawn from experiments with different configurations.}

\subsection{Symbolic Reasoning and Online Evolution Traces}
\label{sec:app_symbolic_cases}

The following qualitative cases illustrate how SIDs can be manipulated as
structured symbols and how the same memory address can accumulate more reliable
experience during online evolution. We transcribe the operation, intermediate
memories, and generated result so that each example can be read as a complete
case rather than as a schematic diagram. These examples clarify the operation
semantics but do not replace the quantitative evaluation.

\begin{tcolorbox}[breakable, enhanced, colback=orange!5, colframe=orange!40, title={\textbf{Symbolic Case 1: Address Fusion ($A+B\rightarrow C$)}}, fonttitle=\bfseries\small, coltitle=black, colbacktitle=orange!15, arc=2pt, boxrule=0.5pt]
\textbf{Inputs.} Two source experiences are addressed by
\texttt{<SID\_L1\_33><SID\_L2\_4><SID\_L3\_14><SID\_L4\_6>} and
\texttt{<SID\_L1\_32><SID\_L2\_23><SID\_L3\_10><SID\_L4\_2>}. The first records a failure
caused by relying on a single source (e.g., Wikipedia) when checking a
director's birth date. The second records a related failure in which conflicting
source data led to an incorrect date. Both entries therefore emphasize
multi-source verification and arithmetic or timeline checks.

\textbf{Operation.} The fusion prompt asks the model to combine the shared
problem type, evidence requirements, failure modes, and reusable procedures,
while rejecting unsupported facts and avoiding a simple concatenation of the two
entries.

\textbf{Output.} The generated experience is written to
\texttt{<SID\_L1\_6><SID\_L2\_2><SID\_L3\_5><SID\_L4\_6>} and states: use multiple
authoritative sources (official biographies, film databases, and public records),
cross-check dates against the person's work and career timeline, resolve source
conflicts in favor of the more reliable source, and confirm the final answer
with at least two independent references.

\textbf{Interpretation.} The target is not a textual concatenation of the two
sources. The model extracts their common verification structure and writes it to
a new symbolic address, showing how multiple addresses can provide ingredients
for a reusable experience.
\end{tcolorbox}

\begin{tcolorbox}[breakable, enhanced, colback=blue!5, colframe=blue!40, title={\textbf{Symbolic Case 2: Address Decomposition ($C\rightarrow A,B$)}}, fonttitle=\bfseries\small, coltitle=black, colbacktitle=blue!12, arc=2pt, boxrule=0.5pt]
\textbf{Target.} The target experience at
\texttt{<SID\_L1\_25><SID\_L2\_13><SID\_L3\_2><SID\_L4\_4>} concerns identifying a person's
main occupation and returning a concise, normalized description. Its abstract
procedure is to identify the entity, obtain a short biography, extract the key
occupation label, and normalize multiple occupations when necessary.

\textbf{Operation.} A judge compares candidate source pairs according to shared
or complementary skills. The selected pair is
\texttt{A=<SID\_L1\_14><SID\_L2\_13><SID\_L3\_2><SID\_L4\_4>} and
\texttt{B=<SID\_L1\_25><SID\_L2\_2><SID\_L3\_7><SID\_L4\_4>}. A contributes entity
disambiguation for generic roles or names; B contributes structured biographical
search and normalized occupation extraction.

\textbf{Output.} The reconstructed target
\texttt{C'=<SID\_L1\_25><SID\_L2\_13><SID\_L3\_2><SID\_L4\_4>} preserves the target task
and makes the supporting steps explicit: identify ambiguous entities or
occupations, disambiguate with context or authoritative sources, extract the
most specific supported occupation, and return it in a standardized form.

\textbf{Interpretation.} Decomposition retrieves complementary subskills rather
than copying one memory entry. The resulting card preserves the target task
while making the supporting operations explicit and reusable.
\end{tcolorbox}

\begin{tcolorbox}[breakable, enhanced, colback=green!5, colframe=green!40!black, title={\textbf{Online Evolution Case: \texttt{<SID\_L1\_6><SID\_L2\_19><SID\_L3\_7><SID\_L4\_2>}}}, fonttitle=\bfseries\small, coltitle=black, colbacktitle=green!12, arc=2pt, boxrule=0.5pt]
This case follows one literary-question memory through iterations 0, 3, and 5.
The address remains fixed while the payload is revised after new trajectories.

\begin{description}[leftmargin=*,style=nextline]
\item[Iteration 0.] The initial experience says that literary-work questions
require identifying the work, author, and relevant characters, then checking the
requested details against the original text. It warns that secondary summaries
can mix up similar characters and names.
\item[Iteration 3.] A later trajectory adds an explicit source hierarchy: recall
the canonical work and character first, consult the original text or an
authoritative literary reference, and reject conclusions supported only by
secondary analysis when the original and the reference disagree.
\item[Iteration 5.] The evolved entry specializes the rule for questions about
characters and roles. It requires extracting the exact name and role from the
primary text, checking the surrounding context and timeline, and re-running the
original-text verification whenever the character identity is ambiguous.
\end{description}

\textbf{Interpretation.} The iterations show selective revision rather than
blind appending. Each update retains the transferable literary-analysis
procedure, removes ambiguity, and adds verification requirements learned from
the new trajectory. The case therefore illustrates the invariance--evolution
trade-off: content is corrected and enriched without invalidating the learned
address used to retrieve it.
\end{tcolorbox}

\begin{tcolorbox}[breakable, enhanced, colback=orange!5, colframe=orange!40, title={\textbf{Case 1: Factual Retrieval and Multi-hop Reasoning}}, fonttitle=\bfseries\small, coltitle=black, colbacktitle=orange!15, arc=2pt, boxrule=0.5pt]

\textbf{Query:} ``Simpleton Tunnel links Italy with which country?''

\medskip
\textbf{MemRetriever execution:}
\begin{itemize}[leftmargin=*,itemsep=1pt]
\item Memory Lookup: $\langle$SID\_L1\_56$\rangle$$\langle$SID\_L2\_1$\rangle$$\langle$SID\_L3\_1$\rangle$$\langle$SID\_L4\_6$\rangle$
\item \textbf{Retrieved experience:}
\begin{itemize}[itemsep=0pt]
\item \emph{Problem addressed:} Handling geographic-name queries; quickly identifying target place-name entities and query intent (geographic location, administrative division, historical changes, transportation, surrounding facilities, neighborhood districts, etc.).
\item \emph{Practice:} Strictly follow standard geographic naming conventions; precisely match official administrative divisions and geographic information; avoid confusing colloquial or informal names; support multi-language translation and phonetic/approximate name variants.
\end{itemize}
\item \textbf{Rewritten experience:} Recognize place-name misspellings (especially for cross-border infrastructure like tunnels and bridges). \emph{Practice:} Parse the target entity in the question, note common spelling variants; search standard names and key evidence (location, endpoints); identify the two connected countries (excluding the one already given); output the other country as the answer.
\end{itemize}

\medskip
\textbf{StackPlanner execution:}
\begin{itemize}[leftmargin=*,itemsep=1pt]
\item \textbf{Input:} [Query] + [Rewritten experience]
\item \textbf{Reasoning:} The user asks about the ``Simpleton Tunnel'' (likely ``Simplon Tunnel'') connecting Italy to another country. Since the question concerns external factual information, a web-search verification is appropriate.
\item \textbf{Dispatches Research Agent:} Search for the correct name of ``Simplon Tunnel'' and the countries it connects, focusing on basic infrastructure linking Italy to neighboring countries.
\item \textbf{Research Agent findings:}
\begin{itemize}[itemsep=0pt]
\item Simplon Tunnel -- Wikipedia: ``connects Brig and Domodossola, Italy, through the Alps'' (score=0.891)
\item Simplon Pass -- Wikipedia: ``connects Brig in the canton of Valais with Domodossola in Piedmont (Italy)'' (score=0.788)
\end{itemize}
\item \textbf{Research report returned to lead agent:} The correct name is the \textbf{Simplon Tunnel} (German: Simplontunnel; Italian: Traforo del Sempione). It connects \textbf{Brig, Switzerland} and \textbf{Domodossola, Italy}, under the Lepontine Alps. Length: 19.8\,km, 1906--1982 the world's longest railway tunnel. Construction: started 1898, completed with two single-track bores 15 years apart. Purpose: high-altitude trans-border freight and passenger rail. Connected countries: \textbf{Switzerland} (Valais) and \textbf{Italy} (Piedmont).
\end{itemize}

\medskip
\textbf{Answer:} \textcolor{red!70!black}{\textbf{Switzerland}}
\end{tcolorbox}

\begin{tcolorbox}[breakable, enhanced, colback=blue!5, colframe=blue!40, title={\textbf{Case 2: Mathematical Reasoning + Memory Evolution}}, fonttitle=\bfseries\small, coltitle=black, colbacktitle=blue!12, arc=2pt, boxrule=0.5pt]

\textbf{Query:} ``A coach is arranging a starting 5-player lineup for a basketball team. There are 12 players in total, and Ace and Zeppo must be in the starting lineup. How many different lineups are there?''

\medskip
\textbf{StackPlanner execution:}
\begin{itemize}[leftmargin=*,itemsep=1pt]
\item \textbf{Input:} [Query]
\item \textbf{Reasoning:} The user asks: given two players already confirmed as starters, how many ways to form the remaining lineup. This is a fixed-element combination counting problem; prior experience can be directly applied.
\item \textbf{Dispatches Research Agent:} Calculate the number of ways to choose 3 starters from the remaining 10 players given Ace and Zeppo are fixed.
\item \textbf{Research findings:}
\begin{itemize}[itemsep=0pt]
\item Fixed starters: 2 (Ace, Zeppo)
\item Remaining player pool: 10
\item Additional players to select: 3
\item Selection order does not affect the result $\to$ use combination
\item $C(10,3) = 120$
\end{itemize}
\item \textbf{Conclusion:} There are 120 different starting lineups.
\end{itemize}

\medskip
\textbf{MemEvolver execution:}
\begin{itemize}[leftmargin=*,itemsep=1pt]
\item MemE Lookup: $\langle$SID\_L1\_3$\rangle$$\langle$SID\_L2\_8$\rangle$$\langle$SID\_L3\_4$\rangle$$\langle$SID\_L4\_2$\rangle$
\item \textbf{Retrieved experience at this SID:} Handling combination counting problems. \emph{Practice:} Compute the number of selectable objects and the number still to be chosen; determine whether order matters; if order-independent, use $C(n,k)$.
\item \textbf{Rewritten (evolved) experience:} \underline{Handle fixed-member combination selection problems} (e.g., fixed starting players, fixed candidates, fixed seat positions). \emph{Practice:} Identify which members are already fixed, so they need not be re-selected; update the remaining candidate count and remaining selection count; determine if the problem is combination or permutation; if order-independent, use $C(n,k)$; output the final count and verify that it satisfies all constraints.
\end{itemize}

\medskip
\textbf{Answer:} \textcolor{red!70!black}{\textbf{120}}
\end{tcolorbox}

\begin{tcolorbox}[breakable, enhanced, colback=green!5, colframe=green!40!black, title={\textbf{Case 3: Memory Evolution at an Unpopulated Leaf}}, fonttitle=\bfseries\small, coltitle=black, colbacktitle=green!12, arc=2pt, boxrule=0.5pt]

\textbf{Query:} ``Ian won \$100 in a lottery. He decided to use it to pay off debts. First he paid Colin \$20; then he paid Helen double what he paid Colin; finally he paid Benedict half what he paid Helen. How much does Ian have left?''

\medskip
\textbf{StackPlanner execution:}
\begin{itemize}[leftmargin=*,itemsep=1pt]
\item \textbf{Input:} [Query]
\item \textbf{Reasoning:} The user asks about the remaining amount after sequential payments involving proportional relationships. This requires step-by-step sequential computation.
\item \textbf{Dispatches Research Agent:} Compute Ian's remaining balance after three payments.
\item \textbf{Research findings:}
\begin{itemize}[itemsep=0pt]
\item Initial amount: \$100
\item Payment to Colin: \$20
\item Payment to Helen: $20 \times 2 = \$40$
\item Payment to Benedict: $40 \div 2 = \$20$
\item Total paid: $20 + 40 + 20 = \$80$
\end{itemize}
\item \textbf{Conclusion:} Remaining = $100 - 80 = \$20$.
\end{itemize}

\medskip
\textbf{MemEvolver execution:}
\begin{itemize}[leftmargin=*,itemsep=1pt]
\item MemE Lookup: $\langle$SID\_L1\_20$\rangle$$\langle$SID\_L2\_11$\rangle$$\langle$SID\_L3\_0$\rangle$$\langle$SID\_L4\_3$\rangle$
\item \textbf{Retrieved experience at this SID:} \textbf{Leaf unpopulated}---no existing experience stored.
\item \textbf{Newly inserted experience:} Handle problems involving initial amounts, sequential payments, and proportional relationships between payment amounts. \emph{Practice:} First record the initial amount; describe each payment in order following the problem statement; for proportional terms like ``double'' or ``half,'' always compute from the immediately preceding payment amount; complete one payment at a time, updating the remaining balance; finally output the remaining amount and verify that all payment amounts sum correctly.
\end{itemize}

\medskip
\textbf{Answer:} \textcolor{red!70!black}{\textbf{\$20}}
\end{tcolorbox}









\section{Full Prompt Specifications}
\label{sec:app_full_prompts}

For completeness and reproducibility, this appendix collects the verbatim prompts used across the \methodname{} pipeline: experience generation, trajectory summarization, skill-cluster and SID-leaf synthesis, embedding generation, StackPlanner outcome-reward rollouts, the MemRetriever (MemR) runtime, and the MemEvolver (MemE) runtime. Placeholders in braces (e.g., \texttt{\{question\}}) are filled at runtime.

\subsection{Experience Generation Prompts}

\begin{promptbox}{General-purpose datasets}
\begin{lstlisting}[style=promptstyle]
You are solving a benchmark training question.

Answer the question as accurately as possible. Use StackPlanner tools or subagents only when they are genuinely useful.
Leave a compact, observable solve trace before the final answer so this run can be distilled into reusable experience later.

For factual, entity-comparison, multi-hop, or time-sensitive questions, do not answer from a bare guess. Verify the key entities/attributes with available tools or with explicit evidence from the provided context. If no tool is needed or available, still write the evidence/checks you used.

Use this visible structure:
Evidence:
- <key fact, calculation, or tool result used>
Verification:
- <answer-format check, entity disambiguation, arithmetic check, or uncertainty note>

Question:
{question}

Return your final answer in this exact trailing form:
Final answer: <answer>
\end{lstlisting}
\end{promptbox}

\begin{promptbox}{WebShop}
\begin{lstlisting}[style=promptstyle]
You are collecting a WebShop benchmark experience.

Goal: satisfy the shopping instruction by selecting the best matching product or product option.
You must use the available WebShop benchmark tools when they exist:
1. call `webshop_search` with a query and the full instruction,
2. call `webshop_inspect` on promising ASINs,
3. call `webshop_select` with the final ASIN and selected options to obtain the benchmark reward/success signal.
If these tools are not available, explicitly say the result is an offline selection and do not claim verified success.

Shopping instruction:
{question}

Available metadata excerpt:
{metadata}

Focus on reusable SOP details: query formulation, attribute filtering, price/option matching, product-page verification, final selection, and the observed reward/success signal.

Return your final answer in this exact trailing form:
Final answer: <product id / asin / selected product>
\end{lstlisting}
\end{promptbox}

\begin{promptbox}{ALFWorld}
\begin{lstlisting}[style=promptstyle]
You are collecting an ALFWorld/TextWorld household-task experience.

Goal: complete the embodied household task. The ALFWorld tools are enabled in this StackPlanner session.
You must call the tools; do not produce an offline plan before trying them.
1. call `alfworld_reset` with the task filename,
2. inspect the returned observation and admissible commands,
3. call `alfworld_step` repeatedly with admissible commands until the environment returns done/won, or until progress is clearly blocked,
4. report the final score/done/won signal.
Only mark the result as offline/unverified if `alfworld_reset` itself returns an explicit tool error. If a prior memory says tools were unavailable, ignore that memory for this run.

Task/environment description:
{question}

Available metadata excerpt:
{metadata}

Use ALFWorld-style action thinking: inspect the room, find relevant objects, open/close containers if needed, take objects, navigate to target receptacles, put/use/clean/heat/cool as required, and verify completion from the environment score/done/won fields. Preserve reusable SOP details and common failure checks.

Return your final answer in this exact trailing form:
Final answer: <completed / failure reason with final score/done/won>
\end{lstlisting}
\end{promptbox}

\subsection{Trajectory Summarization Prompts}
\label{sec:app_trajectory_prompts}

\begin{promptbox}{Single-trajectory summary}
\begin{lstlisting}[style=promptstyle]
You compress agent execution traces into faithful, reusable trajectory summaries.

Return only valid JSON. All values must be strings. Use exactly these fields:
run_id, evaluation, prediction, trajectory_summary, decisive_steps, evidence_or_calculation, success_signal, failure_signal, low_evidence_notes.

Rules:
- Preserve what the run actually did; do not invent tools, sources, calculations, or evidence.
- trajectory_summary should be a numbered step list, not a one-line abstract.
- decisive_steps should name the step(s) that most affected the final answer.
- evidence_or_calculation should record the concrete evidence, equation, lookup, environment observation, or answer-format check visible in the run.
- success_signal should explain why the run is reusable if it is correct; otherwise say no verified success trajectory exists.
- failure_signal should explain the likely divergence if the run is wrong/unverified/low-evidence; for correct runs, record realistic residual risks.
- If the run only has a direct final answer, mark it low-evidence instead of pretending a procedure was observed.
\end{lstlisting}
\end{promptbox}

\begin{promptbox}{Multi-trajectory summary}
\begin{lstlisting}[style=promptstyle]
You extract rich, reusable experience from repeated benchmark attempts.

Return only valid JSON. All values must be strings. Use exactly these fields:
skill_name, problem_type, problem_addressed, applicability, preconditions, procedure, decision_rules, verification_checks, failure_recovery, anti_patterns, practice, lessons_learned, success_experience, failure_reflection, steps_summary, summary, retrieval_cues, failure_modes, verification, evidence.

Write a useful memory for a future agent, not a generic checklist.

Requirements:
- Use the ground truth/reference as the anchor.
- Compare correct, failed, and unverified trajectories.
- Treat a citation as evidence only when the visible source content directly supports the claim; preserve low-evidence caveats instead of upgrading weak or tangential sources into verified facts.
- success_experience should be 2-4 short paragraphs when enough evidence exists: what correct trajectories actually did, what was decisive, what variation is safe, and how to reuse it.
- failure_reflection should be 2-4 short paragraphs when failures exist: wrong predictions, divergence point, likely root cause, warning signs, and exact repair protocol.
- summary should be 3-6 compact paragraphs with these ideas: outcome contrast, success pattern, failure pattern, reusable recipe, verification gates.
- procedure and practice should be executable enough for a future agent to follow.
- Do not claim success for WebShop/ALFWorld/offline interactive tasks without a real reward or environment signal.
- Do not average the attempts into vague advice; preserve the contrast that made the group informative.
\end{lstlisting}
\end{promptbox}

\begin{promptbox}{Interactive-dataset rich summary}
\begin{lstlisting}[style=promptstyle]
Generate a reusable rich summary memory for one real interactive benchmark run.

Return only valid JSON with exactly one string field: rich_summary.

This rich_summary is the main text embedded by RQK, so it must be model-written, diverse, and reusable rather than a fixed field dump.

Requirements:
- Ground the memory in the actual run data, tool observations, and environment reward/done/won/success signals.
- Write natural field-note prose, not a repeated template with the same labels for every sample.
- Preserve a portable SOP: what to inspect first, how to choose actions/products, what gates prove success, and what warning signs should trigger repair.
- For WebShop, include search-query strategy, attribute/option/price checks, product-page verification, final select behavior, and reward/success evidence when visible.
- For ALFWorld, include reset/task-file context, observation/admissible-command use, navigation/object/container operations, final score/done/won evidence, and failure checks when progress stalls. Preserve the original task verb: if the task says "look at", do not rewrite it as a "take" task; describe any take action only as an observed final action.
- Do not claim success unless the payload includes a verified reward/success/won signal.
- Do not invent counterfactual mechanics.
- Do not copy raw JSON or full transcripts; distill them into actionable memory.
- Do not merely restate the exact target ASIN/file as the skill. The reusable part is the decision process.
- Keep it compact enough for retrieval: usually 4-8 short paragraphs or tight bullets, 350-900 words.

Payload:
{rich_json}
\end{lstlisting}
\end{promptbox}

\subsection{Skill-Cluster and SID-Leaf Synthesis Prompts}
\label{sec:app_sid_synthesis_prompts}

\begin{promptbox}{Skill-cluster synthesis}
\begin{lstlisting}[style=promptstyle]
You are synthesizing reusable agent skill memories for RQ-KMeans training.

Below are skill records that were assigned to the same semantic cluster. They may mention
different repositories, file paths, brands, specific commands, or reference files. Your task is
to extract the shared, reusable problem-solving pattern and remove case-specific details.

Write one generalized experience record that:
1. Captures the common problem type and when to use it.
2. Preserves reusable procedures, decision rules, verification checks, and pitfalls.
3. Keeps important technology/domain names only when they are common to the cluster.
4. Avoids source paths, sample ids, one-off repository names, reference filenames, and long command dumps.
5. Is diverse and broadly reusable, not a vague label.

Return ONLY a valid JSON object with these string fields:
skill_name, problem_type, problem_addressed, applicability, preconditions, procedure,
decision_rules, practice, lessons_learned, verification_checks, retrieval_cues, summary

Cluster metadata:
- cluster_id: {cluster_id}
- original_records: {record_count}
- unique_descriptions: {unique_count}
- top_categories: {categories}
- top_skill_names: {skill_names}
\end{lstlisting}
\end{promptbox}

\begin{promptbox}{SID-leaf experience merge}
\begin{lstlisting}[style=promptstyle]
You are an expert at synthesizing problem-solving experiences. Below are {n} experience summaries that all belong to the same semantic cluster (SID: {sid}). They describe similar types of problems and strategies.

Your task: Merge them into ONE comprehensive experience summary that:
1. Captures the common problem type and core strategy
2. Preserves the most useful specific techniques and decision rules
3. Includes both success patterns and failure pitfalls if mentioned
4. Is concise but complete --- no redundancy

Output ONLY the merged experience summary, no preamble.
\end{lstlisting}
\end{promptbox}

\subsection{Embedding Generation Prefix}
\label{sec:app_embedding_prompt}

\begin{promptbox}{RQ-KMeans embedding prefix}
\begin{lstlisting}[style=promptstyle]
Instruct: Encode the following agent experience as a reusable problem-solving memory for semantic retrieval and SID clustering. Focus on the problem type, reusable skill, preconditions, procedure, decision rules, verification checks, outcome, and failure recovery. Prefer semantic problem-solving patterns over dataset names, run ids, wording artifacts, or incidental entities unless they are essential to the skill.
Query: {rich_summary}
\end{lstlisting}
\end{promptbox}

\subsection{StackPlanner Outcome-Reward Prompts}
\label{sec:app_outcome_prompts}

The StackPlanner is prompted as the LangGraph lead agent during outcome-reward rollouts. The \texttt{\{memory\_section\}} placeholder is filled with the with-memory or the baseline block shown below.

\begin{promptbox}{WebShop rollout}
\begin{lstlisting}[style=promptstyle]
Solve this WebShop product-identification task. Use web_search when useful. Identify the single product that best satisfies every requested attribute. Return only its 10-character ASIN (for example B012345678), with no explanation.

{memory_section}WebShop request:
{query}

# ---- {memory_section} with-memory variant ----
Retrieved reusable memory:
{retrieved_experience}

# ---- {memory_section} baseline variant ----
No retrieved memory is provided.
\end{lstlisting}
\end{promptbox}

\begin{promptbox}{ALFWorld rollout}
\begin{lstlisting}[style=promptstyle]
Solve this ALFWorld task by interacting with the real TextWorld environment. First call alfworld_reset with exactly the task_file and split shown below. Then repeatedly choose one command exactly from the latest admissible_commands and call alfworld_step. Re-plan from every observation. Continue until a tool response returns won=true, or stop after 30 step calls. Do not invent internal PDDL identifiers and do not merely print an action plan; only tool-executed environment success counts.

{memory_section}Task request:
{task_request}

task_file: {alfworld_task_file}
split: {alfworld_split}
\end{lstlisting}
\end{promptbox}

\begin{promptbox}{General QA (counterfactual baseline)}
\begin{lstlisting}[style=promptstyle]
Solve the question using your normal capabilities. No retrieved memory is provided. Return only the final answer and no explanation.

Question:
{query}
\end{lstlisting}
\end{promptbox}

\subsection{MemRetriever and MemEvolver Runtime Prompts}

\begin{promptbox}{MemRetriever system prompt}
\begin{lstlisting}[style=promptstyle]
You are MemRetriever, a memory retrieval agent.
For every query, choose exactly one four-layer Semantic ID (SID) in the exact form <SID_L1_X><SID_L2_Y><SID_L3_Z><SID_L4_W>, then make exactly one memory_lookup call whose sid_list contains only that SID. You may reason briefly in <think>...</think>. After the tool response, output exactly one non-empty <answer>...</answer> containing a concise, actionable summary of the retrieved experience. Do not call the tool again, output multiple SIDs, or copy the entire memory. If no relevant memory is returned, output <answer>No relevant experience found.</answer>
\end{lstlisting}
\end{promptbox}

\begin{promptbox}{\texttt{memory\_lookup} tool description}
\begin{lstlisting}[style=promptstyle]
type: function
function:
  name: memory_lookup
  description: >-
    Look up experience memories from the memory bank using Semantic IDs (SIDs). Returns the top-k most relevant experience summaries for each SID provided.
  parameters:
    type: object
    properties:
      sid_list:
        type: array
        items:
          type: string
        description: >-
          A one-element list containing the canonical four-layer SID selected by MemRetriever. Example: ["<SID_L1_5><SID_L2_3><SID_L3_2><SID_L4_7>"]. Do NOT use natural-language descriptions, multiple SIDs, or partial SIDs as values.
    required:
      - sid_list
\end{lstlisting}
\end{promptbox}

\begin{promptbox}{LLM relevance judge prompt}
\begin{lstlisting}[style=promptstyle]
Judge whether the retrieved reusable experience is useful for the query.
Rubric: 2 = directly useful: it contains a task-specific strategy, evidence, constraints, or procedure that would materially help answer the query;
1 = only broad task-type relevance: it is directionally related but generic;
0 = unrelated, misleading, or too generic to help.
Do not reward a shared dataset label by itself. Judge the actual content.
Return strict JSON only: {"score": 0, "reason": "short reason"}.

QUERY:
{query}

CONTENT:
{retrieved_experience}
\end{lstlisting}
\end{promptbox}

\begin{promptbox}{MemEvolver system prompt}
\begin{lstlisting}[style=promptstyle]
You are MemEvolver, a memory evolution agent.
Given an execution trajectory, you update the memory bank with evolved experience:

1. <think>: Analyse the trajectory thoroughly -- understand what task was attempted, what knowledge domain it covers, what succeeded and failed, and hypothesise which SID cluster captures this domain best. Consider the trajectory's implicit lessons.

2. <tool_call>: Call memory_lookup with your chosen SID to retrieve the existing experience.

3. <think>: Compare the trajectory's insights against the retrieved experience -- identify what is already captured, what is new, what is contradicted, and how to merge them. If no existing experience is found, derive the experience entirely from the trajectory.

4. <answer>: Produce an evolved experience that synthesises the old memory with new insights from the trajectory. The evolved experience should be reusable: general enough to transfer to similar future tasks, specific enough to be actionable. Do not copy the trajectory verbatim -- distil the transferable lessons.
\end{lstlisting}
\end{promptbox}

\end{document}